%% file: example_paper.tex
\documentclass[nohyperref]{article}
\input{math_commands.tex}

\usepackage{microtype}
\usepackage{graphicx}
\usepackage{booktabs} % for professional tables

\usepackage{hyperref}

\usepackage[accepted]{icml2022}

\usepackage{amsmath}
\usepackage{amssymb}
\usepackage{mathtools}
\usepackage{amsthm}

\usepackage[capitalize,noabbrev]{cleveref}

\usepackage{subcaption}
\usepackage{amsmath}
\usepackage{amsthm}
\usepackage{amssymb}
\usepackage{hhline}
\usepackage{multirow}
\usepackage{wrapfig}
\usepackage{caption}
\usepackage{diagbox}
\usepackage{slashbox}
\usepackage{bbm}
\usepackage{tabularx}

\usepackage{thmtools}
\usepackage{thm-restate}

\usepackage{xpatch}
\usepackage{esint}

\makeatletter
\@for\next:={int,iint,iiint,iiiint,dotsint,oint,oiint,sqint,sqiint,
  ointctrclockwise,ointclockwise,varointclockwise,varointctrclockwise,
  fint,varoiint,landupint,landdownint}\do{%
    \expandafter\edef\csname\next\endcsname{%
      \noexpand\DOTSI
      \expandafter\noexpand\csname\next op\endcsname
      \noexpand\ilimits@
    }%
  }
\makeatother

\makeatletter
\xpatchcmd{\thmt@restatable}% Edit \thmt@restatable
{\csname #2\@xa\endcsname\ifx\@nx#1\@nx\else[{#1}]\fi}% Replace this code
{\ifthmt@thisistheone\csname #2\@xa\endcsname\ifx\@nx#1\@nx\else[{#1}]\fi\else\csname #2\@xa\endcsname\fi}% with this code
{}{} % execute code for success/failure instances
\makeatother

\makeatletter
\newcommand{\pushright}[1]{\ifmeasuring@#1\else\omit\hfill$\displaystyle#1$\fi\ignorespaces}
\newcommand{\pushleft}[1]{\ifmeasuring@#1\else\omit$\displaystyle#1$\hfill\fi\ignorespaces}
\makeatother

\newcommand*{\Scale}[2][4]{\scalebox{#1}{$#2$}}%

\newcommand{\vect}[1]{\boldsymbol{#1}}
\newcommand{\RN}[1]{%
    \textup{\lowercase\expandafter{\it \romannumeral#1}}%
}

\icmltitlerunning{Deep Ensemble as a Gaussian Process Approximate Posterior}

\begin{document}

\twocolumn[
\icmltitle{Deep Ensemble as a Gaussian Process Approximate Posterior}



\icmlsetsymbol{equal}{*}

\begin{icmlauthorlist}
\icmlauthor{Zhijie Deng}{thu}
\icmlauthor{Feng Zhou}{thu}
\icmlauthor{Jianfei Chen}{thu}
\icmlauthor{Guoqiang Wu}{sdu}
\icmlauthor{Jun Zhu}{thu}
\end{icmlauthorlist}

\icmlaffiliation{thu}{Tsinghua University}
\icmlaffiliation{sdu}{Shandong University}

\icmlcorrespondingauthor{Zhijie Deng}{dengzhijiethu@gmail.com}
\icmlcorrespondingauthor{Jun Zhu}{dcszj@tsinghua.edu.cn}

\icmlkeywords{Machine Learning, ICML}

\vskip 0.3in
]



\printAffiliationsAndNotice{}  % leave blank if no need to mention equal contribution

\begin{abstract}
Deep Ensemble (DE) is an effective alternative to Bayesian neural networks for uncertainty quantification in deep learning. The uncertainty of DE is usually conveyed by the functional inconsistency among the ensemble members, say, the disagreement among their predictions. Yet, the functional inconsistency stems from unmanageable randomness and may easily collapse in specific cases. To render the uncertainty of DE reliable, we propose a refinement of DE where the functional inconsistency is explicitly characterized, and further tuned w.r.t. the training data and certain priori beliefs. Specifically, we describe the functional inconsistency with the empirical covariance of the functions dictated by ensemble members, which, along with the mean, define a Gaussian process (GP). Then, with specific priori uncertainty imposed, we maximize functional evidence lower bound to make the GP specified by DE approximate the Bayesian posterior. In this way, we relate DE to Bayesian inference to enjoy reliable Bayesian uncertainty. Moreover, we provide strategies to make the training efficient. Our approach consumes only marginally added training cost than the standard DE, but achieves better uncertainty quantification than DE and its variants across diverse scenarios.
\end{abstract}
% \vspace{-2.5ex}
\section{Introduction}
% \vspace{-.5ex}
Bayesian treatment of deep neural networks (DNNs) is promised to enjoy principled Bayesian uncertainty while unleashing the capacity of DNNs, with {Bayesian neural networks} (BNNs) as popular examples~\citep{mackay1992practical,hinton1993keeping,neal1995bayesian,graves2011practical}.
Nevertheless, despite the surge of advance in BNNs~\citep{louizos2016structured,zhang2018noisy}, many existing BNNs still face obstacles in accurate and scalable inference~\citep{sun2018functional}, and exhibit limitations in uncertainty quantification and out-of-distribution robustness~\citep{ovadia2019can}. % when compared with \emph{Deep Ensemble}~\citep{lakshminarayanan2017simple}.

\begin{figure}[t]
% \vspace{ex}
\centering
\begin{minipage}{\linewidth}
% \begin{figure}
\centering
\rotatebox[origin=l]{90}{\makebox[0.45in]{\scriptsize DE}}%
% \rotatebox[origin=c]{90}{\makebox[1in]{DE}}%
    \begin{subfigure}[b]{0.22\linewidth}
    \centering
    \includegraphics[width=\linewidth]{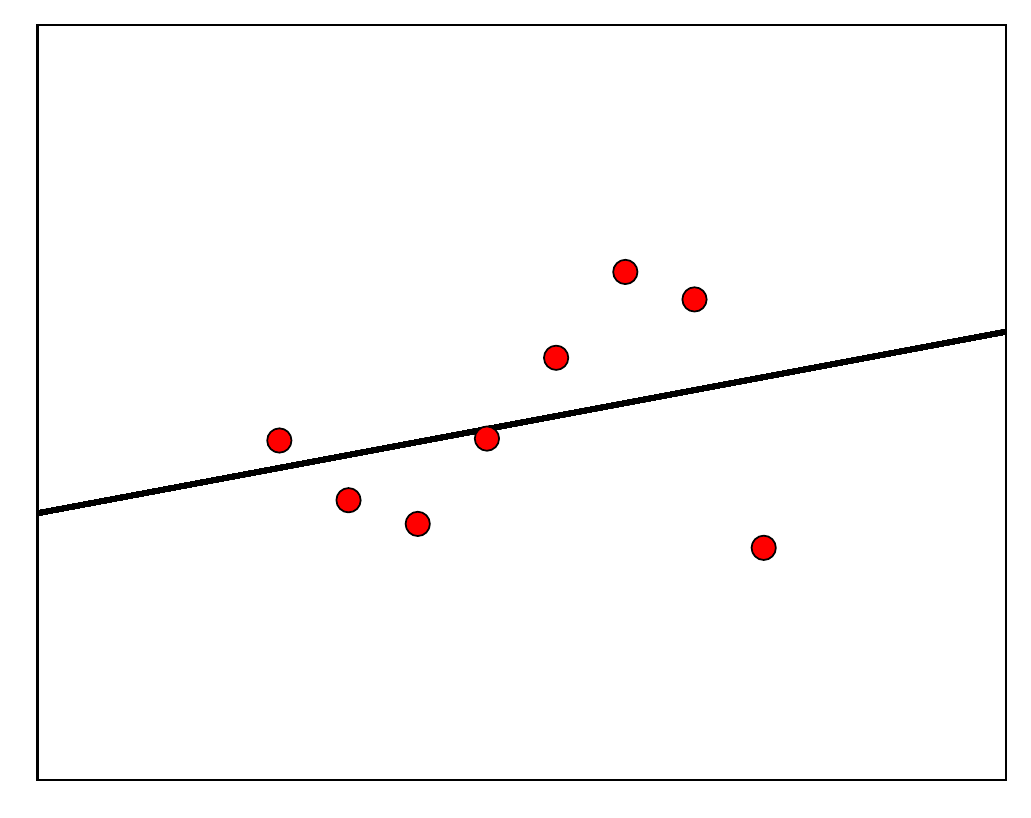}
        \vspace{-2.5ex}
        % \caption{\scriptsize }
        % \label{fig:ood-cifar}
    \end{subfigure}
    \begin{subfigure}[b]{0.22\linewidth}
    \centering
    \includegraphics[width=\linewidth]{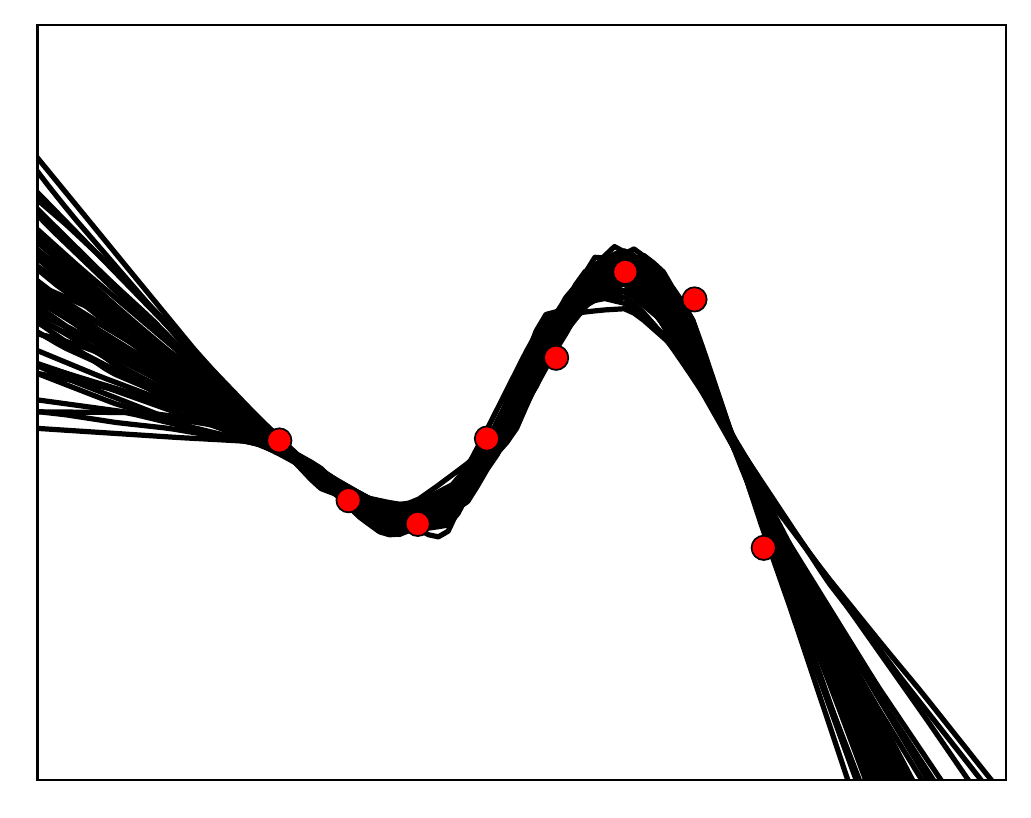}
        \vspace{-2.5ex}
        % \caption{\scriptsize }
        % \label{fig:ood-imagenet}
    \end{subfigure}
    \begin{subfigure}[b]{0.22\linewidth}
    \centering
        \includegraphics[width=\textwidth]{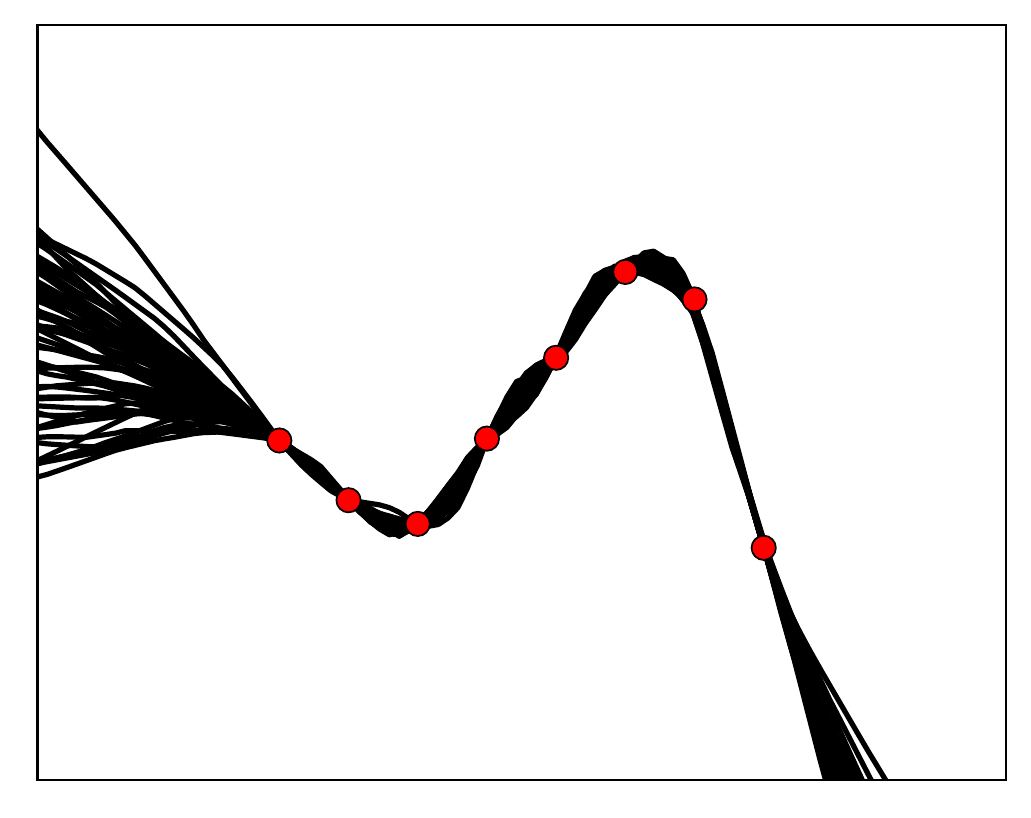}
        \vspace{-2.5ex}
        % \caption{\scriptsize }
        % \label{fig:ood-reg-cifar}
    \end{subfigure}
     \begin{subfigure}[b]{0.22\linewidth}
    \centering
        \includegraphics[width=\textwidth]{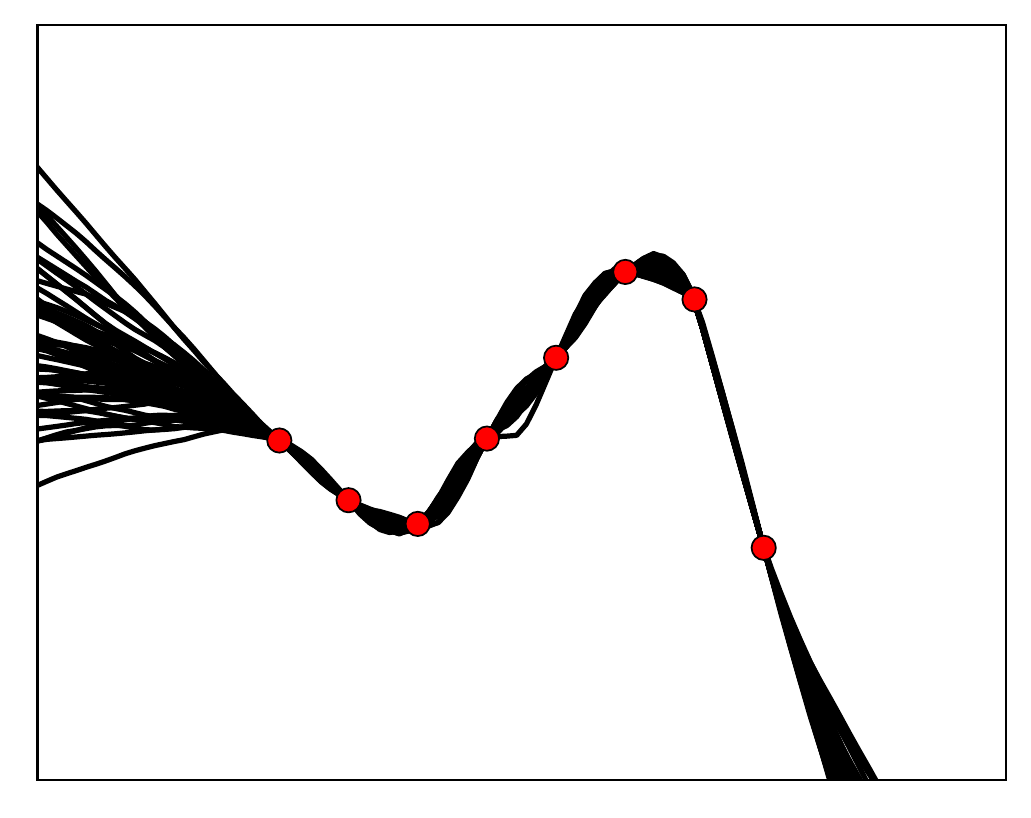}
        \vspace{-2.5ex}
        % \caption{\scriptsize }
        % \label{fig:ood-reg-cifar}
    \end{subfigure}
\end{minipage}
\begin{minipage}{\linewidth}
% \begin{figure}
\centering
\rotatebox[origin=l]{90}{\makebox[0.2in]{\scriptsize rDE}}%
    \begin{subfigure}[b]{0.22\linewidth}
    \centering
    \includegraphics[width=\linewidth]{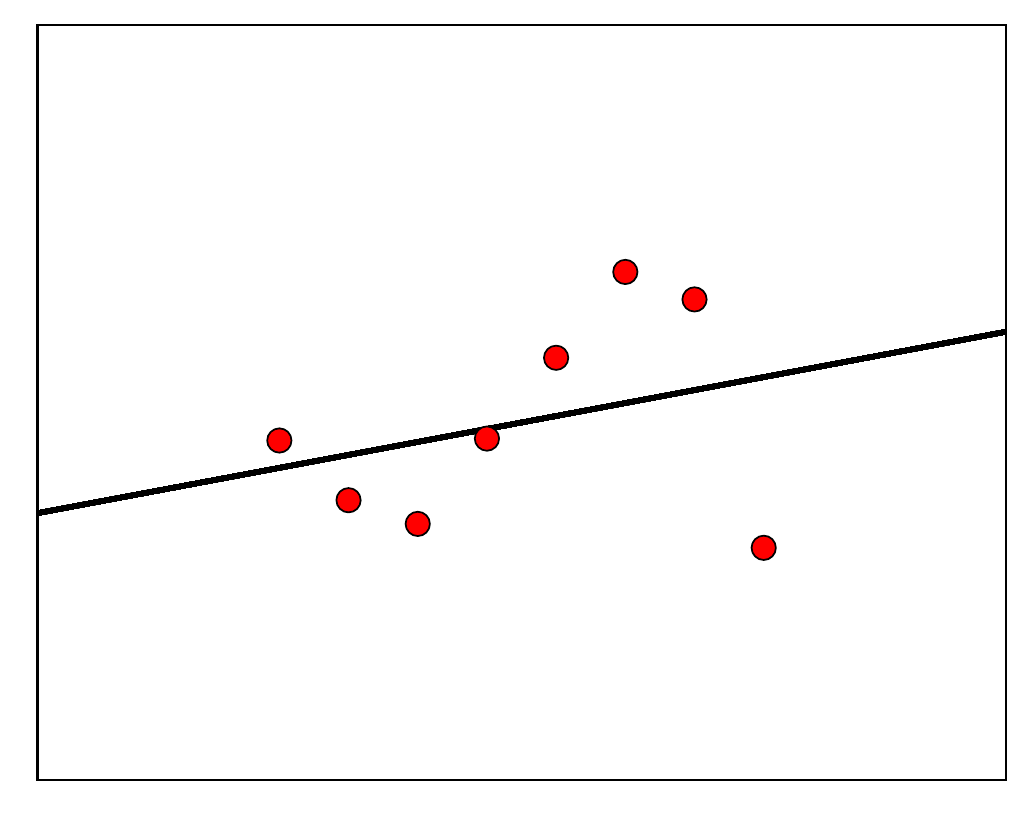}
        \vspace{-5ex}
        % \caption{\scriptsize }
        % \label{fig:ood-cifar}
    \end{subfigure}
    \begin{subfigure}[b]{0.22\linewidth}
    \centering
    \includegraphics[width=\linewidth]{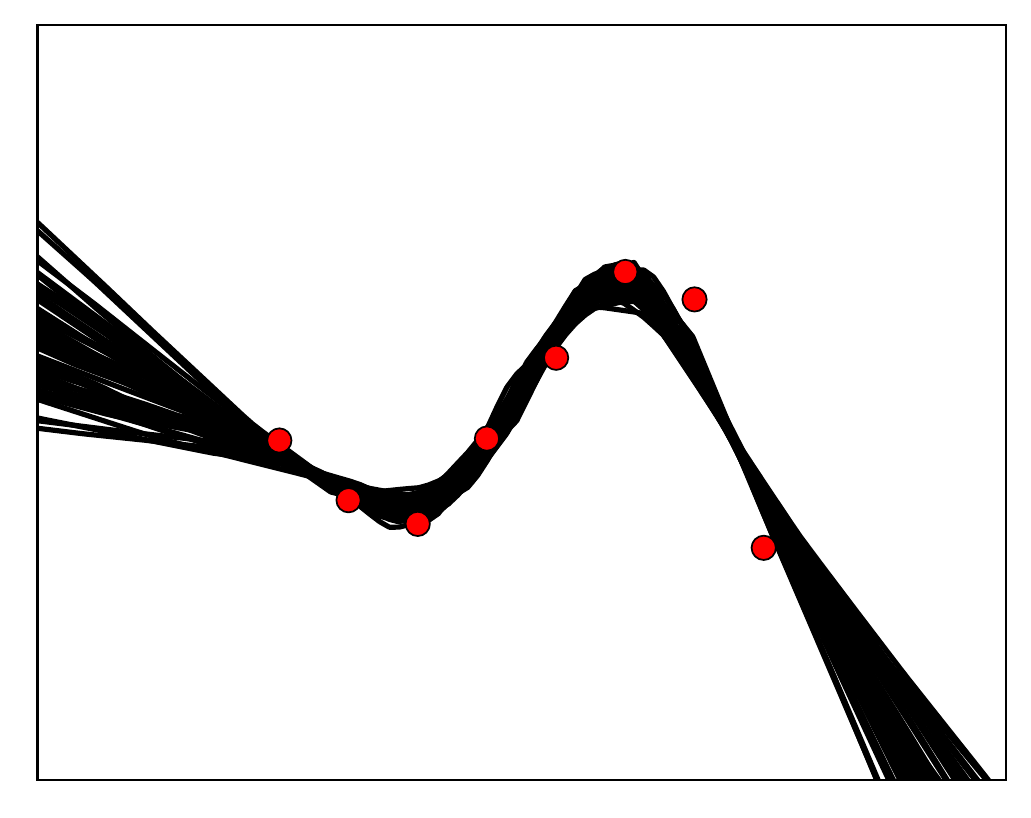}
        \vspace{-5ex}
        % \caption{\scriptsize }
        % \label{fig:ood-imagenet}
    \end{subfigure}
    \begin{subfigure}[b]{0.22\linewidth}
    \centering
        \includegraphics[width=\textwidth]{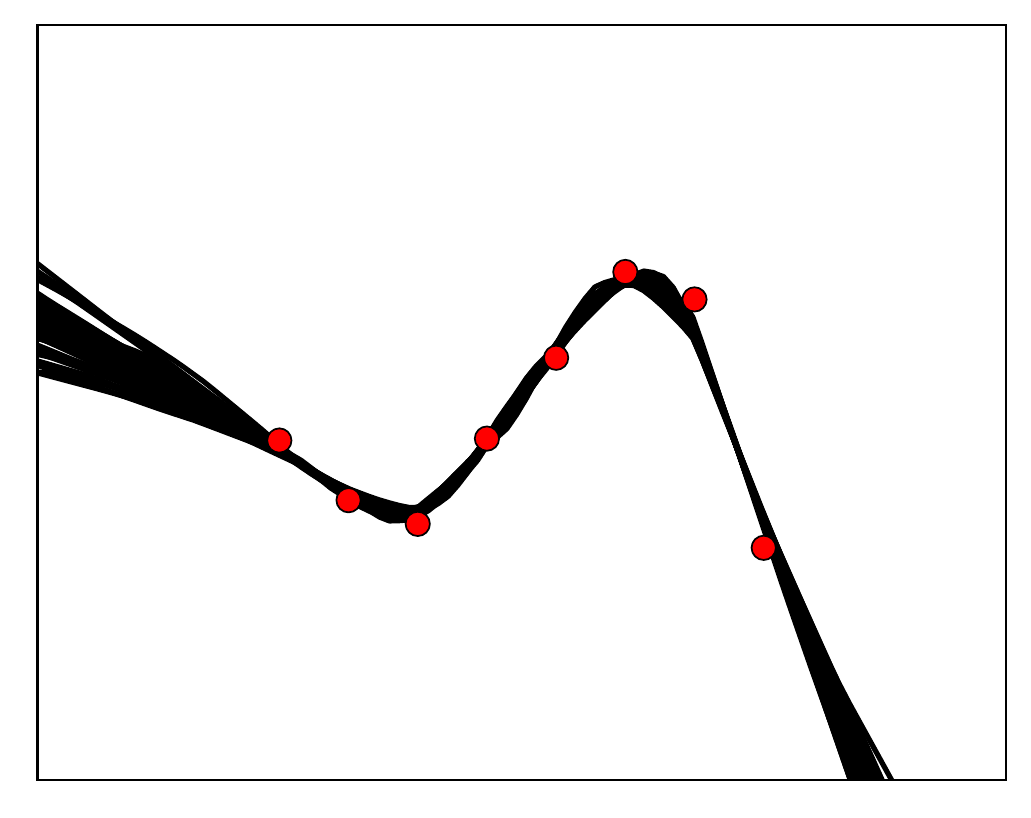}
        \vspace{-5ex}
        % \caption{\scriptsize }
        % \label{fig:ood-reg-cifar}
    \end{subfigure}
     \begin{subfigure}[b]{0.22\linewidth}
    \centering
        \includegraphics[width=\textwidth]{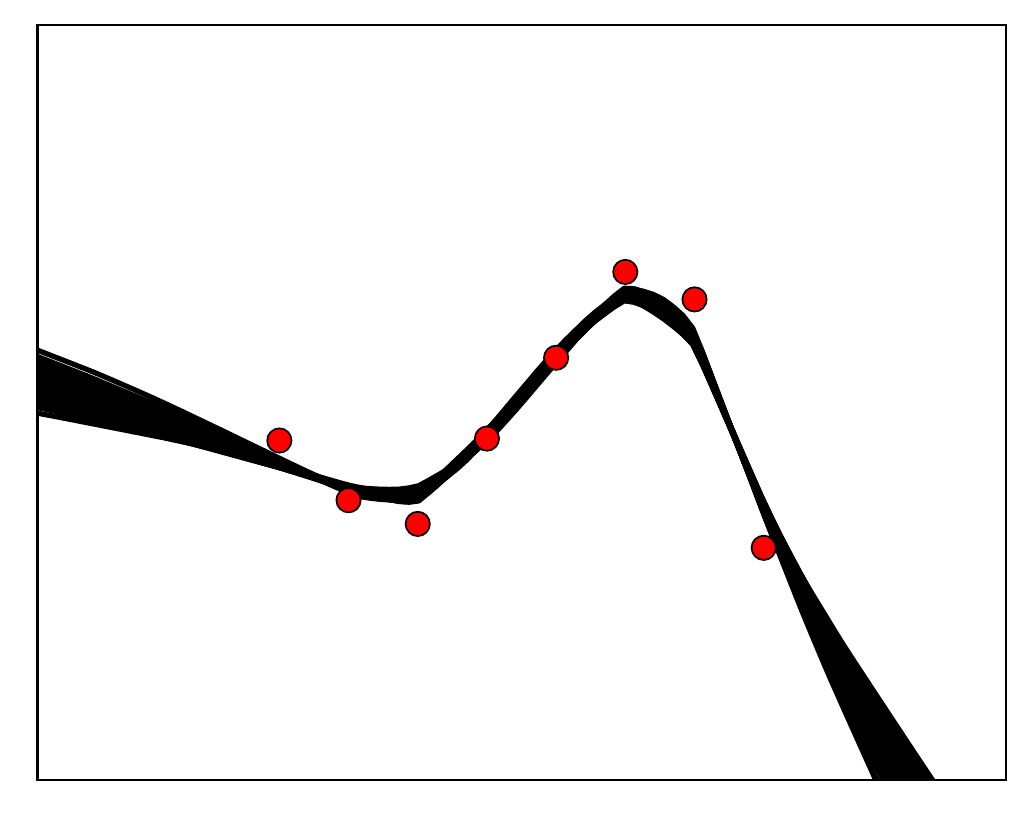}
        \vspace{-5ex}
        % \caption{\scriptsize }
        % \label{fig:ood-reg-cifar}
    \end{subfigure}
\end{minipage}
\begin{minipage}{\linewidth}
% \begin{figure}
\centering
\rotatebox[origin=l]{90}{\makebox[0.85in]{\scriptsize NN-GP}}%
\begin{subfigure}[b]{0.22\linewidth}
    \centering
    \includegraphics[width=\linewidth]{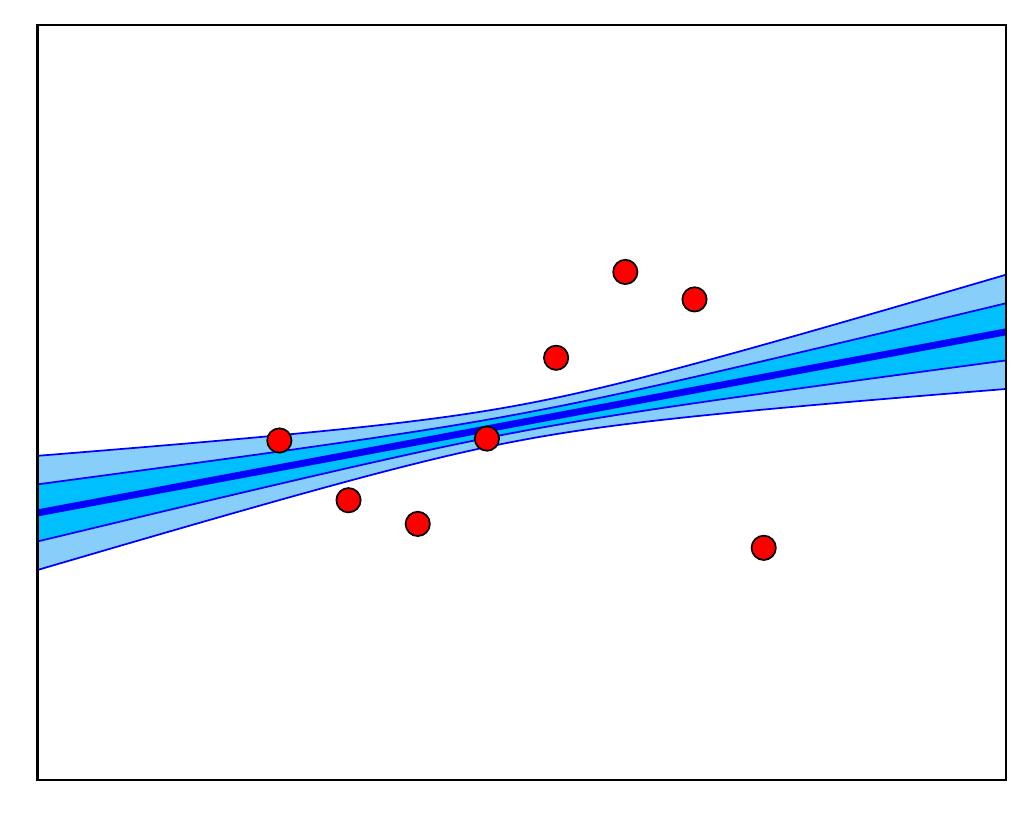}
        \vspace{-3.5ex}
        \caption*{\scriptsize 0 hidden layer}
        % \label{fig:ood-cifar}
    \end{subfigure}
    \begin{subfigure}[b]{0.22\linewidth}
    \centering
    \includegraphics[width=\linewidth]{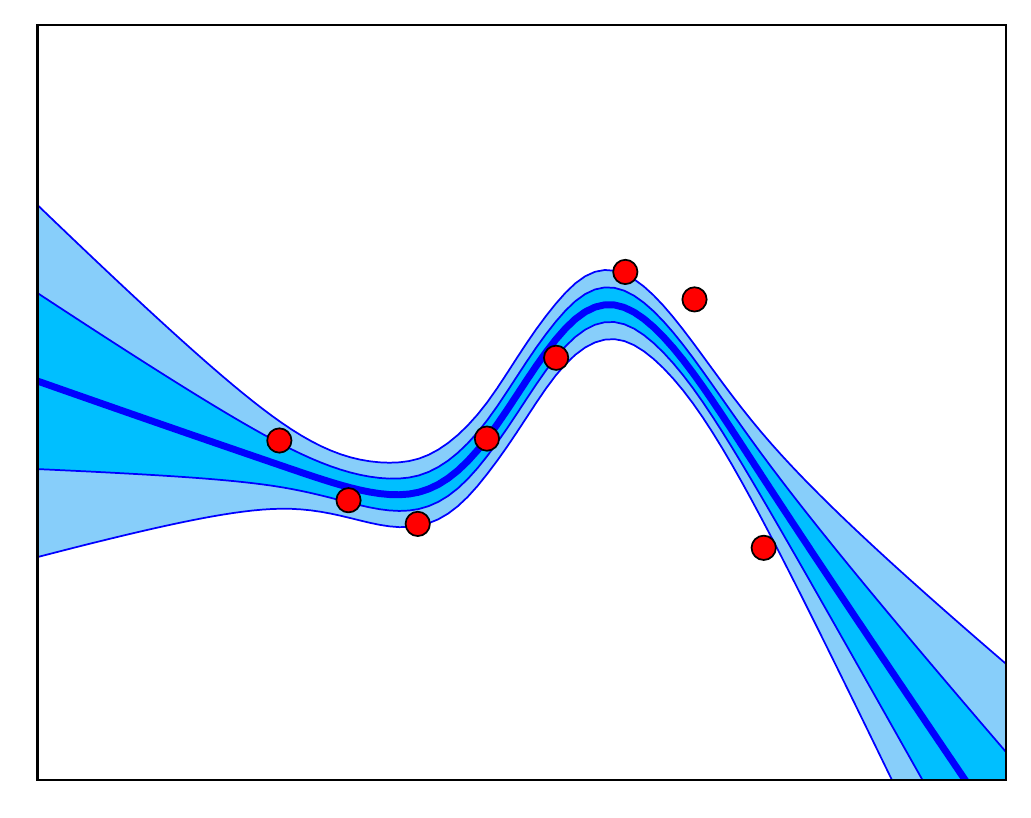}
        \vspace{-3.5ex}
        \caption*{\scriptsize 1 hidden layer}
        % \label{fig:ood-imagenet}
    \end{subfigure}
    \begin{subfigure}[b]{0.22\linewidth}
    \centering
        \includegraphics[width=\textwidth]{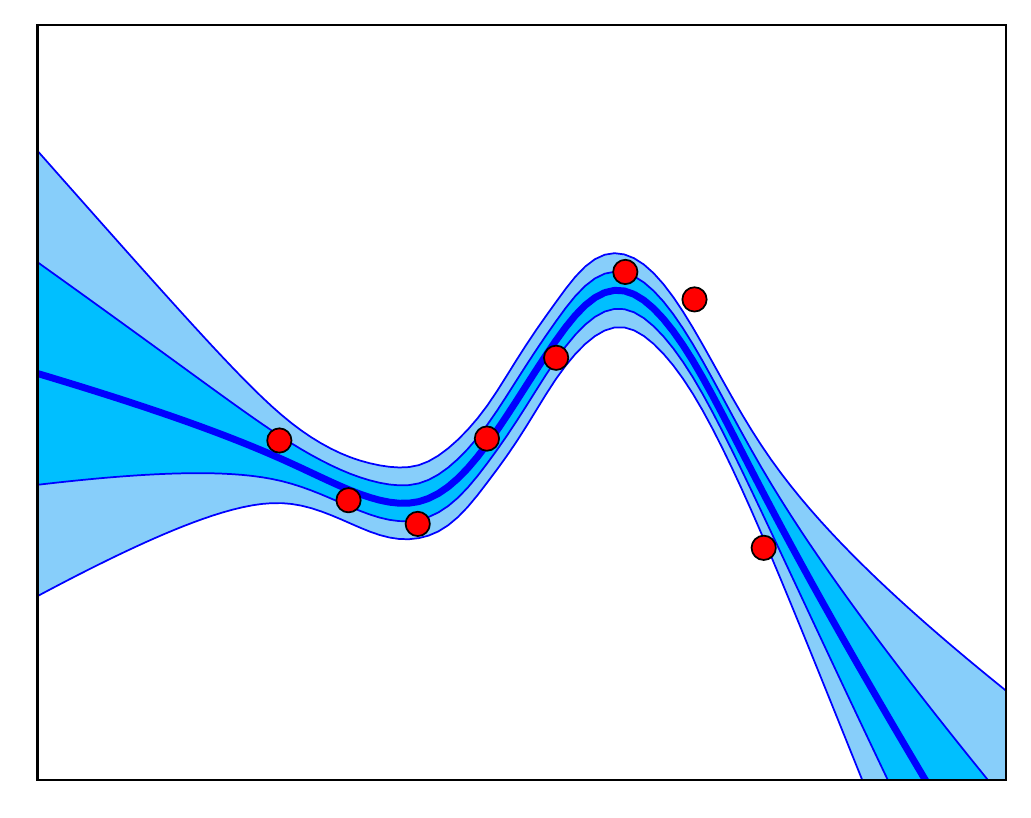}
        \vspace{-3.5ex}
        \caption*{\scriptsize 2 hidden layers}
        % \label{fig:ood-reg-cifar}
    \end{subfigure}
     \begin{subfigure}[b]{0.22\linewidth}
    \centering
        \includegraphics[width=\textwidth]{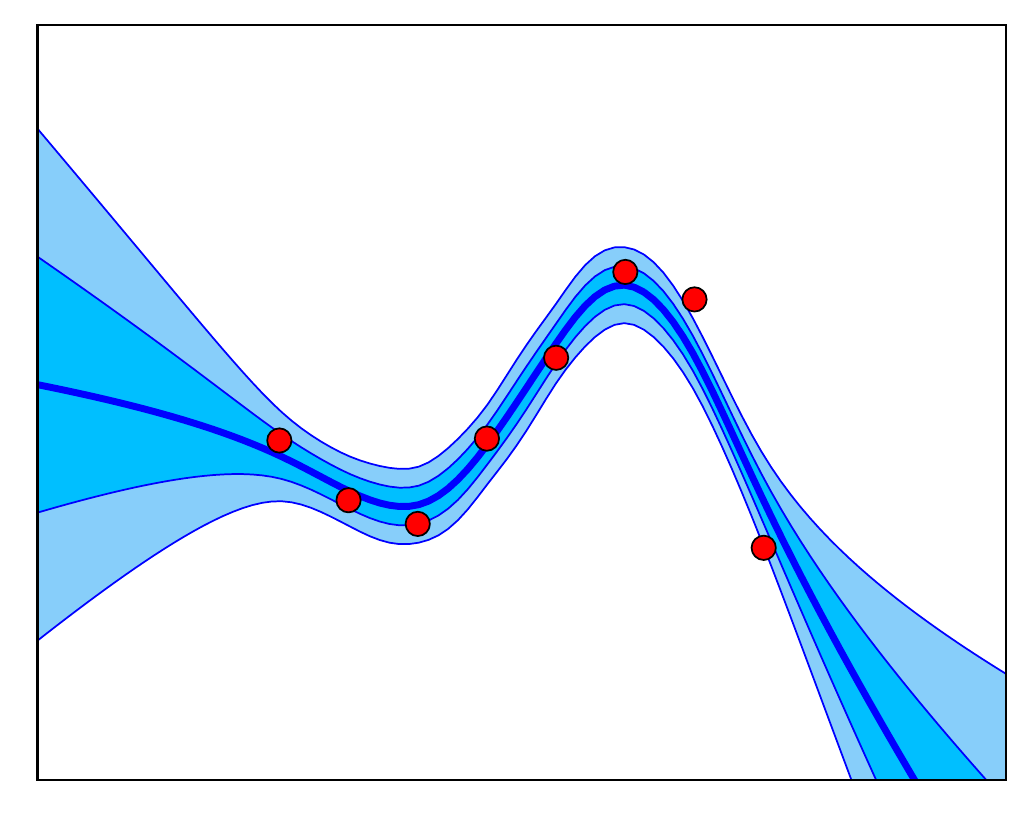}
        \vspace{-3.5ex}
        \caption*{\scriptsize 3 hidden layers}
        % \label{fig:ood-reg-cifar}
    \end{subfigure}
\end{minipage}
\vspace{-1.5ex}
\caption{\footnotesize 1-D regression on $y=\sin 2x + \epsilon, \epsilon \sim \mathcal{N}(0, 0.2)$. 
Red dots refer to the training data (the rightest dot is deliberately perturbed to cause significant data noise). 
Rows: DE, rDE and NN-GP (the gold standard). 
Columns: using multilayer perceptrons (MLPs) architectures with 0 hidden layer, 1 hidden layer of size 64, 2 hidden layers of size 128, and 3 hidden layers of size 256. 
DE and rDE use $50$ MLPs for ensemble. 
The weight-space priors for rDE and NN-GP are Gaussian distributions. 
Black lines for DE and rDE refer to the predictions of the ensemble members.
Dark blue curves and shaded regions for DE-GP refer to mean predictions and uncertainty. 
Compared to NN-GP, DE and rDE suffer from over-confidence and less calibrated uncertainty estimates.}
\label{fig:illu}
\vspace{-2ex}
\end{figure}

Alternatively, {Deep Ensemble} (DE)~\citep{lakshminarayanan2017simple} trains multiple independent, randomly-initialized DNNs for ensemble, presenting higher flexibility and effectiveness than BNNs.
% with prediction disagreement as uncertainty.
% However, the “uncertainty” yielded by DE totally stems from the randomness in DNNs initialization and stochastic gradient descent (SGD), distinct from the Bayesian uncertainty . 
% So, there is no guarantee that the uncertainty estimates given by DE are reliable.
Practitioners tend to interpret the \emph{functional inconsistency} among the ensemble members, say, the disagreement among their predictions, as a proxy of DE's uncertainty. %\jianfei{actually this sounds worse than functional inconsistency. Maybe add a sentence before to briefly introduce what is functional inconsistency } 
However, the functional inconsistency stems from the unmanageable randomness in DNN initialization and stochastic gradient descent (SGD), thus is likely to collapse in specific cases (see \cref{fig:illu}). 
% However, it is hard to interpret DE as a Bayesian approach which seeks for the Bayesian posterior of a certain model, and there is no guarantee that the uncertainty estimates given by DE are reliable. 
To fix this issue, recent works like RMS~\citep{lu2017ensemble,osband2018randomized,pearce2020uncertainty} and NTKGP~\citep{NEURIPS2020_0b1ec366} refine DE to be a Bayesian inference approach in the spirit of ``{sample-then-optimize}''~\citep{matthews2017sample}. %, %e.g.,\junz{these are examples of "recent works"? may put them just after "recent works", or make the connection more explicit.} {Randomised maximum a posteriori (MAP) Sampling} (RMS)~\citep{lu2017ensemble,osband2018randomized,pearce2020uncertainty}, and {Neural Tangent Kernel Gaussian process} (NTK-GP)~\citep{NEURIPS2020_0b1ec366}, etc. 
Yet, these works often make strong assumptions like Gaussian likelihoods and {linearized}/{infinite-width} models.%, having difficulties to generalize.

% In this work, we endow DE with a principled learning criterion under the function-space Bayesian inference paradigm.
% refurbish DE as a Bayesian approach from a function-space viewpoint. 
% It is trivial to regard DE as an approximate posterior in the form of mixture of deltas (MoD). 
% However, the singularity issue of MoD can yield ill-defined objectives for Bayesian learning~\citep{hron2018variational}.
This work aims at tackling these issues. %addressing the unreliability issue of the uncertainty of DE without relying on restrictive assumptions. 
We first reveal that the unreliable uncertainty of DE arises from the gap that the functional inconsistency are not properly adapted w.r.t. the training data and certain priori beliefs, but is leveraged to quantify post data uncertainty in the test phase. 
To bridge the gap, we propose to explicitly incorporate the functional inconsistency into modeling. 
Viewing the ensemble members as a set of basis functions, the functional inconsistency can be formally described by their empirical covariance, which, along with their mean, specify a \emph{Gaussian process} (GP). 
Such a GP is referred to as DE-GP for short hereinafter. 
Given this setup, we advocate tuning the whole DE-GP, including its mean and covariance, w.r.t. the training data under the functional variational inference (fVI) paradigm~\citep{sun2018functional} given specific priors. 
Namely, we regard DE-GP as an parametric approximate posterior so as to enjoy the principled Bayesian uncertainty. 
By fVI, DE-GP can handle classification problems directly and exactly, without casting them into regression ones as done by some existing Bayesian variants of DE~\citep{NEURIPS2020_0b1ec366}.

% In the perspective of posterior approximation, organizing DE as DE-GP actually shifts the family of the approximate posteriors from mixture of deltas, which have degenerated supports and often lead to ill-defined objectives for approximate inference~\citep{hron2018variational}, to amenable Gaussian processes. %\jianfei{what's the benefit of doing so}

% DE-GP is parametric, so we train it to approach the true posterior over functions under the principle of functional variational inference (fVI). 
% The learning objective is  {functional evidence lower bound} (fELBO)~\citep{sun2018functional,rudner2021rethinking}. 
% We leverage the theoretical results of \citep{sun2018functional} that the KL divergence between stochastic processes can be framed into that between the marginal distributions of function evaluations.\junz{this sentence can be removed to avoid distraction from our contributions.}
Without loss of generality, we adopt a prior also in the form of GP, then the gradients of the KL divergence between the DE-GP approximate posterior and the GP prior involved in fVI can be easily estimated without reliance on complicated gradient estimators~\cite{shi2018spectral}. 
We provide recipes to make the training even faster, then the additional computation overhead induced by DE-GP upon DE is minimal.
We also identify the necessity of including an extra weight-space regularization term to guarantee the generalization performance of DE-GP when using deep architectures.
Empirically, DE-GP outperforms DE and its variants on various regression datasets, and presents superior uncertainty estimates and out-of-distribution robustness without compromising accuracy in standard image classification tasks. 
DE-GP also shows promise in solving contextual bandit problems, where the uncertainty guides exploration.

\vspace{-1.5ex}
\section{Related Work}
\vspace{-.5ex}
% With the background section covering most related literature on BNNs and NN-GPs, we mainly review related work on Deep Ensemble and its Bayesian refinement here.

\textbf{Bayesian neural networks.} 
Bayesian treatment of DNNs is an emerging topic yet with a long history~\citep{mackay1992bayesian,hinton1993keeping,neal1995bayesian,graves2011practical}.
BNNs can be learned by variational inference~\citep{blundell2015weight,hernandez2015probabilistic,louizos2016structured,zhang2018noisy,khan2018fast,deng2020bayesadapter}, Laplace approximation~\citep{mackay1992bayesian,ritter2018scalable}, Markov chain Monte Carlo~\citep{welling2011bayesian,chen2014stochastic,zhang2019cyclical}, particle-optimization based variational inference~\citep{liu2016stein}, Monte Carlo dropout~\citep{gal2016dropout}, and other methods~\citep{maddox2019simple}.
To avoid the difficulties of posterior inference in weight space, some recent works advocate performing Bayesian inference in function space~\citep{sun2018functional,rudner2021rethinking,wang2018function}. 
In function space, BNNs of infinite or even finite width equal to GPs~\citep{neal1996priors,lee2018deep,novak2018bayesian,khan2019approximate}, which provides supports for constructing an approximate posterior in the form of GP.

\textbf{Deep Ensemble.} As an alternative to BNNs, DE~\cite{lakshminarayanan2017simple} has shown promise in diverse uncertainty quantification scenarios~\citep{ovadia2019can}, yet lacks a proper Bayesian interpretation. 
\citet{NEURIPS2020_322f6246} interpreted DE as a method that approximates the Bayesian posterior predictive, but it is hard to judge whether the approximation is reliable or not in practice.
The notion of ``sample-then-optimize''~\citep{matthews2017sample} has also been considered to make DE Bayesian. 
For example, RMS~\citep{lu2017ensemble,osband2018randomized,pearce2020uncertainty} regularizes the ensemble members towards randomised priors to obtain posterior samples, while it typically assumes \emph{linear} data likelihood which is impractical for deep models and classification tasks.
\citet{NEURIPS2020_0b1ec366} proposed to add a randomised function to each ensemble member to realize a function-space Bayesian interpretation, but the method is asymptotically exact in the \emph{infinite width} limit and is limited to regressions.
By contrast, DE-GP works without restrictive assumptions. 
In parallel, \citet{d2021repulsive} proposed to add a repulsive term to DE but the method relies on less scalable gradient estimators.

\begin{figure*}[t]
% \vspace{-1ex}
\centering
\begin{subfigure}[b]{0.19\linewidth}
    \centering
    \includegraphics[width=\linewidth]{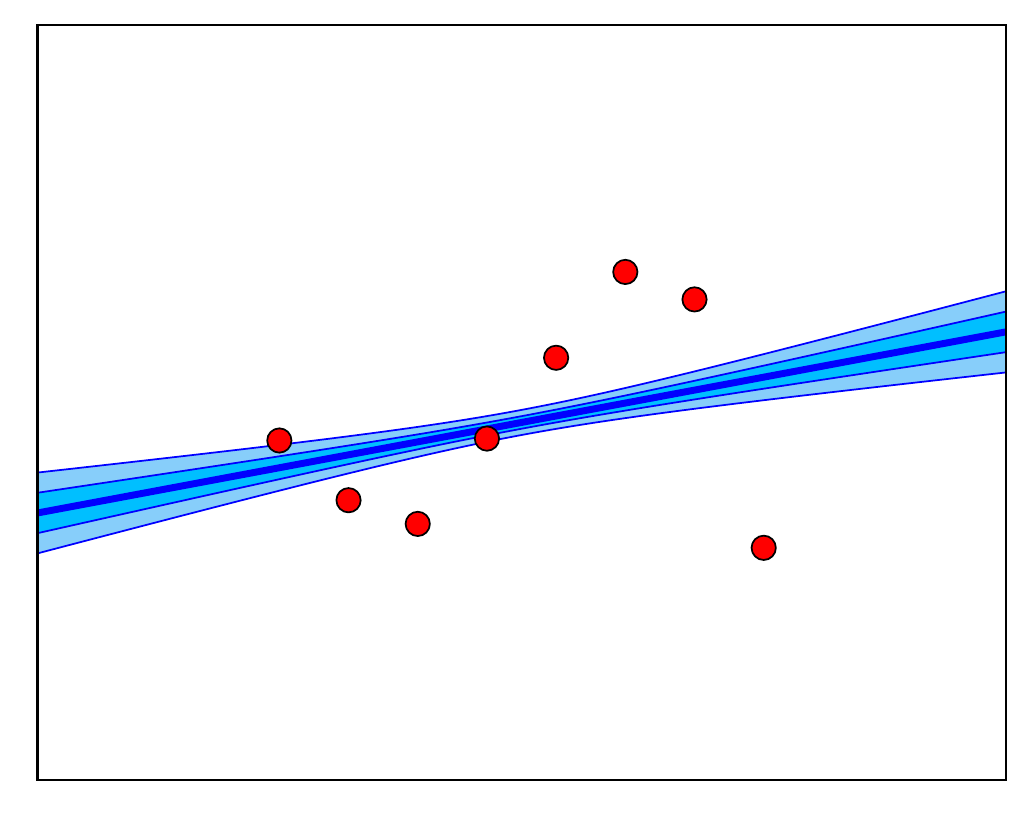}
        \vspace{-4.ex}
       \caption*{\scriptsize 0 hidden layer}
        % \label{fig:ood-cifar}
    \end{subfigure}
    \begin{subfigure}[b]{0.19\linewidth}
    \centering
    \includegraphics[width=\linewidth]{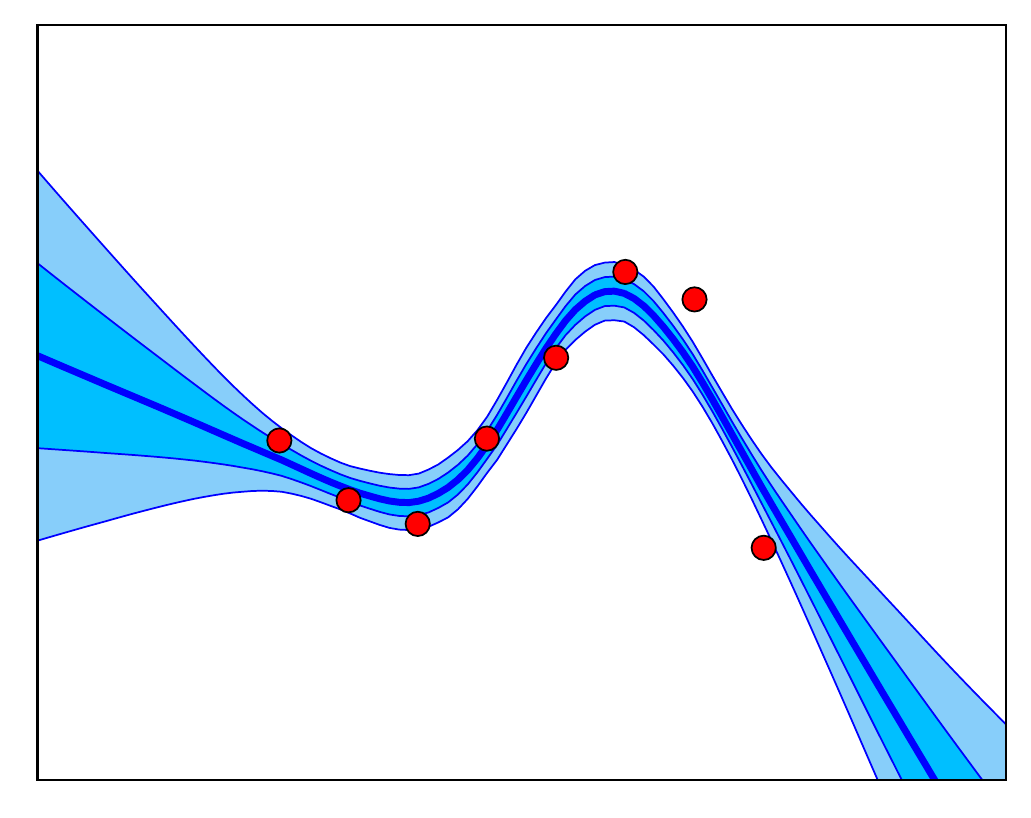}
        \vspace{-4ex}
       \caption*{\scriptsize 1 hidden layer}
        % \label{fig:ood-imagenet}
    \end{subfigure}
    \begin{subfigure}[b]{0.19\linewidth}
    \centering
       \includegraphics[width=\textwidth]{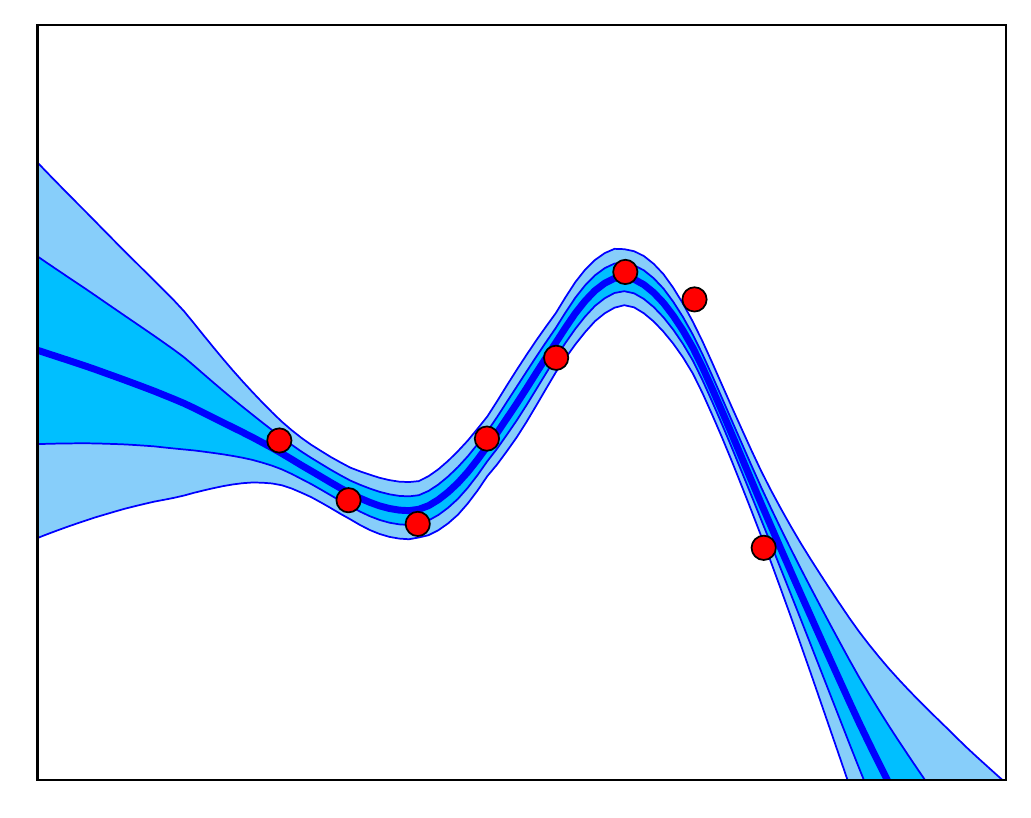}
        \vspace{-4.ex}
       \caption*{\scriptsize 2 hidden layers}
        % \label{fig:ood-reg-cifar}
    \end{subfigure}
    \begin{subfigure}[b]{0.19\linewidth}
    \centering
       \includegraphics[width=\textwidth]{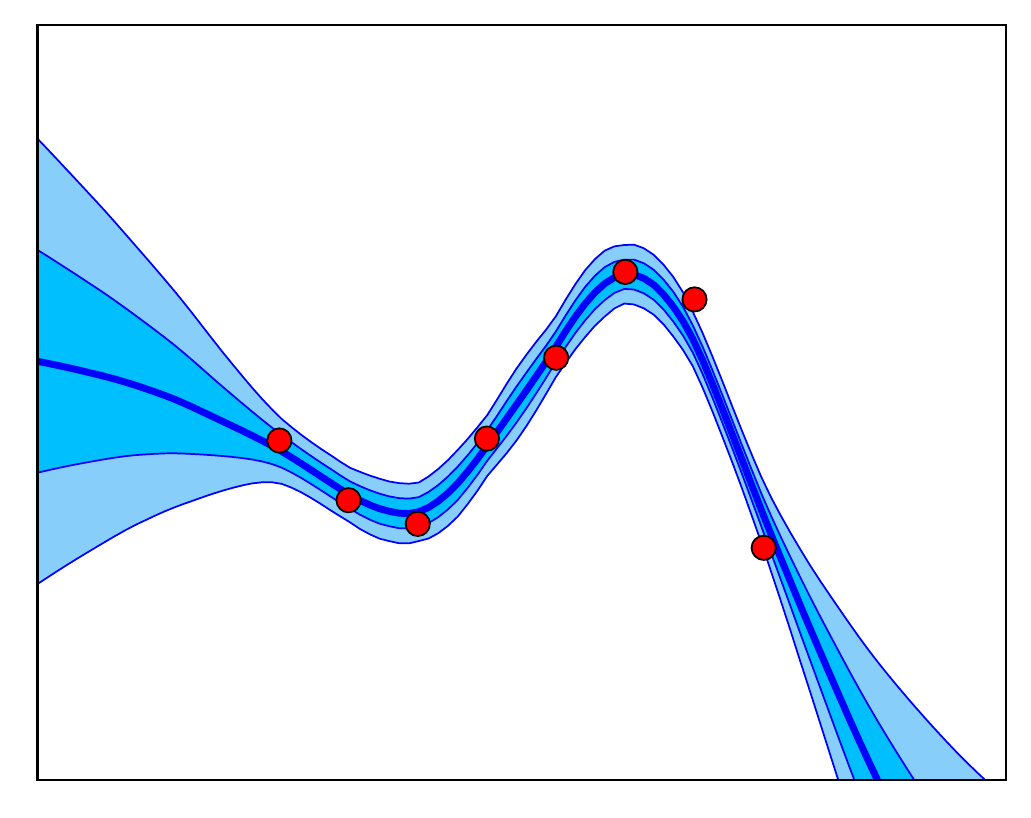}
        \vspace{-4.ex}
       \caption*{\scriptsize 3 hidden layers}
        % \label{fig:ood-reg-cifar}
    \end{subfigure}
    \begin{subfigure}[b]{0.19\linewidth}
    \centering
       \includegraphics[width=\textwidth]{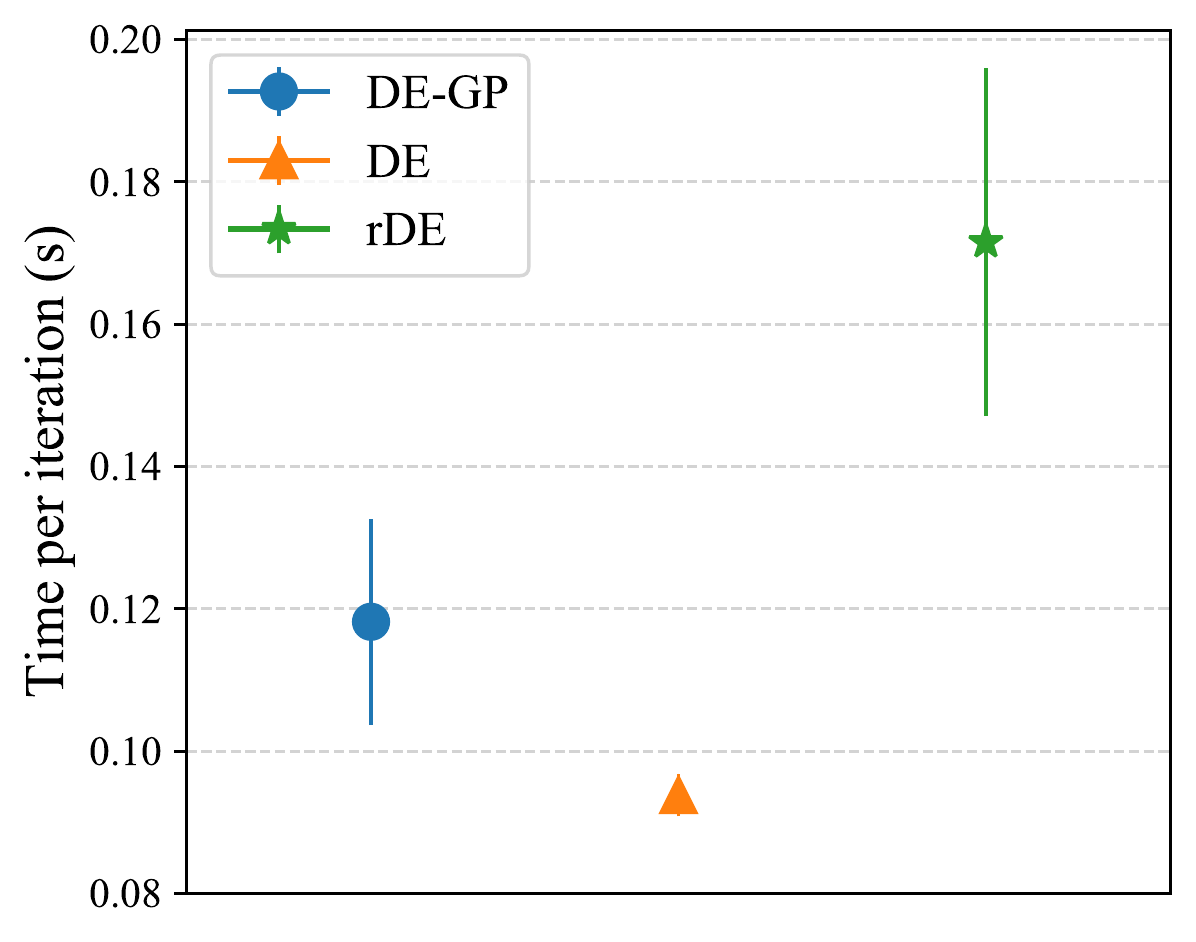}
        \vspace{-4.ex}
       \caption*{\scriptsize Training speed comparison}
        % \label{fig:ood-reg-cifar}
    \end{subfigure}
\vspace{-2ex}
\caption{\footnotesize 
DE-GP behaves similarly to the non-parameteric NN-GP (the gold standard), with only marginally added overheads upon DE.
% Results of DE-GP on $y=\sin 2x +\epsilon, \epsilon \sim \mathcal{N}(0, 0.2)$. 
The settings are identical to those in \cref{fig:illu}. 
The training speed is measured with the MLP architecture with 3 hidden layers of size 256.}
\label{fig:illu2}
\vspace{-.0ex}
\end{figure*}

% \vspace{-1ex}
\section{Motivation}
\vspace{-.5ex}
Assuming access to a dataset $\mathcal{D}=(\mathbf{X}, \mathbf{Y})=\{(\vx_i, \vy_i)\}_{i=1}^n$, with $\vx_i \in \mathcal{X}$ as data and $\vy_i$ as $C$-dimensional targets, 
we can use a DNN $g(\cdot, \vw): \mathcal{X} \rightarrow \mathbb{R}^C$ with weights $\vw$ to fit it. 
Despite impressive performance, the regularly trained DNNs are prone to over-confidence, making it hard to decide how certain they are about the predictions. 
Lacking the ability to reliably quantify predictive uncertainty is unacceptable for realistic decision-making scenarios.% in finance, medicine, and self-driving cars. 

A principled mechanism for uncertainty quantification in deep learning is to apply Bayesian treatment to DNNs to reason about Bayesian uncertainty. 
The resulting models are known as BNNs.
In BNNs, $\vw$ is treated as a random variable.
Given some prior beliefs $p(\vw)$, we chase the posterior $p(\vw|\mathcal{D})$. 
In practice, it is intractable to analytically compute the true posterior $p(\vw|\mathcal{D})$ due to the high non-linearity of DNNs, so some approximate posterior $q(\vw)$ is usually found by techniques like variational inference~\cite{blundell2015weight}, Laplace approximation~\cite{mackay1992bayesian}, Monte Carlo dropout~\citep{gal2016dropout}, etc.

BNNs marginalize over the posterior to predict for new data $\vx^*$ (a.k.a. the \emph{posterior predictive}):
\begin{equation}
\label{eq:post-pred}
\small
\begin{aligned}
    p(y|\vx^*, \mathcal{D}) & = \mathbb{E}_{p(\vw|\mathcal{D})} p(y|\vx^*, \vw)  \\
    & \approx \mathbb{E}_{q(\vw)} p(y|\vx^*, \vw) \approx \frac{1}{S} \sum_{s=1}^S p(y|\vx^*, \vw_s),
\end{aligned}
\end{equation}
where $\vw_s\sim q(\vw), s=1,...,S$. 
This procedure propagates the embedded model uncertainty into the prediction. 
However, most of the existing BNN approaches face obstacles in precise posterior inference due to non-trivial and convoluted posterior dependencies~\citep{louizos2016structured,zhang2018noisy,shi2018kernel,sun2018functional}, and deliver unsatisfactory uncertainty estimates and out-of-distribution (OOD) robustness~\citep{ovadia2019can}.

As a workaround of BNNs, Deep Ensemble (DE)~\citep{lakshminarayanan2017simple} deploys a set of $M$ DNNs $\{g(\cdot, \vw_i)\}_{i=1}^M$ to interpret the data from different angles. In this sense, we can view DE as an approximate posterior in the form of a mixture of deltas (MoD) $q(\vw)=\frac{1}{M}\sum_{i=1}^M \delta (\vw - \vw_i)$. 
Nevertheless, instead of being tuned under Bayesian inference principles, the ensemble members are \emph{independently} trained under deterministic learning principles like maximum likelihood estimation (MLE) and maximum a posteriori (MAP) (we refer to the resulting models as DE and regularized DE (rDE) respectively): 
% \vspace{-0.5ex}
\begin{equation}
% \vspace{-1ex}
\label{eq:de}
\small
\begin{aligned}
\max_{\vw_1, ..., \vw_M}\mathcal{L}_\text{DE} =&\max_{\vw_1, ..., \vw_M} \frac{1}{M}\sum_{i=1}^M \log p(\mathcal{D}|\vw_i),\\
\max_{\vw_1, ..., \vw_M}\mathcal{L}_\text{rDE} =&\max_{\vw_1, ..., \vw_M} \frac{1}{M}\sum_{i=1}^M [\log p(\mathcal{D}|\vw_i) + \log p(\vw_i)].
\end{aligned}
\end{equation}
The randomness in initialization and SGD diversifies the ensemble members, making them explore distinct modes of the non-convex loss landscape of DNNs~\citep{fort2019deep,NEURIPS2020_322f6246}, and in turn boost the ensemble performance. 
In parallel, researchers have found that DE is also promising in quantifying uncertainty~\citep{lakshminarayanan2017simple,ovadia2019can}.
The \emph{functional inconsistency} among the ensemble members is usually interpreted as a proxy of DE's uncertainty. 
However, we question the efficacy of the uncertainty of DE given that the functional inconsistency arises from the randomness in DNN initialization and SGD instead of a Bayesian treatment, i.e., the true model uncertainty is not explicitly considered during training. 
To confirm this apprehension, we evaluate DE and rDE on a simple 1-D regression problem. 
We choose neural network Gaussian process (NN-GPs)~\cite{neal1996priors} as a golden standard because it equals to {BNNs} in the infinite-width limit and allows for analytical function-space posterior inference. 
We depict the results in \cref{fig:illu}.

The results echo our concerns on DE. It is evident that (\RN{1}) DE and rDE collapse to a single model in the linear case (the first column), attributed to that the loss surface is convex w.r.t. the model parameters; (\RN{2}) DE and rDE reveal minimal uncertainty in in-distribution regions, although there is severe data noise; (\RN{3}) the uncertainty of DE and rDE deteriorates as the model size increases regardless of whether there is over-fitting. % \zhijie{don't know how to explain this? any ideas?} %  when there is data noise.\jianfei{how do the symptoms cause by the reason}
% These phenomena are unexpected. 

In essence, the functional inconsistency in DE has not been properly adapted w.r.t. the priori uncertainty and training data, but is used to quantify post data uncertainty. 
Such a gap is thought of as the cause of the unreliability issue of DE's uncertainty.
Having identified this, we propose to incorporate the functional inconsistency into modeling, and perform Bayesian inference to tune the whole model to enjoy principled Bayesian uncertainty. 
We describe how to realize this in the following.

% So, a proper Bayesian exposition for DE is still a paramount need.

% \vspace{-1.ex}
\section{Methodology}
% \vspace{-.5ex}
This section provides the details for the modeling, inference, and training procedure of the proposed DE-GP. 
We impress the readers in advance by the results of DE-GP on the aforementioned regression problem shown in \cref{fig:illu2}. 
% We describe the modeling, inference, and algorithm for the proposed DE-GP in

% \vspace{-.5ex}
\subsection{Modeling}
% \vspace{-.5ex}
% In fact, an ensemble $\{g(\cdot, \vw_i)\}_{i=1}^M$ implicitly defines an approximate posterior in the form of MoD. 
% However, MoD typically leads to ill-defined objectives for approximate inference~\citep{hron2018variational}. %, so we cannot trivially push it towards the true posterior over functions. 
Viewing the ensemble members $\{g(\cdot, \vw_i)\}_{i=1}^M$ as a set of basis functions, the functional inconsistency among them can be formally represented by the empirical covariance of these functions in the following form:
\begin{equation}
\small
\Scale[1]{k(\vx, \vx') := \frac{1}{M}\sum_{i=1}^M \left(g_i(\vx)-m(\vx)\right) \left(g_i(\vx')-m(\vx')\right)^\top},
\end{equation}
where $g_i$ refers to $g(\cdot, \vw_i)$ and $m(\vx) := \frac{1}{M}\sum_{i=1}^M g_i(\vx)$. 
From the definition, $k$ is a matrix-valued kernel, with values in the space of $C \times C$ matrices.

Then, the incorporation of functional inconsistency amounts to building a model with $k(\vx, \vx')$. 
Naturally, the DE-GP $\mathcal{GP}(f| m(\vx), k(\vx, \vx'))$ comes into the picture. 
Nevertheless, $k(\vx, \vx')$ is of low rank, so we opt to add a small scaled identity matrix $\lambda\mathbf{I}_C$\footnote{$\mathbf{I}_C$ refers to the identity matrix of size $C \times C$.} upon $k(\vx, \vx')$ to avoid singularity. 
Unless specified otherwise, we refer to the resulting covariance kernel as $k(\vx, \vx')$ in the following.

We can train both DE-GP and DE to approximate the true Bayesian posteriors associated with some priors. In this sense, DE-GP shifts the family of the approximate posteriors from MoDs, which are singular and often lead to ill-defined learning objectives~\cite{hron2018variational}, to the amenable GPs. %\junz{make this mixture of deltas more explicit when introducing DE} \junz{can we say something like DE is a special (degenerating) case of DE-GP?}
The variations in $k(\vx, \vx')$ are confined to having up to $M-1$ rank, echoing the recent investigations showing that low-rank approximate posteriors for deep models conjoin effectiveness and \emph{efficiency}~\citep{maddox2019simple,izmailov2020subspace,dusenberry2020efficient}. 
% Note that the popular Nystr{\"o}m method~\citep{williams2001using} also uses a low-rank matrix to approximate the original kernel matrix. % confirms the effectiveness of  approximation to a troublesome large kernel matrix.
% \jianfei{when $M\rightarrow \infty$, can this recover exact Bayesian inference?}
% which is likely to be reasonable given the previous investigations~\citep{{maddox2019simple,izmailov2020subspace,dusenberry2020efficient}}.
 %, eliminating the need for intensive hyper-parameter tuning as for constant kernels. 
% By adapting to the observations, the basis functions can check  a rich variety of  assumptions without hand-tailor
% ``the rich variety that GP kernels can represent''

Akin to the kernels in \citep{wilson2016deep}, the DE-GP kernels are highly flexible, and may {automatically} discover the underlying structures of high-dimensional data without manual participation.

% \vspace{-.5ex}
\subsection{Inference}
% \vspace{-0.5ex}
\label{sec3.2}
As stated above, we take DE-GP as a parametric approximate posterior, i.e., $q(f)=\mathcal{GP}(f|m(\vx), k(\vx, \vx'))$, and push it to approach the true posterior over functions associated with specific priors by fVI~\cite{sun2018functional}.

\textbf{Prior} We can freely choose a distribution over functions (i.e., a stochastic process) as the prior attributed to the fVI paradigm. 
Without loss of generality, we use the MC estimates of NN-GPs (MC NN-GPs)~\citep{novak2018bayesian} as the prior because (\RN{1}) they correspond to the Gaussian priors on weights $p(\vw)=\mathcal{N}(\vect{0}, \texttt{diag}(\vect{\sigma_0^2}))$ ; (\RN{2}) they carry valuable inductive bias of certain \emph{finitely wide} DNN architectures; (\RN{3}) they are more accessible than NN-GPs in practice; (\RN{4}) they have been evaluated in some related works like \cite{wang2018function}. 
Concretely, supposing a \emph{finitely wide} DNN composed of a feature projector ${h(\cdot, \vw): \mathcal{X} \rightarrow \mathbb{R}^{\hat{C}}}$ and a linear readout layer with weight variance $\sigma_{w}^2$ and bias variance $\sigma_{b}^2$, the MC NN-GP prior is $p(f)=\mathcal{GP}(f|0, k_p(\vx, \vx'))$, where
% \vspace{-0.5ex}
\begin{equation}
\label{eq:enngp}
\small
\begin{aligned}
k_p(\vx, \vx') &:= (\sigma_{w}^2 \hat{k}_p(\vx, \vx') + \sigma_{b}^2)\mathbf{I}_C, \\ 
\text{and} \,\, \hat{k}_p(\vx, \vx') &:= \frac{1}{S\hat{C}}\sum_{s=1}^S h(\vx, \vw_s)^\top h(\vx', \vw_s).
\end{aligned}
\end{equation}
$\vw_s$ are i.i.d. samples from $p(\vw)$.
We highlight that $\hat{k}_p$ is scalar-valued yet ${k}_p$ is matrix-valued. 
Keep in mind that the MC NN-GP priors can be defined with distinct architectures from the DE-GP approximate posteriors. 

In practice, DE-GP benefits from learning in the parametric family specified by DNNs ensemble, hence can even outperform the analytical NN-GP posteriors on some metrics. 
%We actually use the MC estimate of the NN-GP prior during training.

% As will be shown later, the variational inference in the GP family is much simpler than that using general stochastic processes.

% We specify the functional prior as the aforementioned NN-GPs $p(f)=\mathcal{GP}(f|0, k(\vx, \vx'))$
% % as the inference resides in function space, in contrast to the variational inference in weight space, which entails the prior and the variational in the same parametric family.}
% to enjoy the inductive bias carried by specific NN architectures.\footnote{Note that the NN-GP prior and the DE-GP variational posterior can be defined with various architectures. We actually use the MC estimate of the NN-GP prior during training.} 

% Yet, the current implementations routinely suffer from scalability issues and struggle to handle modern network configurations.
% Thus, \cite{novak2018bayesian} developed {an} MC estimate of $k$ using \emph{finitely wide} NNs, in a similar spirit to the classic random feature approaches~\citep{rahimi2007random}.

% To set up the NN-GP prior with a deep architecture, we set the prior variance on weight following common practice like {He initialization}~\citep{he2015delving}. 
% The MC estimate of $k(\vx, \vx')$ equals to the MC NN-GP kernels defined in Eq~(\ref{eq:enngp}). 

\textbf{fELBO} Following \cite{sun2018functional}, we maximize the functional ELBO (fELBO) to achieve fVI: %, an extension of \cref{eq:w-elbo}:
\begin{equation}
\label{eq:f-elbo}
\small
    \max_{q(f)} \mathbb{E}_{q(f)}[\log p(\mathcal{D}|f)] - \KL[q(f)\Vert p(f)].
\end{equation}
% with $p(\mathcal{D}|f) = \prod_i p(y_i|f(\vx_i))$. 
Notably, there is a KL divergence between two GPs, which, on its own, is challenging to cope with. 
Fortunately, as proved by \cite{sun2018functional}, we can take the KL divergence between the {marginal distributions} of function evaluations as a substitute for it, giving rise to a more tractable objective:
\begin{equation}
\label{eq:lb}
\small
\begin{aligned}
\Scale[0.9]{\mathcal{L} = \mathbb{E}_{q(f)}\left[ \sum_{(\vx_i,\vy_i)\in\mathcal{D}} \log p(\vy_i|f(\vx_i))\right] - \KL\left[q(\mathbf{f}^{\tilde{\mathbf{X}}})\Vert p(\mathbf{f}^{\tilde{\mathbf{X}}})\right],} %\\
% & = \log p(\mathcal{D}) - \KL[q(\mathbf{f}^{\tilde{\mathbf{X}}})\Vert p(\mathbf{f}^{\tilde{\mathbf{X}}}|\mathcal{D})] \leq \log p(\mathcal{D}),
\end{aligned}
\end{equation}
where $\tilde{\mathbf{X}}$ denotes a \emph{measurement set} including all training inputs $\mathbf{X}$, and $\mathbf{f}^{\tilde{\mathbf{X}}}$ is the concatenation of the vectorized outputs of $f$ for $\tilde{\mathbf{X}}$, i.e., $\mathbf{f}^{\tilde{\mathbf{X}}} \in \mathbb{R}^{|\tilde{\mathbf{X}}|C}$.\footnote{We use $|\tilde{\mathbf{X}}|$ to notate the size of a set $\tilde{\mathbf{X}}$.}

It has recently been shown that the fELBO is often ill-defined, since the KL divergence in function space is infinite~\cite{burt2020understanding}, which may lead to several pathologies. 
Yet, \cref{fig:illu2} and \cref{fig:illu}, which form posterior approximation quality checks, prove the empirical efficacy of the used fELBO. 

% In this way, we can obtain a tractable lower bound $\mathcal{L}$ for the log marginal likelihood.
% \end{restatable}
% This is a corollary of the Theorem 2 in \citep{sun2018functional} which is for general stochastic processes.\junz{if this is straightforward, don't put it as a theorem; put the deviation in plain text, to avoid over-claim. readers may expect nontrivial results from a theorem.}

% \vspace{-0.5ex}
\subsection{Training}
% \vspace{-0.5ex}
We outline the training procedure of DE-GP in \cref{algo:1}, and elaborate some details below. 

% \vspace{-0.5ex}
\subsubsection{Mini-batch Training}
% \vspace{-0.5ex}
In deep learning scenarios, DE-GP should proceed by mini-batch training. 
At each step, we manufacture a stochastic measurement set with a mini-batch $\mathcal{D}_s=(\mathbf{X}_{s}, \mathbf{Y}_{s})$ from the training data $\mathcal{D}$ and some random samples $\mathbf{X}_{\nu}$ from a continuous distribution (e.g., a uniform distribution) $\nu$ supported on $\mathcal{X}$. 
Then, we adapt the objective defined in \cref{eq:lb} to the following form:
\begin{equation}
\label{eq:lb-s}
\small
\begin{aligned}
\underset{ \vw_1, ...,\vw_M}{\max} \,\mathcal{L} = \underset{ \vw_1, ...,\vw_M}{\max} \,& \Scale[0.95]{\mathbb{E}_{f\sim q(f)}\left[\sum_{(\vx_i,\vy_i)\in\mathcal{D}_s}\log p(\vy_i|f(\vx_i))\right]} \\
& \; \Scale[0.95]{- \, \alpha \KL\left[q(\mathbf{f}^{\tilde{\mathbf{X}}_s})\Vert p(\mathbf{f}^{\tilde{\mathbf{X}}_s})\right]},
\end{aligned}
\end{equation}
where ${\tilde{\mathbf{X}}_s}$ indicates the union of ${\mathbf{X}_{s}}$ and ${\mathbf{X}_{\nu}}$. 
We opt to fix the hyper-parameters specifying the MC NN-GP prior $p(f)$, but to tune the coefficient $\alpha$ to better trade off between data evidence and priori regularization rather than fixing $\alpha$ as 1.
When tuning $\alpha$, we intentionally set it as large as possible to avoid colder posteriors and worse uncertainty estimates.

The importance of the incorporation of extra measurement points $\mathbf{X}_{\nu}$ depends on the data and the problem at hand. 
\cref{app:illu-abl} reports a study on the aforementioned 1-D regression where the DE-GP is trained without the incorporation of $\mathbf{X}_{\nu}$, and the results are still seemingly promising.

% on weights for defining the prior GP 

% \vspace{-.5ex}
\subsubsection{Analytical Estimation of KL}
% \vspace{-.5ex}
We then provides strategies to efficiently estimate the KL term in \cref{eq:lb-s}. 
Thanks to the variational inference in the GP family, the marginal distributions in the KL are both multivariate Gaussians, i.e., $q(\mathbf{f}^{\tilde{\mathbf{X}}_s}) = \mathcal{N}(\mathbf{f}^{\tilde{\mathbf{X}}_s}|\mathbf{m}^{\tilde{\mathbf{X}}_s}, \mathbf{k}^{\tilde{\mathbf{X}}_s,\tilde{\mathbf{X}}_s})$, $p(\mathbf{f}^{\tilde{\mathbf{X}}_s}) = \mathcal{N}(\mathbf{f}^{\tilde{\mathbf{X}}_s}|\mathbf{0}, \mathbf{k}_p^{\tilde{\mathbf{X}}_s,\tilde{\mathbf{X}}_s})$, with the kernel matrices $\mathbf{k}^{\tilde{\mathbf{X}}_s,\tilde{\mathbf{X}}_s}, \mathbf{k}_p^{\tilde{\mathbf{X}}_s,\tilde{\mathbf{X}}_s} \in \mathbb{R}^{|\tilde{\mathbf{X}}_s|C \times |\tilde{\mathbf{X}}_s|C}$ as the joints of pair-wise outcomes. 

Thus, the marginal KL divergence and its gradients can be estimated exactly without resorting to complicated approximations~\cite{sun2018functional,rudner2021rethinking}. 
We further offer prescriptions for efficiently computing the inversion of $\mathbf{k}_p^{\tilde{\mathbf{X}}_s,\tilde{\mathbf{X}}_s}$ and the determinant of $\mathbf{k}^{\tilde{\mathbf{X}}_s,\tilde{\mathbf{X}}_s}$ involved in the KL.
% It is easy to show that
% \begin{equation}
% \label{eq:grad}
% \small
% \nabla_{\mathbf{m}_q^{\mathbf{X}}} \KL[q(\mathbf{f}^{\mathbf{X}})\Vert p(\mathbf{f}^{\mathbf{X}}|\mathcal{D})] = (\mathbf{k}_p^{\mathbf{X},\mathbf{X}})^{-1} \mathbf{m}_q, \nabla_{\mathbf{k}_q^{\mathbf{X},\mathbf{X}}} \KL[q(\mathbf{f}^{\mathbf{X}})\Vert p(\mathbf{f}^{\mathbf{X}}|\mathcal{D})] = \frac{1}{2}((\mathbf{k}_p^{\mathbf{X},\mathbf{X}})^{-1} - (\mathbf{k}_q^{\mathbf{X},\mathbf{X}})^{-1}).
% \end{equation}
%Nevertheless, when the output dimension $C$ is large, the involved matrix inversion on $\mathbf{k}_q^{\tilde{\mathbf{X}}_s,\tilde{\mathbf{X}}_s}, \mathbf{k}^{\tilde{\mathbf{X}}_s,\tilde{\mathbf{X}}_s}$ may be prohibitive due to $\mathcal{O}(|\tilde{\mathbf{X}}_s|^3C^3)$ time complexity. 

As discussed in \cref{sec3.2}, there is a simple structure in $\mathbf{k}_p^{\tilde{\mathbf{X}}_s,\tilde{\mathbf{X}}_s}$, so we can write it in the form of Kronecker product:
\begin{equation}
\label{eq:p-k}
\small
\mathbf{k}_p^{\tilde{\mathbf{X}}_s,\tilde{\mathbf{X}}_s} = (\sigma_{w}^2 \mathbf{\hat{k}}_p^{\tilde{\mathbf{X}}_s, \tilde{\mathbf{X}}_s} + \sigma_{b}^2) \otimes \mathbf{I}_C,
\end{equation}
where $\mathbf{\hat{k}}_p^{\tilde{\mathbf{X}}_s, \tilde{\mathbf{X}}_s} \in \mathbb{R}^{|\tilde{\mathbf{X}}_s|\times|\tilde{\mathbf{X}}_s|}$ corresponds to the evaluation of kernel $\hat{k}_p$.
Hence we can exploit the property of Kronecker product to inverse $\mathbf{k}_p^{\tilde{\mathbf{X}}_s,\tilde{\mathbf{X}}_s}$ in $\mathcal{O}(|\tilde{\mathbf{X}}_s|^3)$ complexity.

Besides, as $\mathbf{k}^{\tilde{\mathbf{X}}_s,\tilde{\mathbf{X}}_s}$ is low-rank, we can leverage the {matrix determinant lemma}~\citep{harville1998matrix} to compute the determinant of $\mathbf{k}^{\tilde{\mathbf{X}}_s,\tilde{\mathbf{X}}_s}$ in $\mathcal{O}(|\tilde{\mathbf{X}}_s|CM^2)$ time given that usually $M \ll |\tilde{\mathbf{X}}_s|C$ (e.g., $10$ $\ll$ $256C$).

\begin{figure}[t]
% \vspace{-0.2cm}
\centering
\includegraphics[width=0.95\linewidth]{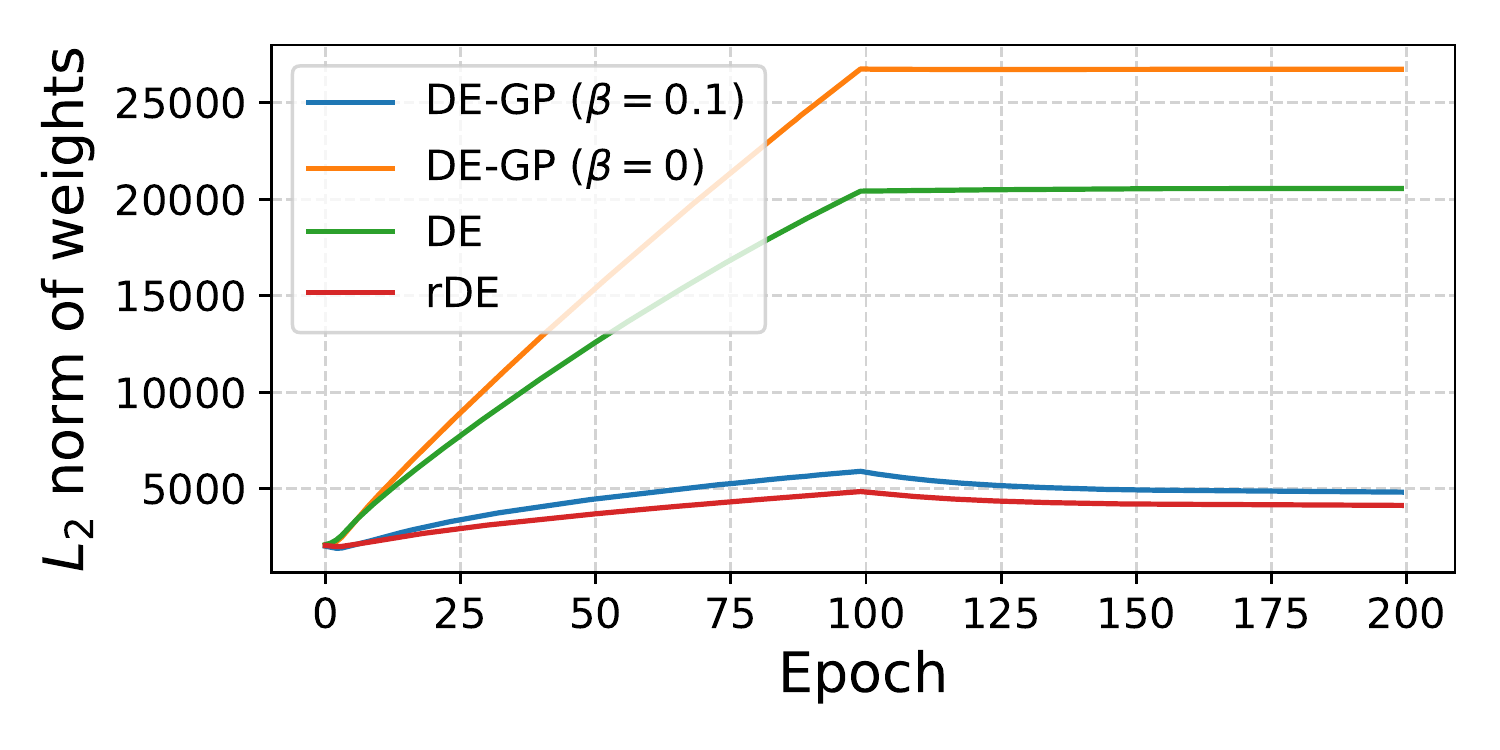}
\vspace{-0.35cm}
\caption{\footnotesize The $L_2$ norm of weights varies w.r.t. training step. The models are trained on CIFAR-10 with ResNet-20 architecture. DE-GP ($\beta=0$) finds solutions with high complexity and poor test accuracy (see \cref{table:1}), yet DE-GP ($\beta=0.1$) settles this.}
\label{fig:w-norm}
\vspace{-0.05cm}
\end{figure}

% \vspace{-0.5ex}
\subsubsection{An Extra Regularization on Weights}
% \vspace{-0.5ex}
Generally, we usually introduce an $L_2$ regularization term on weights when optimizing over a parametric function class like DNNs. 
In this sense, we can generalize \cref{eq:lb-s} as:
\begin{equation}
\label{eq:lb-s2}
\small
\begin{aligned}
\underset{ \vw_1, ...,\vw_M}{\max} \,\mathcal{L} &= \Scale[1]{\mathbb{E}_{f\sim q(f)}\left[\sum_{(\vx_i,\vy_i)\in\mathcal{D}_s}\log p(\vy_i|f(\vx_i))\right]} \\
& \;\;\; \Scale[1]{- \,\alpha \KL\left[q(\mathbf{f}^{\tilde{\mathbf{X}}_s})\Vert p(\mathbf{f}^{\tilde{\mathbf{X}}_s})\right] - \beta \sum_i||\vw_i||_2^2},
\end{aligned}
\end{equation}
where $\beta$ is set as 0 in default. 
Ideally, the KL divergence in \cref{eq:lb-s2} is enough to help DE-GP to resist over-fitting (in function space). 
Its effectiveness is evidenced by the results in \cref{fig:illu2}, but we have empirically observed that it may lose efficacy when facing deep architectures like ResNets~\cite{he2016deep}.

Specifically, we conducted a set of experiments with DE, rDE, and DE-GP using ResNet-20 architecture on CIFAR-10 benchmark~\cite{krizhevsky2009learning}.
We plot how the $L_2$ norm of $\{\vw_i\}_{i=1}^M$ varies w.r.t. training step in \cref{fig:w-norm} and display the comparison on test accuracy in \cref{table:1}.

An immediate conclusion is that the KL in \cref{eq:lb-s2} cannot cause proper regularization effects on weights for DE-GP when using ResNet-20,\footnote{This is interesting. We deduce that this may be partly attributed to the high non-linearity of DNNs.} thus the learned DE-GP suffers from high complexity and hence poor performance. 

Having identified this, we suggest activating the $L_2$ weight regularization term in \cref{eq:lb-s2} \emph{when handling deep architectures}. 
In practice, we can set $\beta$ according to commonly used weight decay coefficient. 
We trained a DE-GP with $L_2$ weight regularization of intensity $0.1$ for the above case. 
The results in \cref{fig:w-norm} and \cref{table:1} testify its effectiveness. 

% Readers familiar with Bayesian inference may concern about that the extra regularization upon fVI leads to biased posterior inference. 
% We admit that this is the case. \junz{these two sentences seem to make things complicated. It's better to focus on providing insights to justify the regularization.}
This extra regularization may introduce bias to the posterior inference corresponding to the imposed MC NN-GP prior.
But, if we think of the extra regularization as a kind of extra prior knowledge, we can then justify it within the posterior regularization scheme~\cite{ganchev2010posterior} (see \cref{app:postreg}).

\begin{algorithm}[t] 
\caption{\small The training of DE-GP} 
\label{algo:1}
\footnotesize
\begin{algorithmic}[1]
\STATE {\bfseries Input:} $\mathcal{D}$: dataset; $\Scale[0.95]{\{g(\cdot, \vw_i)\}_{i=1}^M}$: a deep ensemble of size $M$; $k_p$: MC NN-GP prior kernel; $\Scale[0.95]{\alpha, \beta}$: trade-off coefficients; $\Scale[0.95]{\nu}$: distribution for sampling extra measurement points; $U$: number of MC samples for estimating the expected log-likelihood
\WHILE{not converged}
    \STATE $\Scale[0.92]{\mathcal{D}_s=(\mathbf{X}_{s}, \mathbf{Y}_{s}) \subset \mathcal{D},\, \mathbf{X}_{\nu} \sim \nu,\tilde{\mathbf{X}}_s=\{\mathbf{X}_{s},\, \mathbf{X}_{\nu}\}}$
    \STATE $\Scale[0.92]{\mathbf{g}_i^{\tilde{\mathbf{X}}_s}=g(\tilde{\mathbf{X}}_s, \vw_i),\, i=1,...,M}$
    \STATE $\Scale[0.92]{\mathbf{m}^{\tilde{\mathbf{X}}_s}=\frac{1}{M}\sum_i \mathbf{g}_i^{\tilde{\mathbf{X}}_s}}$
    \STATE $\Scale[0.92]{ \mathbf{k}^{\tilde{\mathbf{X}}_s,\tilde{\mathbf{X}}_s} = \frac{1}{M}\sum_{i=1}^M (\mathbf{g}_i^{\tilde{\mathbf{X}}_s}-\mathbf{m}^{\tilde{\mathbf{X}}_s}) (\mathbf{g}_i^{\tilde{\mathbf{X}}_s}-\mathbf{m}^{\tilde{\mathbf{X}}_s})^\top + \lambda \mathbf{I}_{|\tilde{\mathbf{X}}_s|C}}$%, $q(\mathbf{f}^{\tilde{\mathbf{X}}_s}) = \mathcal{N}(\mathbf{f}^{\tilde{\mathbf{X}}_s}|\mathbf{m}_q^{\tilde{\mathbf{X}}_s}, \mathbf{k}_q^{\tilde{\mathbf{X}}_s,\tilde{\mathbf{X}}_s})$
    \STATE $\Scale[0.92]{\mathbf{k}_p^{\tilde{\mathbf{X}}_s,\tilde{\mathbf{X}}_s}=k_p(\tilde{\mathbf{X}}_s, \tilde{\mathbf{X}}_s)}$ 
    \STATE $\Scale[0.87]{\mathcal{L}_1 = \frac{1}{U}\sum_{i=1}^U\sum_{(\vx,\vy)\in \mathcal{D}_s} \log p(\vy|\mathbf{f}_i(\vx)),\,\mathbf{f}_i \sim \mathcal{N}(\mathbf{m}^{\tilde{\mathbf{X}}_s}, \mathbf{k}^{\tilde{\mathbf{X}}_s,\tilde{\mathbf{X}}_s})}$
    \STATE $\Scale[0.92]{\mathcal{L}_2=\KL[\mathcal{N}(\mathbf{m}^{\tilde{\mathbf{X}}_s}, \mathbf{k}^{\tilde{\mathbf{X}}_s,\tilde{\mathbf{X}}_s})\Vert \mathcal{N}(\mathbf{0}, \mathbf{k}_p^{\tilde{\mathbf{X}}_s,\tilde{\mathbf{X}}_s})]}$ 
    \STATE $\Scale[0.92]{\mathcal{L}_3 = \sum_i||\vw_i||_2^2}$
    \STATE $\Scale[0.92]{\vw_i = \vw_i + \eta \nabla_{\vw_i}(\mathcal{L}_1 - \alpha\mathcal{L}_2 -\beta\mathcal{L}_3),\, i=1,...,M}$
\ENDWHILE
%   \KwRet the model
\end{algorithmic}
\end{algorithm}

\begin{figure*}[t]
% \vspace{-0.1cm}
% \vspace{-.5ex}
\centering
\begin{minipage}{0.49\linewidth}
% \begin{figure}
\centering
    \begin{subfigure}[b]{\linewidth}
    \centering
    \includegraphics[width=\linewidth]{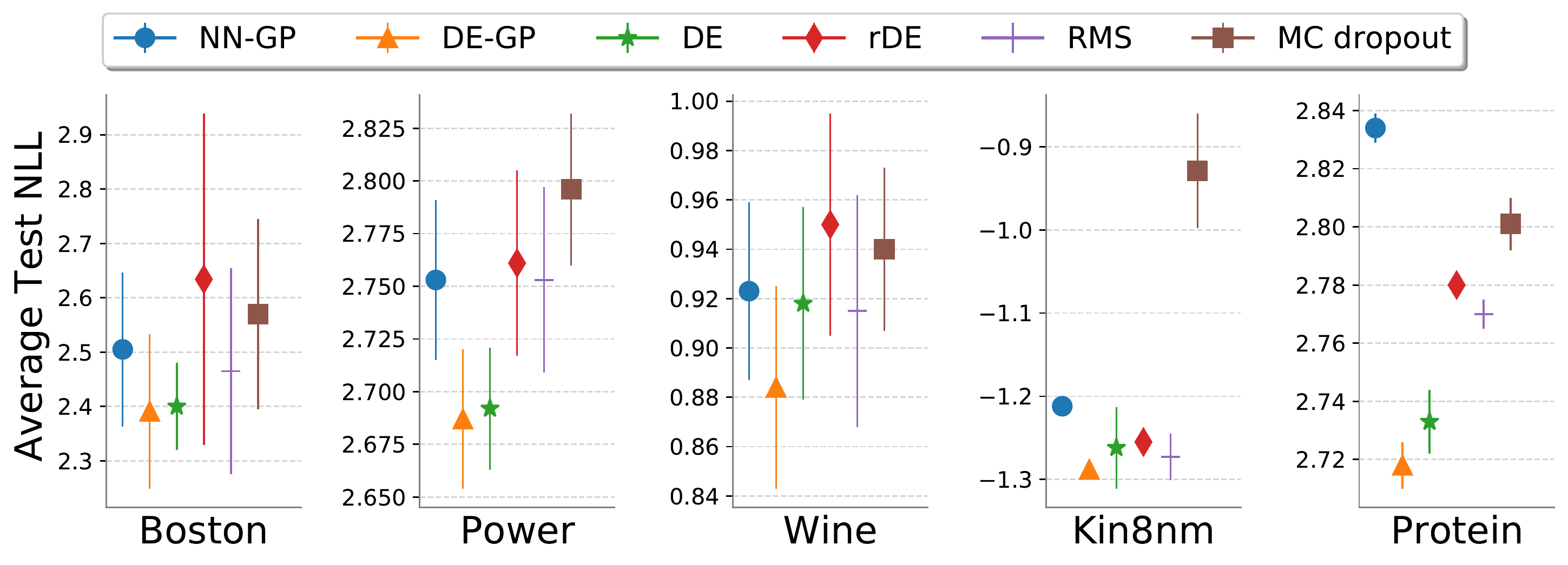}
        % \vspace{-4.ex}
        % \caption{\scriptsize }
        % \label{fig:ood-cifar}
    \end{subfigure}
\end{minipage}
\begin{minipage}{0.49\linewidth}
% \begin{figure}
\centering
    \begin{subfigure}[b]{\linewidth}
    \centering
    \includegraphics[width=\linewidth]{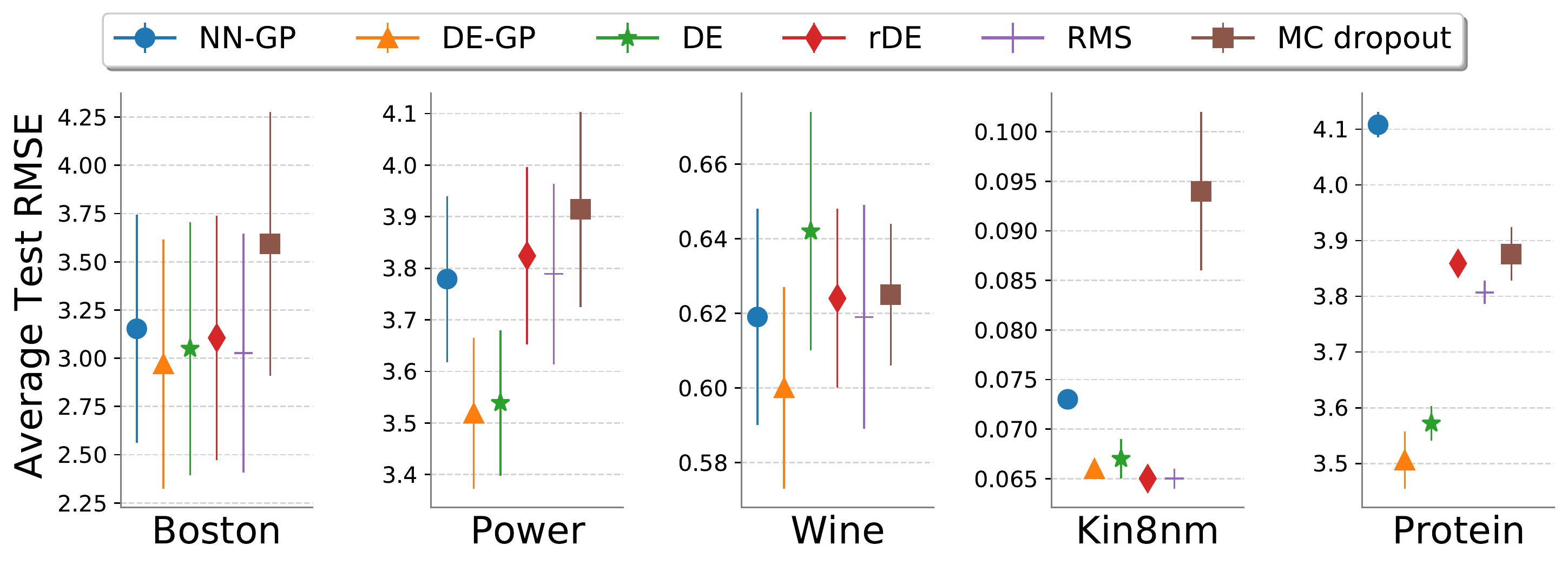}
        % \vspace{-4.ex}
        % \caption{\scriptsize }
        % \label{fig:ood-cifar}
    \end{subfigure}
\end{minipage}
\vspace{-2ex}
\caption{\small Comparison on average test NLL and RMSE on UCI regression problems. The lower the better.}
\vspace{-1ex}
\label{fig:uci}
\end{figure*}

\begin{figure*}[t]
\vspace{-0.1cm}
\centering
\begin{minipage}{0.99\linewidth}
% \begin{figure}
\centering
    \begin{subfigure}[b]{0.315\linewidth}
    \centering
    \includegraphics[width=\linewidth]{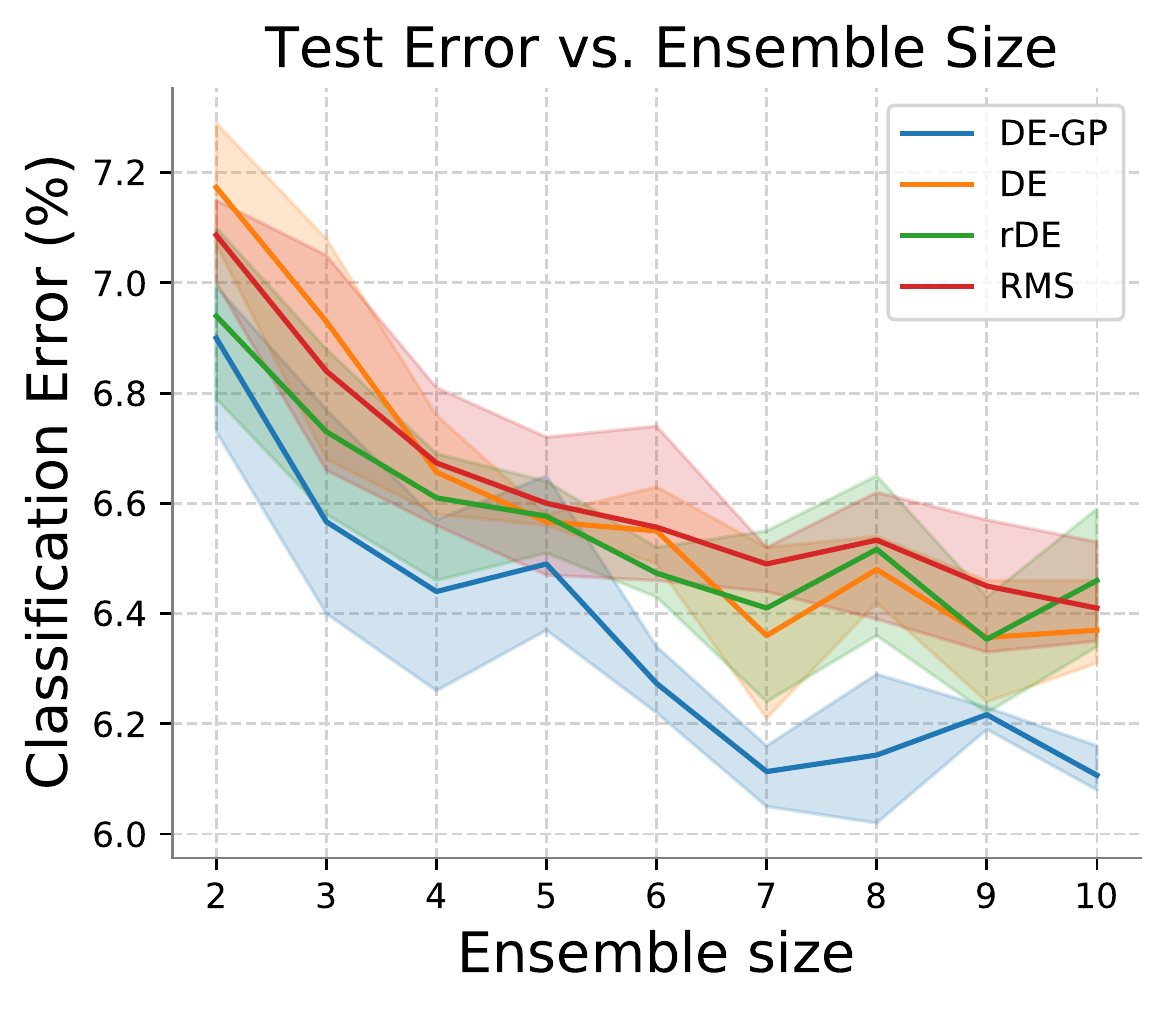}
        % \vspace{-4.ex}
    % \caption{\scriptsize }
        % \label{fig:ood-cifar}
    \end{subfigure}
    \begin{subfigure}[b]{0.32\linewidth}
    \centering
    \includegraphics[width=\linewidth]{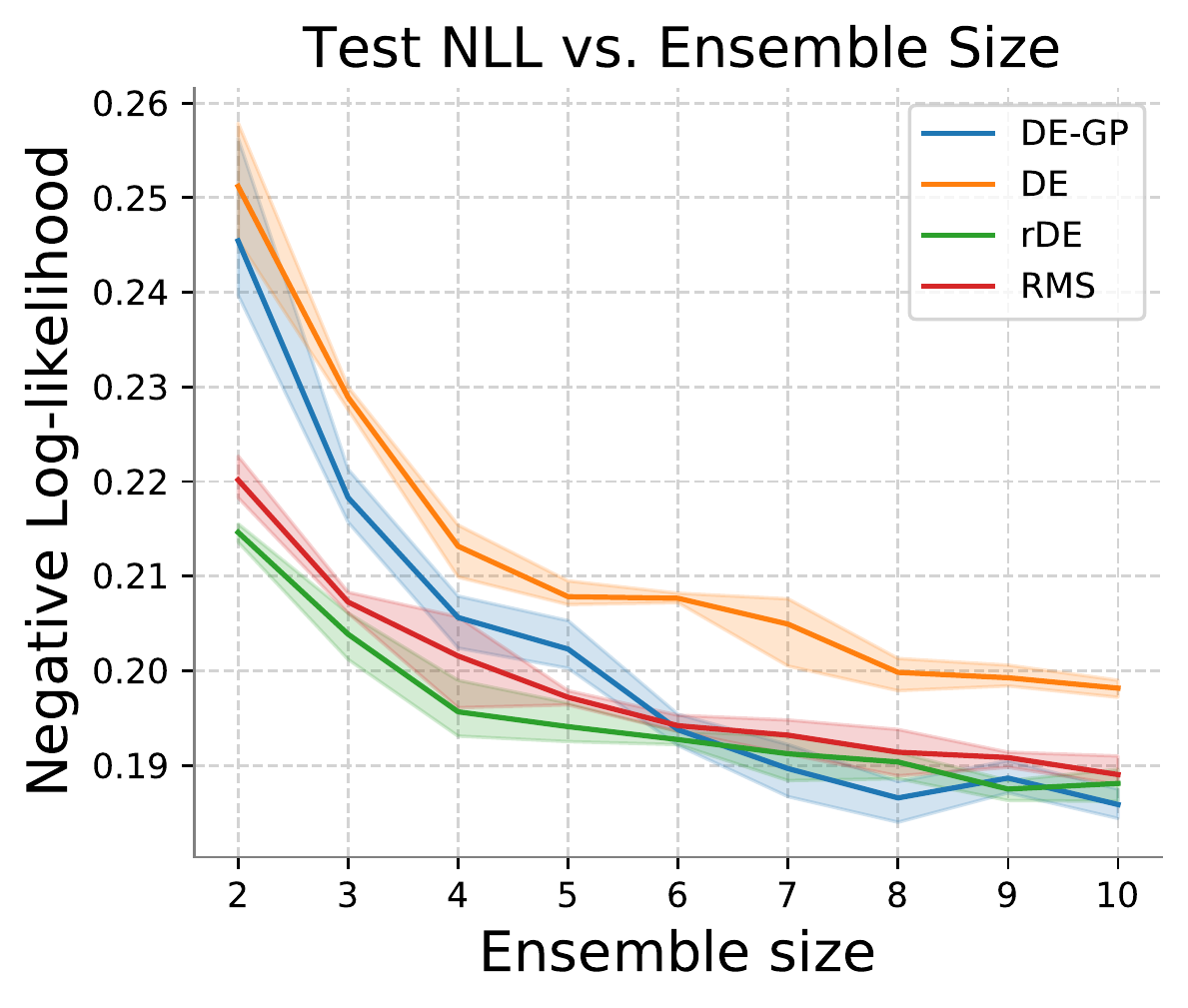}
        % \vspace{-4.ex}
    % \caption{\scriptsize }
        % \label{fig:ood-cifar}
    \end{subfigure}
    \begin{subfigure}[b]{0.33\linewidth}
    \centering
    \includegraphics[width=\linewidth]{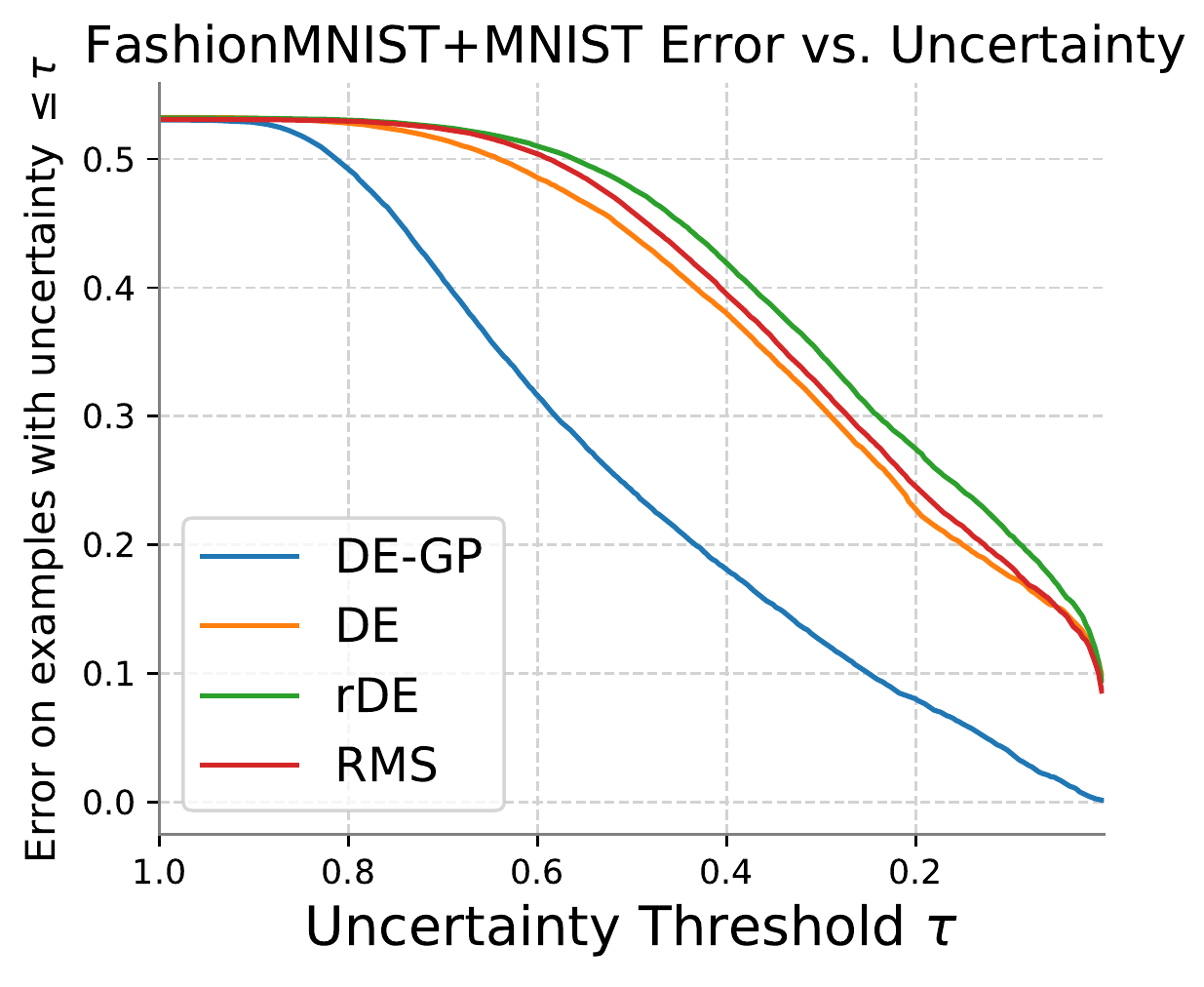}
        % \vspace{-4.ex}
    % \caption{\scriptsize }
        % \label{fig:ood-cifar}
    \end{subfigure}
\end{minipage}
\vspace{-2ex}
\caption{\small (Left): Test error varies w.r.t. ensemble size on Fashion-MNIST. 
(Middle): Test NLL varies w.r.t. ensemble size on Fashion-MNIST. 
(Right): Test error versus uncertainty plots for methods trained on Fashion-MNIST and tested on both Fashion-MNIST and MNIST. Ensemble size is fixed as 10.}
\label{fig:fmnist}
\vspace{-2ex}
\end{figure*}

% \vspace{-.5ex}
\subsection{Discussions}
% \vspace{-.5ex}
\label{sec:discussion}
\textbf{Diversity.} The diversity among the ensemble members in function space is explicitly encouraged by the KL divergence term in \cref{eq:lb-s2}.
Nonetheless, the expected log-likelihood in \cref{eq:lb-s2} enforces each ensemble member to yield the same, correct outcomes for the training data. 
Thereby, the diversity mainly exists in the regions far away from the training data (see \cref{fig:illu2}). 
Yet, the diversity in DE does not have a clear theoretical support. 

\textbf{Efficiency.}
Compared to the overhead introduced by DNNs, the effort for estimating the KL in \cref{eq:lb-s2} is negligible. 
The added cost of DE-GP primarily arises from the extra measurement points and the evaluation of the prior kernels. 
% When adopting a classic kernel like RBF, the kernel estimation is cheap and the whole pipeline is on par with vanilla DE in speed; but when using advanced kernels like the NN-GP kernels, the computational efficiency deteriorates.
In practice, we use a small batch size for the extra measurement points.
We build the MC NN-GP prior kernels with cheap architectures and perform MC estimation in parallel. 
Eventually, DE-GP is only marginally slower than DE. 

\textbf{Weight sharing.} DE-GP does not care about how $g_i$ are parameterized, so we can perform weight sharing among $g_i$, for example, using a shared feature extractor and $M$ independent MLP classifiers to construct $M$ ensemble members~\citep{Deng_2021_CVPR}. 
With shared weights, DE-GP is still likely to be reliable because our learning principle induces diversity in function space regardless of the weights. 
Experiments in \cref{sec:img-clf} validate this.

\textbf{Limitations.} Despite being Bayesian in principle, DE-GP loses parallelisability. Nevertheless, it may be a common issue, e.g., the methods of \cite{d2021repulsive} also entail concurrent updates of the ensemble members.

% is likely to be less flexible than DE and other variants because all the involved $M$ models need to be updated simultaneously.  %\jianfei{do reviewers ask us to discuss the limitations?}
% So, we recommend using DE-GP the ML practitioners need to make a trade-off between theoretical soundness and flexibility in practical usage.

% However, the promise of BNNs is undermined by insurmountable inference challenges in practice, and BNNs typically reveal poor uncertainty quality of OOD robustness~\citep{ovadia2019can}.

% \vspace{-1.ex}
\section{Experiments}
% \vspace{-1ex}
\label{sec:exp}
We perform extensive evaluation to demonstrate that DE-GP yields better uncertainty estimates than the baselines, while preserving non-degraded predictive performance. 
The baselines include DE, rDE, NN-GP, RMS, etc. 
In \emph{all} experiments, we estimate the MC NN-GP prior kernels with $10$ MC samples and set the sampling distribution for extra measurement points $\nu$ as the uniform distribution over the data region. 
The number of MC samples for estimating the expected log-likelihood (i.e., $U$ in \cref{algo:1}) is $256$. 
Unless otherwise stated, we set the regularization constant $\lambda$ as $0.05$ times of the average eigenvalue of the central covariance matrices, and set the weight and bias variance for defining the MC NN-GP prior kernel at each layer as $2/\texttt{fan\_in}$ and $0.01$, where $\texttt{fan\_in}$ is the number of input features, as suggested by \citet{he2015delving}.
% Code is provided in the supplementary material. 

% \vspace{-1ex}
\subsection{Illustrative 1-D Regression}
% \vspace{-1ex}
We build a regression problem with $8$ data from $y=\sin 2x+\epsilon, \epsilon \sim \mathcal{N}(0, 0.2)$ as shown in \cref{fig:illu}. 
The rightest datum is deliberately perturbed to cause data noise. 
For NN-GP, we analytically estimate the GP kernel and perform GP regression without training DNNs. 
For DE-GP, DE, and rDE, we train 50 MLPs. 
By default, we set $\alpha=1$, $\beta=0$.

\cref{fig:illu} and \cref{fig:illu2} show the comparison on prediction and training efficiency. 
As shown, DE-GP delivers calibrated uncertainty estimates across settings, on par with the non-parametric Bayesian baseline NN-GP. 
Yet, DE and rDE suffer from degeneracy issue as the dimension of weights increases. 
Though NN-GP outperforms other methods, the involved analytical GP regression may have scalability and effectiveness issues when facing modern architectures~\citep{novak2018bayesian}, while DE-GP does not suffer from them.

\begin{table*}[t]
%   \vspace{-0.25cm}
  \centering
 \footnotesize
\caption{\small Test accuracy comparison on CIFAR-10. Results are summarized over 8 trials.}
\vspace{-1ex}
  \label{table:1}
  \begin{tabular}{c||p{17ex}<{\centering}|p{17ex}<{\centering}|p{17ex}<{\centering}|p{17ex}<{\centering}|p{17ex}<{\centering} }%|p{17.5ex}<{\centering}}
  \hline
{Architecture}& \emph{DE-GP ($\beta=0.1$)} & \emph{DE-GP ($\beta=0$)} & \emph{DE} & \emph{rDE} & \emph{RMS} \\% & ResNet-110\\
\hline
ResNet-20 & \textbf{94.67}$\pm$0.04\% &93.71$\pm$0.06\%&93.43$\pm$0.08\%&\textbf{94.58}$\pm$0.05\%&93.63$\pm$0.07\%\\
ResNet-56 &\textbf{95.55}$\pm$0.04\%&94.24$\pm$0.07\%&94.04$\pm$0.07\%&\textbf{95.56}$\pm$0.06\%&94.45$\pm$0.03\%\\ %use results of resnet20 prior
  \hline
   \end{tabular}
    % \vspace{-0.1cm}
%   \vspace{-2ex}
\vspace{0.1cm}
\end{table*}

\begin{figure*}[t]
\vspace{-0.1cm}
\centering
\begin{minipage}{0.99\linewidth}
% \begin{figure}
\centering
    \begin{subfigure}[b]{0.49\linewidth}
    \centering
    \includegraphics[width=\linewidth]{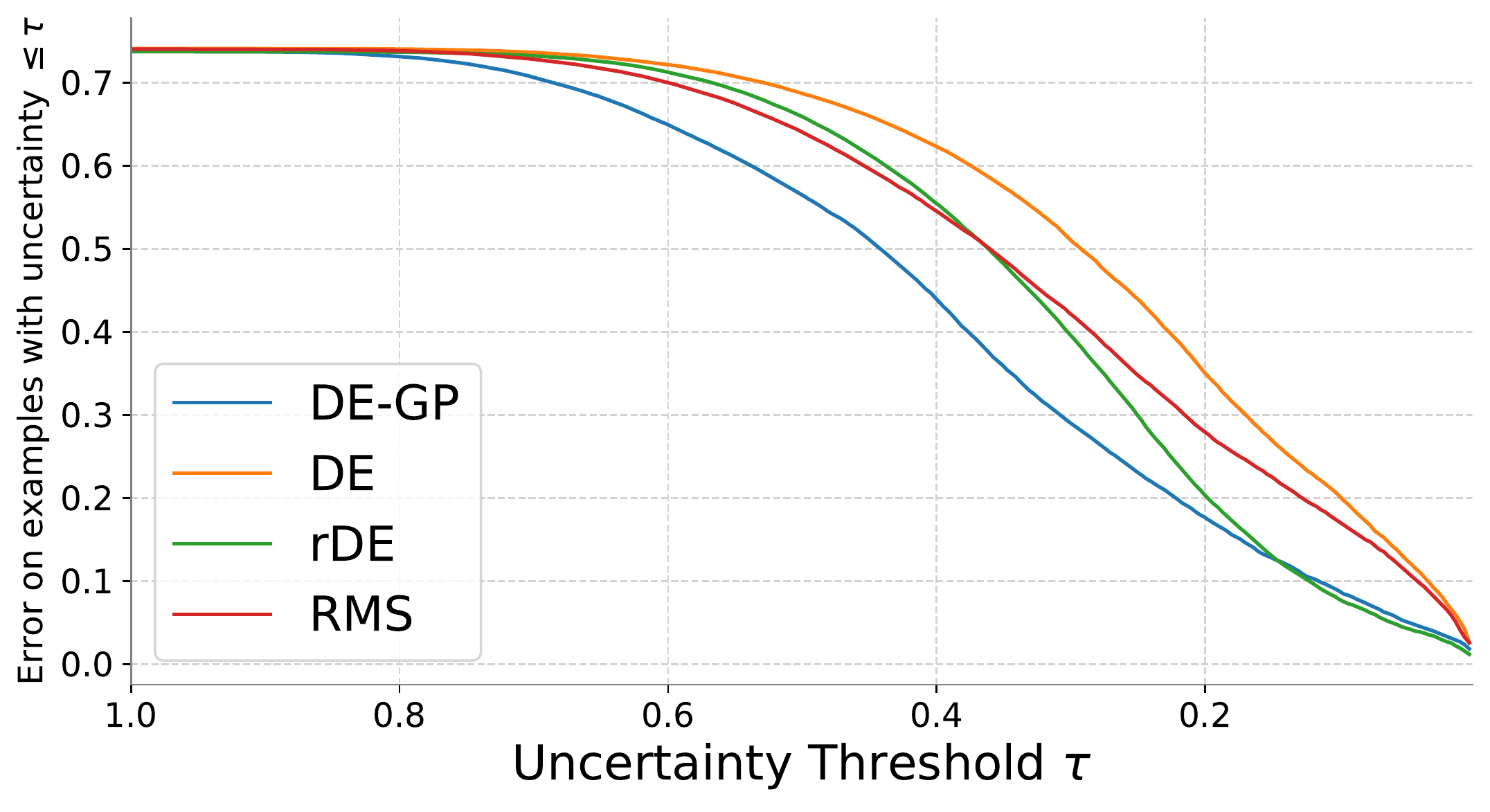}
        % \vspace{-4.ex}
    % \caption{\scriptsize }
        % \label{fig:ood-cifar}
    \end{subfigure}
    \begin{subfigure}[b]{0.49\linewidth}
    \centering
    \includegraphics[width=\linewidth]{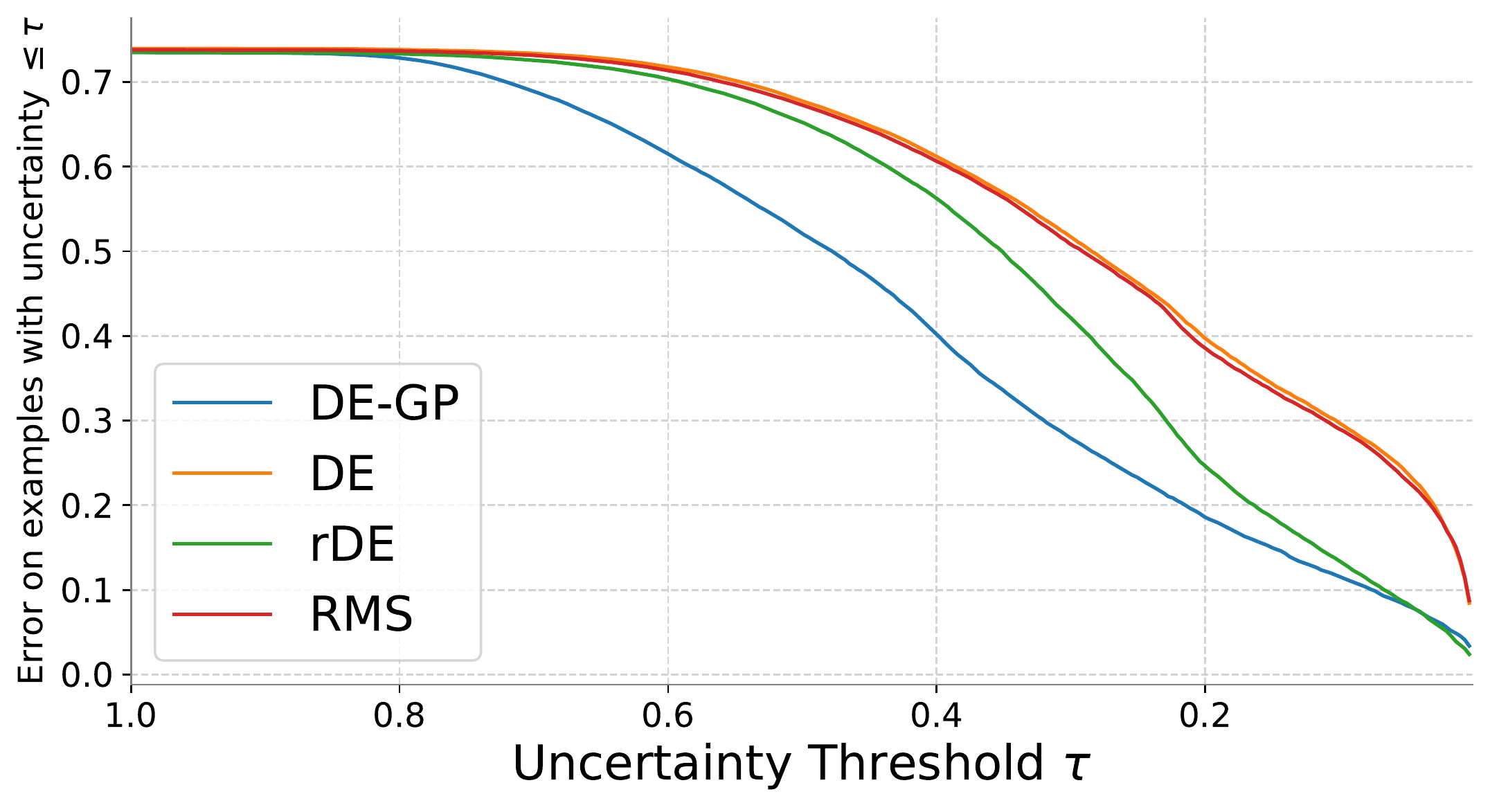}
        % \vspace{-4.ex}
    % \caption{\scriptsize }
        % \label{fig:ood-cifar}
    \end{subfigure}
    % \begin{subfigure}[b]{0.32\linewidth}
    % \centering
    % \includegraphics[width=\linewidth]{cifar/acc_vs_conf_cifar_resnet110.pdf}
    %     % \vspace{-4.ex}
    % \caption{\scriptsize }
    %     % \label{fig:ood-cifar}
    % \end{subfigure}
\end{minipage}
\vspace{-1.ex}
\caption{\small Test error versus uncertainty plots for methods trained on CIFAR-10 and tested on both CIFAR-10 and SVHN with ResNet-20 (Left) or ResNet-56 (Right) architecture.}
\label{fig:acc-vs-conf-cifar}
% \vspace{-0.1cm}
\end{figure*}

\begin{figure*}[t]
% \vspace{-0.1cm}
\centering
\begin{minipage}{0.99\linewidth}
% \begin{figure}
\centering
    \begin{subfigure}[b]{0.49\linewidth}
    \centering
    \includegraphics[width=\linewidth]{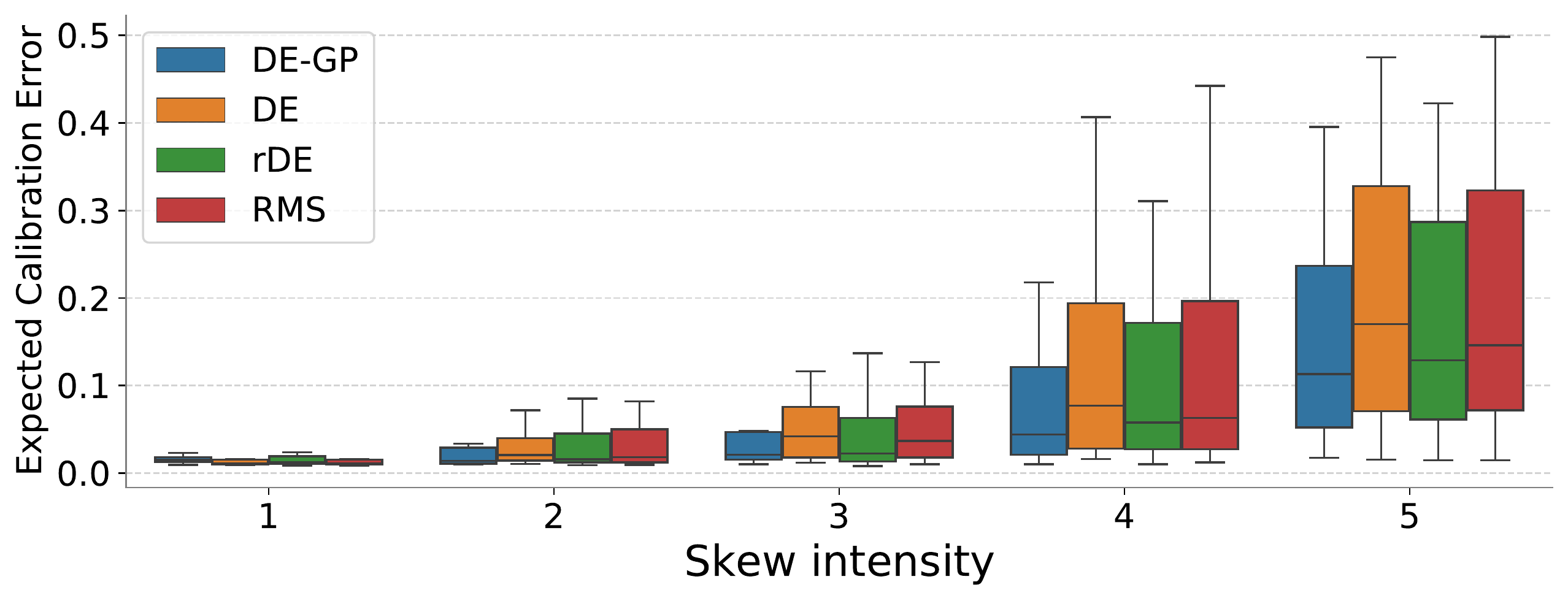}
        % \vspace{-4.ex}
    % \caption{\scriptsize }
        % \label{fig:ood-cifar}
    \end{subfigure}
    \begin{subfigure}[b]{0.49\linewidth}
    \centering
    \includegraphics[width=\linewidth]{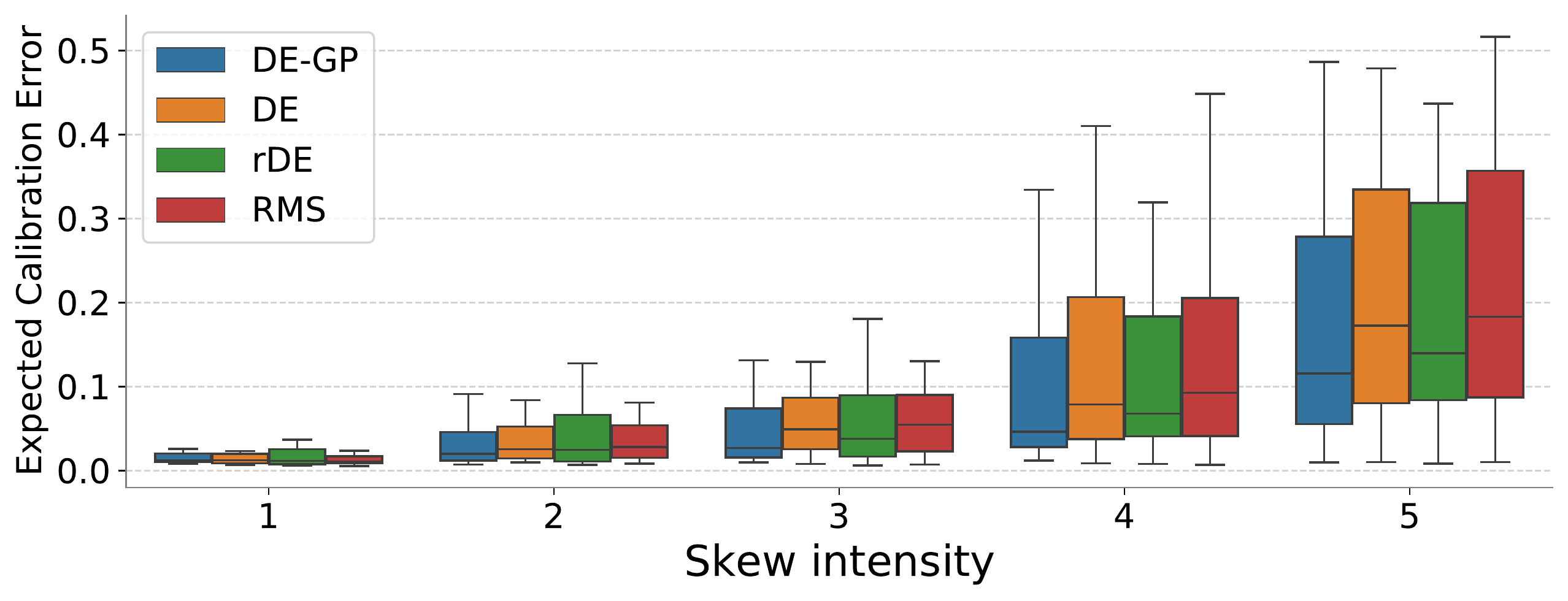}
        % \vspace{-4.ex}
    % \caption{\scriptsize }
        % \label{fig:ood-cifar}
    \end{subfigure}
\end{minipage}
\vspace{-2.ex}
\caption{\small Expected Calibration Error on CIFAR-10 corruptions for models trained with ResNet-20 (Left) or ResNet-56 (Right) architecture. We summarize the results across 19 types of skew in each box.}
\label{fig:cifar-corruptions}
% \vspace{-1ex}
\end{figure*}

\begin{figure*}[!htb]
   \begin{minipage}{0.49\textwidth}
     \centering
     \includegraphics[width=0.49\linewidth]{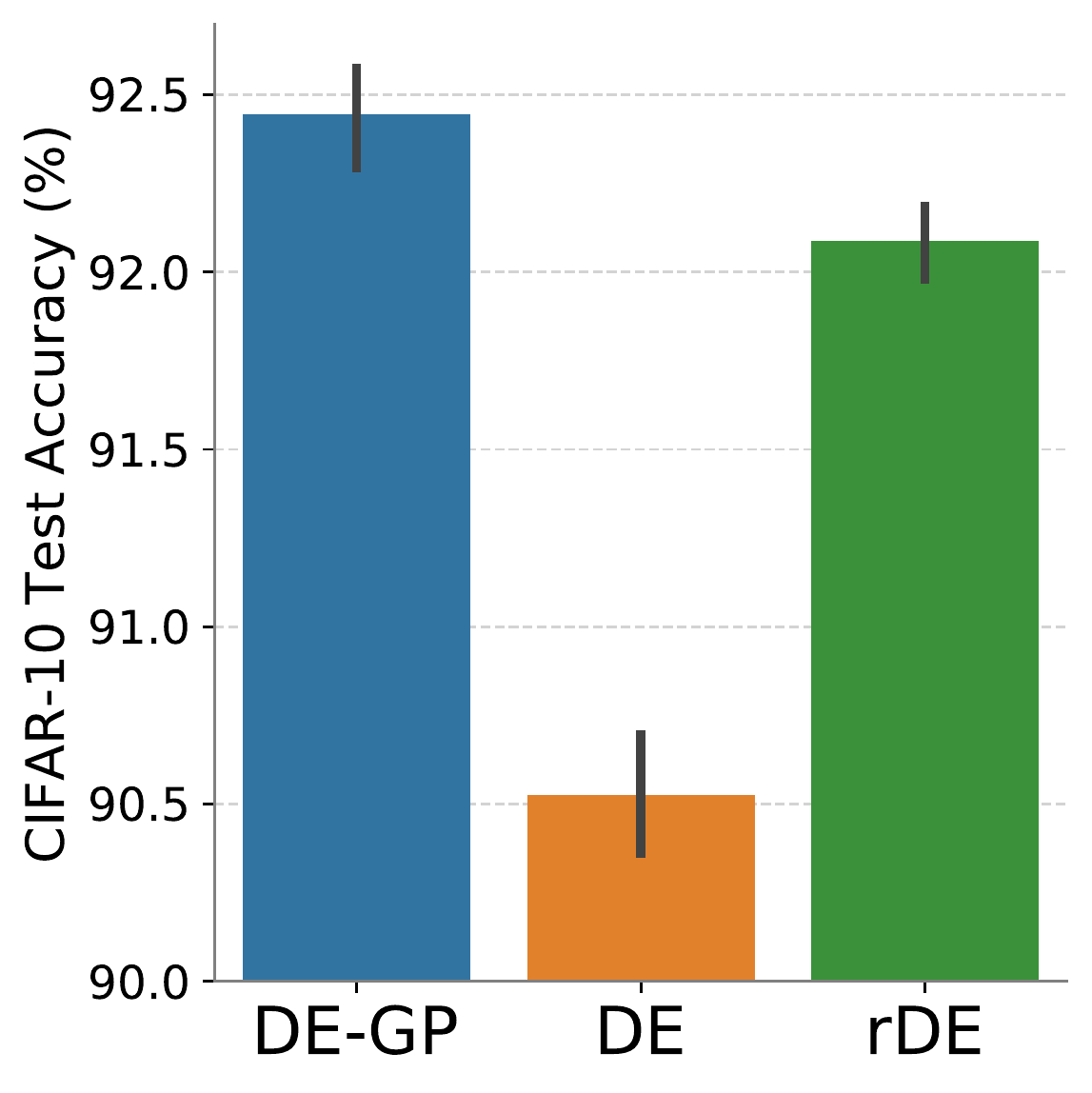}
     \includegraphics[width=0.49\linewidth]{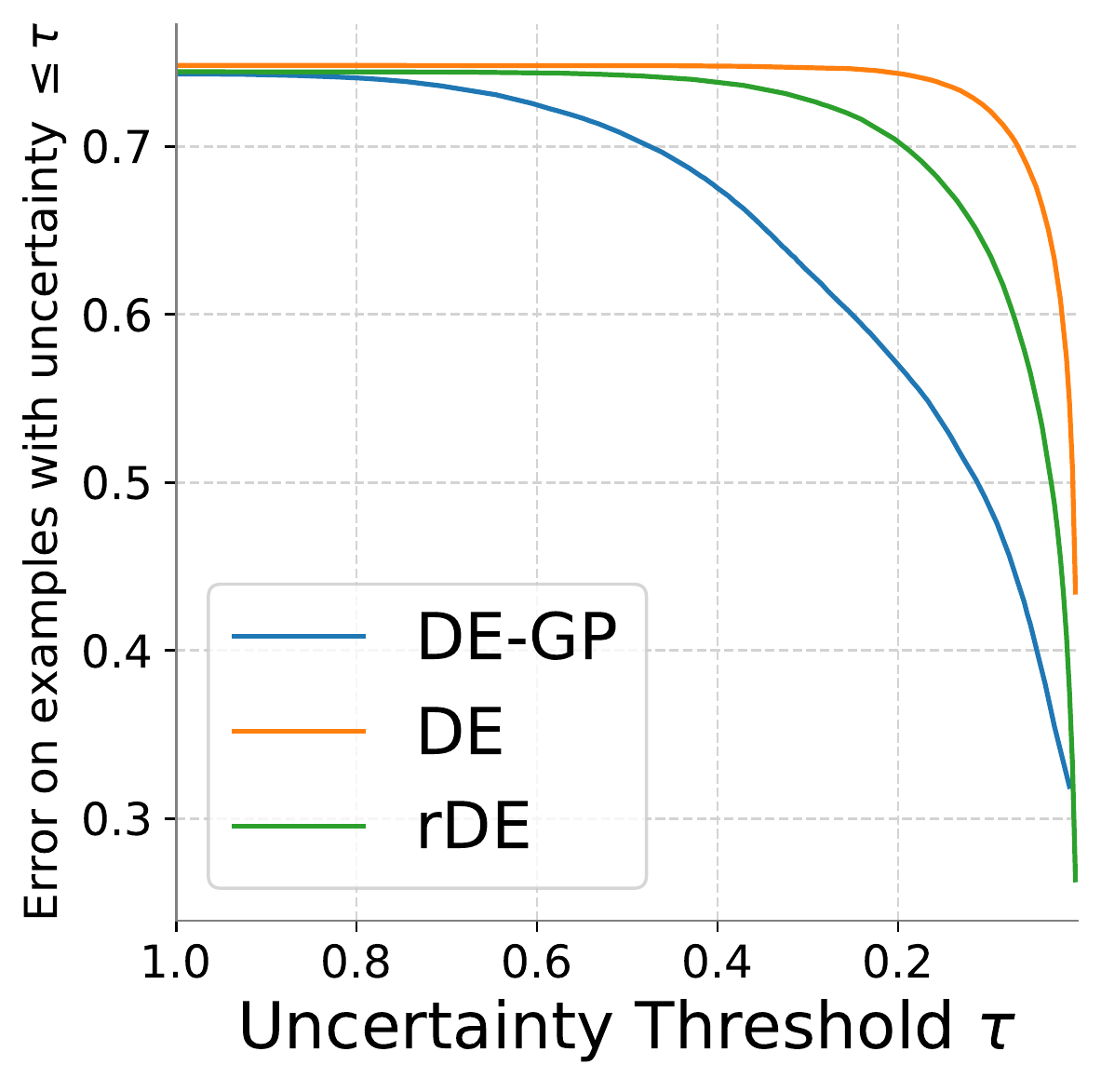}
     \vspace{-1ex}
     \caption{\footnotesize In-distribution test accuracy (Left) and error versus uncertainty plots on the combination CIFAR-10 and SVHN (Right) under weight sharing. (ResNet-20)}
     \label{fig:llrn20-acc-vs-conf}
   \end{minipage}
   \hfill
   \begin{minipage}{0.48\textwidth}
     \centering
     \includegraphics[width=0.95\linewidth]{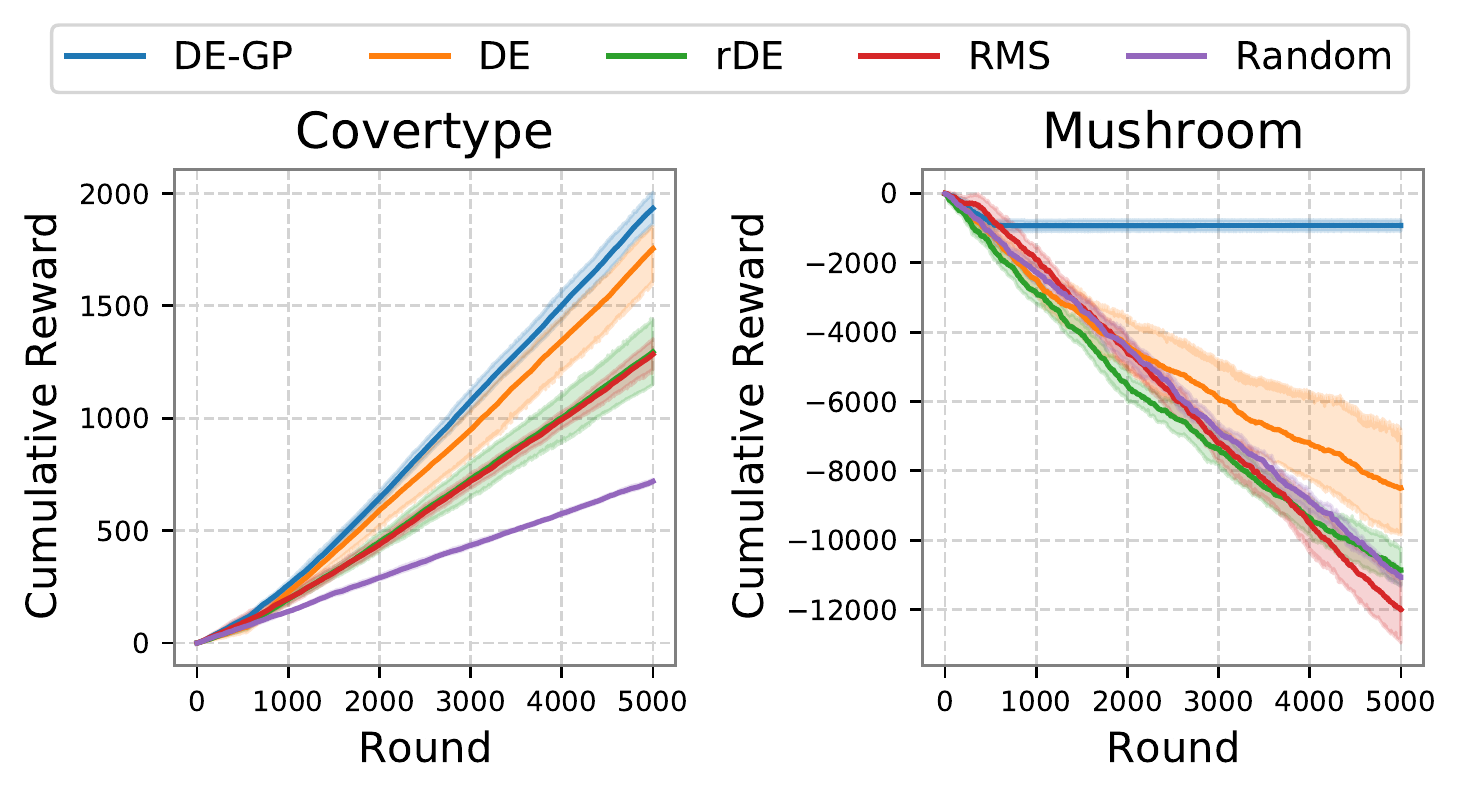}
     \vspace{-1.8ex}
     \caption{\footnotesize Cumulative reward varies w.r.t. round on Covertype (Left) and Mushroom (Right).
    Random corresponds to the Uniform algorithm. Summarized over 5 trials.}
    \label{fig:cb}
   \end{minipage}
 \vspace{-0.1cm}
\end{figure*}

% \vspace{-1ex}
\subsection{UCI Regression}
% \vspace{-1ex}
% Suggested by existing works~\citep{hernandez2015probabilistic,pearce2020uncertainty}, w
We then assess DE-GP on 5 UCI real-valued regression problems.
The used architecture is a MLP with 2 hidden layers of 256 units and ReLU activation. 
10 networks are trained for DE, DE-GP and other variants.
For DE-GP, we set $\beta=0$ and tune $\alpha$ according to validation sets.
%We also tune the hyper-parameters of the baselines for \emph{fairness}.

We perform cross validation with 5 splits. 
\cref{fig:uci} shows the results.
DE-GP surpasses or approaches the baselines across scenarios in aspects of both test negative log-likelihood (NLL) and test root mean square error (RMSE).
% Notably, DE-GP's merits are prominent on the large-scale \emph{Protein} dataset.
DE-GP even beats NN-GP, which is probably attributed to that the variational family specified by DE enjoys the beneficial inductive bias of practically sized SGD-trained DNNs, and DE-GP can flexibly trade off between the likelihood and the prior by tuning $\alpha$. 

% \vspace{-1ex}
\subsection{Classification on Fashion-MNIST and CIFAR-10}
% \vspace{-1ex}
\label{sec:img-clf}
% We next move onto the challenging image classification tasks~\citep{xiao2017fashion,krizhevsky2009learning}.
In the classification experiments, we augment the data log-likelihood (i.e., the first term in \cref{eq:lb-s2}) with a \emph{trainable temperature} to tackle oversmoothing and avoid underconfidence. \footnote{We can make the temperature a Bayesian variable, but it is unnecessary as our model is already Bayesian.}

\textbf{Fashion-MNIST}. We use a widened LeNet5 architecture with batch normalizations (BNs)~\citep{ioffe2015batch} for the Fashion-MNIST dataset~\citep{xiao2017fashion}.
Considering the inefficiency of NN-GP, we mainly compare DE-GP to DE, rDE, and RMS.
% There are 10 ensemble members for each methods.
We set $\alpha$ as well as the regularization coefficients for rDE and RMS all as $0.1$ according to validation accuracy.
For DE-GP, we use $\beta=0$ given the limited capacity of the architecture. 
% We evaluate the trained models on both the in-distribution Fashion-MNIST test set and the out-of-distribution MNIST test set. 
The in-distribution performance is averaged over 8 runs.
\cref{fig:fmnist}-(Left) and \cref{fig:fmnist}-(Middle) display how ensemble size impacts the test results.
Surprisingly, the test error of DE-GP is even lower than the baselines, and its test NLL decreases rapidly as the ensemble size increases.

Besides, to compare the quality of uncertainty estimates, we use the trained models to make prediction and quantify \emph{epistemic uncertainty} for both the in-distribution test set and the out-of-distribution (OOD) MNIST test set. 
All predictions on OOD data are regarded as wrong.
The \emph{epistemic uncertainty} is estimated by the mutual information between the prediction and the variable function: 
\begin{equation}
\label{eq:mi}
\small
\mathcal{I}(f, y|\vx, \mathcal{D}) \approx H\left(\frac{1}{S}\sum_{s=1}^S p(y|f_s(\vx))\right) - \frac{1}{S}\sum_{s=1}^S H\left(p(y|f_s(\vx))\right),
\end{equation}
where $H$ indicates Shannon entropy, with $f_s=g(\cdot, \vw_s)$ for DE, rDE, and RMS, and $f_s \sim q(f; \vw_1,...,\vw_M)$ for DE-GP.
This is a naive extension of the weight uncertainty-based mutual epistemic uncertainty. 
We normalize the uncertainty estimates into $[0, 1]$. 
For each threshold $\tau \in [0, 1]$, we plot the average test error for data with $\leq \tau$ uncertainty in \cref{fig:fmnist}-(Right). 
It is prominent that under various uncertainty thresholds, DE-GP makes fewer mistakes than the baselines, implying that DE-GP succeeds to assign relatively higher uncertainty for the OOD data. 

% the uncertainty estimates of DE-GP are more calibrated than the baselines -- the lower the uncertainty, the more accurate the predictions.

\textbf{CIFAR-10}.
Next, we apply DE-GP to the real-world image classification task CIFAR-10. 
We consider the popular ResNet architectures including ResNet-20 and ResNet-56. 
The ensemble size is fixed as $10$. 
We split the data as training set, validation set, and test set of size 45000, 5000, 10000, respectively.
We set $\beta=0.1$, equivalent with the regularization coefficient on weight in rDE. 
% We select $\alpha$ from $\{0.1, 0.05, 0.01, 0.005\}$ according to validation performance.
We set $\alpha=0.1$ according to an ablation study in \cref{app:alphaabs}. 
We use a lite ResNet-20 architecture without BNs and residual connections to set up the MC NN-GP prior kernel for both the ResNet-20 and ResNet-56 based variational posteriors. 

We present the in-distribution test accuracy in \cref{table:1} and the error versus uncertainty plots on the combination of CIFAR-10 and SVHN test sets in \cref{fig:acc-vs-conf-cifar}. % \ref{fig:cifar-corruptions-nll}, and \ref{fig:cifar-corruptions-acc}.
DE-GP is on par with the practically-used, competing rDE in aspect of test accuracy.
DE-GP ($\beta=0$) shows unsatisfactory
test accuracy, verifying the necessity of incorporating an extra weight-space regularization term when using deep networks. 
The error versus uncertainty plots are similar to those for Fashion-MNIST, substantiating the universality of DE-GP.

We further test the trained methods on CIFAR-10 corruptions~\citep{hendrycks2018benchmarking}, a challenging OOD generalization/robustness benchmark for deep models. 
As shown in \cref{fig:cifar-corruptions} and \cref{app:c10-clf}, DE-GP reveals smaller Expected Calibration Error (ECE)~\citep{guo2017calibration} and lower NLL at various levels of skew, reflecting its ability to make conservative predictions under corruptions.

More results for the deeper ResNet-110 architecture and the more challenging CIFAR-100 benchmark are provided in \cref{app:c10-clf} and \cref{app:cifar100}.

% \begin{figure}[t]
% \vspace{-0.0cm}
% \centering
%     \includegraphics[width=0.49\linewidth]{cifar/rn20ll_acc.pdf}
%     \includegraphics[width=0.49\linewidth]{cifar/acc_vs_conf_cifar_llresnet20.pdf}
% \vspace{-0.7cm}
% \caption{\footnotesize In-distribution test accuracy (Left) and error versus uncertainty plots on the combination CIFAR-10 and SVHN (Right) under weight sharing. (ResNet-20)}
% \label{fig:llrn20-acc-vs-conf}
% \vspace{-2ex}
% \end{figure}

% \begin{figure}[t]
% \vspace{-2.5ex}
% \centering
% \includegraphics[width=\linewidth]{cb/mushroom.pdf}
% \vspace{-4.5ex}
% \caption{\footnotesize Cumulative reward varies w.r.t. round on Covertype (Left) and Mushroom (Right).
% Random corresponds to the Uniform algorithm. Summarized over 5 trails.}
% \label{fig:cb}
% \vspace{-2ex}
% \end{figure}

\textbf{Weight Sharing.} We build a ResNet-20 with 10 classification heads and a shared feature extraction module to evaluate the methods under weight sharing. 
We set a larger value for $\alpha$ for DE-GP to induce higher magnitudes of functional diversity.
The test accuracy (over 8 trials) and error versus uncertainty plots on CIFAR-10 are illustrated in \cref{fig:llrn20-acc-vs-conf}. 
We exclude RMS from the comparison as it assumes i.i.d. ensemble members which may be incompatible with weight sharing.
DE-GP benefits from a function-space diversity-promoting term, hence performs better than
DE and rDE, which purely hinge on the randomness in weight space.
% Find more results in Appendix~\ref{app:c10-clf}.

% \vspace{-1ex}
\subsection{Contextual Bandit}
\vspace{-.5ex}

Finally, we apply DE-GP to contextual bandit, an important decision-making task where the uncertainty helps to guide exploration. 
Following \citep{osband2016deep}, we use DE-GP to achieve efficient exploration inspired by Thompson sampling.
We reuse most of the settings for UCI regression. 
We leverage the GenRL library to build two contextual bandit problems Covertype and Mushroom~\citep{riquelme2018deep}. 
The cumulative reward is depicted in \cref{fig:cb}.
As desired, DE-GP offers better uncertainty estimates and hence beats the baselines by clear margins. 
The potential of DE-GP in more reinforcement learning and Bayesian optimization scenarios deserves future investigation.

\vspace{-1ex}
\section{Conclusion}
% \vspace{-1ex}
In this work, we address the unreliability issue of the uncertainty estimates of Deep Ensemble by defining a Gaussian process with Deep Ensemble and training the model under the principle of functional variational inference. 
Doing so, we have successfully related Deep Ensemble to Bayesian inference to enjoy principled Bayesian uncertainty. 
We offer recipes to make the training feasible, and further identify the necessity of incorporating an extra weight-space regularization term when adopting deep architectures. 
Our method can be implemented easily and efficiently.
Extensive experiments validate the effectiveness of our method.
We hope this work may shed light on the development of better Bayesian deep learning approaches. 

\nocite{langley00}

\bibliography{example_paper}
\bibliographystyle{icml2022}

%%%%%%%%%%%%%%%%%%%%%%%%%%%%%%%%%%%%%%%%%%%%%%%%%%%%%%%%%%%%%%%%%%%%%%%%%%%%%%%
%%%%%%%%%%%%%%%%%%%%%%%%%%%%%%%%%%%%%%%%%%%%%%%%%%%%%%%%%%%%%%%%%%%%%%%%%%%%%%%
% APPENDIX
%%%%%%%%%%%%%%%%%%%%%%%%%%%%%%%%%%%%%%%%%%%%%%%%%%%%%%%%%%%%%%%%%%%%%%%%%%%%%%%
%%%%%%%%%%%%%%%%%%%%%%%%%%%%%%%%%%%%%%%%%%%%%%%%%%%%%%%%%%%%%%%%%%%%%%%%%%%%%%%
\newpage
\appendix
\onecolumn
\section{An Explanation for the Weight-space Regularization}
\label{app:postreg}
Posterior regularization~\citep{ganchev2010posterior,zhu2014bayesian} provides a workaround for Bayesian approaches to impose extra prior knowledge. We can apply posterior regularization to functional variational inference by solving:
\begin{equation}
\small
\max_{q(f)}\mathcal{L}= \mathbb{E}_{q(f)}[\log p(\mathcal{D}|f)] - \KL[q(f)\Vert p(f)]\; \text{s.t.:} \;q(f) \in Q.
\end{equation}
$Q=\{q(f)|\mathbb{E}_{q(f)}\Omega(f)\leq {0}\}$ is a valid set defined in terms of a functional $\Omega$ which delivers some statistic of interest of a function.\footnote{Here we assume one-dimensional outputs for $\Omega$ for notation compactness.} 
For tractable optimization, we can slack the constraint as a penalty:
\begin{equation}
\label{eq:reg-fELBO}
\small
\max_{q(f)}\mathcal{L}'= \mathcal{L} - \beta \max \{\mathbb{E}_{q(f)} \Omega(f), 0\},
\end{equation}
where $\beta$ is a trade-off coefficient.

We next show that the extra weight-space regularization can be derived by imposing the extra prior that \emph{functions drawn from DE-GP should generalize well} to the learning of DE-GP given the above paradigm. 
% \vspace{-0.5ex}
% \subsubsection{Penalize Function (Hypothesis) Complexity}\junz{this is the only subsection of 3.3, we don't need to section-lize this, make the content naturally connected with above.}
% \vspace{-0.5ex}
% The complexity of a function is deeply correlated with its generalization capacity. \guoqiang{1. what does this function mean? hypothesis? 2. generalization capacity? }

\textbf{Binary classification} In the binary classification scenario, $y \in \{-1,1\}$ and $f, g_i: \mathcal{X}\rightarrow \mathbb{R}$.
%we use the margin loss $\ell_{\gamma}(f(\vx), y)=\mathbf{1}_{f(\vx)[y]-\max_{y'\neq y}f(\vx)[y'] < \gamma}$ 
We use $0$-$1$ loss $\ell(f(\vx), y)=\mathbf{1}_{y\neq \text{sign}(f(\vx))}$ to measure the classification error on one datum. 
% Here $\gamma$ is the margin and $f(\vx)[j]$ denotes $j$-th coordinate of the vector $f(\vx)$. 
% Given the dataset $\mathcal{D}$ of $n$ i.i.d. data, the \emph{empirical} risk of a hypothesis (function) is defined as $\hat{R}_{\gamma}(f)=\frac{1}{n}\sum_{i=1}^n \ell_{\gamma}(f(\vx_i), y_i)$.
We assume an underlying distribution $\mu=\mu(\vx, y)$ supported on $\mathcal{X} \times \{-1,1\}$ for generating the training data $\mathcal{D}$, based on which we can define the \emph{true} risk of a function (hypothesis) $f$: $R(f) := \mathbb{E}_{(\vx,y) \sim \mu} \ell(f(\vx), y)$. %$R_0(f) := \mathbb{E}_{(\vx,y) \sim \mu} \ell_{0}(f(\vx), y)$.
We set $\mathbb{E}_{q(f)}\Omega(f):=\mathbb{E}_{q(f)}R(f)$ in the seek of a posterior over functions that can generalize well. %\jianfei{$R$ is both real axis and risk function here. Consider using $\mathbb R$?}

% and we use the 0-1 loss $\ell(f(\vx), y)=\mathbf{1}_{y\neq \text{sign}(f(\vx))}$ to measure the classification error on one datum. 
% Given the dataset $\mathcal{D}$ of $n$ i.i.d. data, the \emph{empirical} risk of a hypothesis (function) is defined as $R_{\mathcal{D}}(f)=\frac{1}{n}\sum_{i=1}^n \ell(f(\vx_i), y_i)$.
% As in typical probably approximately correct (PAC) learning framework, there is an underlying distribution $\mu=\mu(\vx, y)$ supported on $\mathcal{X} \times \mathbb{R}$ for generating $\mathcal{D}$, based on which the \emph{true} risk $R(f) := \mathbb{E}_{(\vx,y) \sim \mu} \ell(f(\vx), y)$ can be defined.

By definition, a hypothesis sample $f \sim q(f)= \mathcal{GP}(m(\vx), k(\vx, \vx'))$ can be decomposed as $f(\vx) = \frac{1}{M}\sum_{i=1}^M g_i(\vx) + \zeta(\vx)$ with $\zeta(\vx) \sim \mathcal{GP}(0, k(\vx, \vx'))$. 
% We then theoretically show that the risk of DE-GP's sample can be bounded from above by the risks of its basis functions.
% Based on this decomposition, we can establish the following theoretical results.
% \begin{restatable}[]{lemm}{Lemm}
% \label{lemm:3}
% For any hypothesis $f$ drawn from the approximate posterior $\mathcal{GP}(m_q(\vx), k_q(\vx, \vx'))$, for any datum $(\vx, y) \in \mathcal{X} \times y$, the following inequality holds:
If $\text{sign}(f(\vx)) \neq y$, it is impossible that $\text{sign}(g_1(\vx)) = y$, ..., $\text{sign}(g_M(\vx)) = y$, and $\text{sign}(\zeta(\vx)) = y$ all hold. 
In other words,
% It is then easy to see that
\begin{equation}
\small
\label{eq:lossf}
{
    \ell(f(\vx), y) \leq  \sum_{i=1}^M[\ell(g_i(\vx), y)] + \ell(\zeta(\vx), y).}
\end{equation}
% \end{restatable}
% This is because a voting is incorrect when at least one of the individuals makes a mistake. 
We can further re-parameterize $\zeta(\vx)$ as $\zeta(\vx)=\frac{1}{\sqrt{M}}\sum_{i=1}^M \epsilon_i (g_i(\vx) - m(\vx)) + \sqrt{\lambda}\epsilon_0, \epsilon_i \sim \mathcal{N}(0, 1), i=0,...,M$, which is essentially a real-valued random function symmetric around 0.
Thus, for any $(\vx, y) \sim \mu$, we have $\mathbb{E}_{q(f)}\ell(\zeta(\vx), y) = \mathbb{E}_{\epsilon_0,...,\epsilon_M} \ell(\zeta(\vx), y) = 1/2$. %since that $\zeta$ is essentially a random guess. %\footnote{This is because $\zeta$ is sampled from a symmetric distribution around 0. N} 
As a result,
\begin{equation}
\small
\begin{aligned}
\mathbb{E}_{q(f)}R(f) &\leq  \mathbb{E}_{q(f)}\mathbb{E}_{(\vx,y) \sim \mu}\sum_{i=1}^M[\ell(g_i(\vx), y)] + \mathbb{E}_{(\vx,y) \sim \mu}[1/2] \\
& = \sum_{i=1}^M [R(g_i)] + 1/2.
\end{aligned}
\end{equation}
Namely, the expected generalization error of the approximately posteriori functions can be bounded from above by those of the DNN basis functions. 
Recalling the theoretical and empirical results showing that DNNs' generalization error $R(g_i)$ can be decreased by controlling model capacity in terms of norm-based regularization $\min_{\vw_i}||\vw_i||_2^2$~\citep{neyshabur2015norm,neyshabur2017exploring,bartlett2017spectrally,jiang2019fantastic}, we obtain an explanation for the extra weight-space regularization. 

\textbf{Multi-class Classification}
In the multi-class classification scenario where $y \in \{1,2,...,C\}$ and $f, g_i: \mathcal{X}\rightarrow \mathbb{R}^C$, we use the loss $\ell(f(\vx), y)=\mathbf{1}_{f(\vx)[y] < \max_{y'\neq y}f(\vx)[y']}$ to measure prediction error where $f(\vx)[j]$ denotes $j$-th coordinate of $f(\vx)$. 
The distinct difference between this scenario and the binary classification scenario is that in this setting, $\zeta(\vx)$ is a vector-valued function:
\begin{equation}
\small
\zeta(\vx)=\frac{1}{\sqrt{M}}\sum_{i=1}^M \epsilon_i (g_i(\vx) - m(\vx)) + \sqrt{\lambda}\vect{\epsilon}_0, 
\end{equation}
where $\vect{\epsilon}_0 \sim \mathcal{N}(\mathbf{0}, \mathbf{I}_C)$ and $\epsilon_i \sim \mathcal{N}(0, 1), i=1,...,M$.
We then make a mild assumption to simplify the analysis.

\begin{restatable}[]{assm}{Assmp}
For any $(\vx, y) \in \mu$, the elements on the diagonal of $k(\vx, \vx)$ have the same value. 
\end{restatable}

This assumption implies that for any $j, j' \in \{1,...,C\}$,
\begin{equation}
\small
\frac{1}{M}\sum_{i=1}^M (g_i(\vx)[j] - m(\vx)[j])^2 + \lambda = \frac{1}{M}\sum_{i=1}^M (g_i(\vx)[j'] - m(\vx)[j'])^2 + \lambda.
\end{equation}
I.e.,
\begin{equation}
\small
\sum_{i=1}^M (g_i(\vx)[j] - m(\vx)[j])^2 = \sum_{i=1}^M (g_i(\vx)[j'] - m(\vx)[j'])^2.
\end{equation}

Therefore, $\zeta(\vx)$ possesses the same variance across its output coordinates and becomes a random guess classifier. 
Based on this, we have $\mathbb{E}_{q(f)}\ell(\zeta(\vx), y) = \mathbb{E}_{\epsilon_0,...,\epsilon_M} \ell(\zeta(\vx), y) = (C-1)/C$.
We can then derive a similar conclusion to that in the binary classification.

\textbf{The validity of Assumption 1.} When the data dimension $|\mathcal{X}|$ is high and the number of training data $n$ is finitely large, with zero probability the sampled data $(\vx, y)\sim \mu$ resides in the training set.
Therefore, only the KL divergence term of the fELBO explicitly affects the predictive uncertainty at $\vx$.
Because the MC NN-GP prior possesses a diagonal structure, it is hence reasonable to make the above assumption.

\section{More of Experiments}
We provide experimental details and more results in this section. 

\subsection{Detailed Settings}
\textbf{Illustrative regression.} 
For the problem on $y=\sin 2x+\epsilon, \epsilon \sim \mathcal{N}(0, 0.1)$, we randomly sample 8 data points from $[-1.5, 1.5]$. 
We add $-1.2$ to the target value of the rightest data point to introduce strong data noise. 
For optimizing the ensemble members, we use a SGD optimizer with $0.9$ momentum and $0.001$ learning rate. 
The learning rate follows a cosine decay schedule. 
The optimization takes $1000$ iterations. 
% We set hidden size as $64$ for the architecture of the MC NN-GP prior kernel for speedup.
The extra measurement points are uniformly sampled from $[-2, 2]$. 
The regularization constant $\lambda$ is set as $1e-4$ times of the average eigenvalue of the central covariance matrices. 

\textbf{UCI regression.} We pre-process the UCI data by standard normalization.
We set the variance for data noise and the weight variance for the prior kernel following \citep{pearce2020uncertainty}.
The batch size for stochastic training is $256$. 
We use an Adam optimizer to optimize for $1000$ epochs.
The learning rate is initialized as $0.01$ and decays by $0.99$ every $5$ epochs.

\textbf{Fashion-MNIST classification.} The used architecture is Conv(32, 3, 1)-BN-ReLU-MaxPool(2)-Conv(64, 3, 0)-BN-ReLU-MaxPool(2)-Linear(256)-ReLU-Linear(10), where Conv($x$, $y$, $z$) represents a 2D convolution with $x$ output channels, kernel size $y$, and padding $z$.
The batch size for training data is $64$. 
We do not use extra measurement points here. 
We use an SGD optimizer to optimize for $24$ epochs.
The learning rate is initialized as $0.1$ and follows a cosine decay schedule. 
We use an Adam optimizer with $1e-3$ learning rate to optimize the temperature.
We use $1000$ MC samples to estimate the posterior predictive and the epistemic uncertainty, because the involved computation is only the cheap softmax transformation on the sampled function values.

\textbf{CIFAR-10 classification.}
We perform data augmentation including random horizontal flip and random crop.
The batch size for training data is $128$. 
We do not use extra measurement points here. 
We use a SGD optimizer with $0.9$ momentum to optimize for $200$ epochs.
The learning rate is initialized as $0.1$ and decays by $0.1$ at $100$-th and $150$-th epochs.
We use an Adam optimizer with $1e-3$ learning rate to optimize the temperature.
We use $1000$ MC samples to estimate the posterior predictive and the epistemic uncertainty. 
Suggested by \citep{ovadia2019can,NEURIPS2020_0b1ec366}, we train models on CIFAR-10, and test them on the combination of CIFAR-10 and SVHN test sets. 
This is a standard benchmark for evaluating the uncertainty on OOD data.

\textbf{Contextual bandit.} We use MLPs with 2 hidden layers of 256 units.
The batch size for training data is 512. We do not use extra measurement points here. 
We update the model (i.e., the agent) for $100$ epochs with an Adam optimizer every $50$ rounds. 
We set $\alpha=1$ and $\beta=0$ without tuning. 
DE, rDE, and RMS all randomly choose an ensemble member at per iteration, but our method randomly draws a sample from the defined Gaussian process for decision. 
This is actually emulating Thompson Sampling and advocated by Bootstrapped DQN~\citep{osband2016deep}.
``Random'' baseline corresponds to the Uniform algorithm.

\begin{figure}[t]
\vspace{-1ex}
\centering
\begin{subfigure}[b]{0.23\linewidth}
    \centering
    \includegraphics[width=\linewidth]{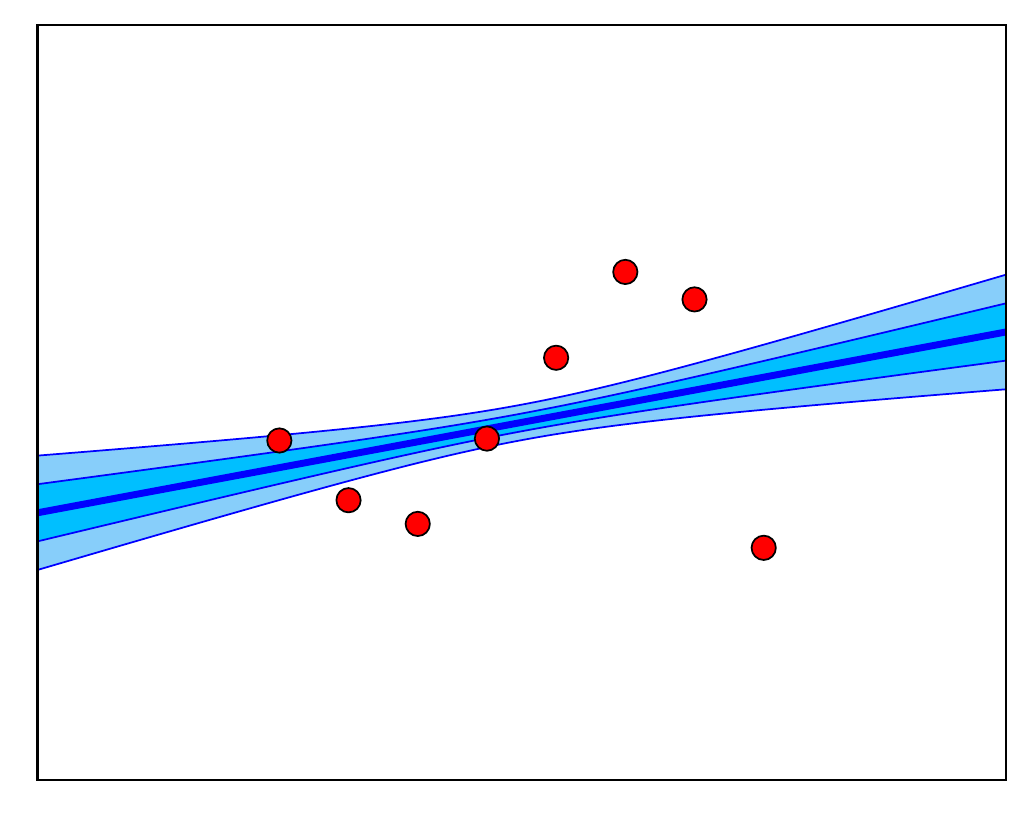}
        \vspace{-4.ex}
       \caption*{\scriptsize MLPs w/o hidden layers}
        % \label{fig:ood-cifar}
    \end{subfigure}
    \begin{subfigure}[b]{0.23\linewidth}
    \centering
    \includegraphics[width=\linewidth]{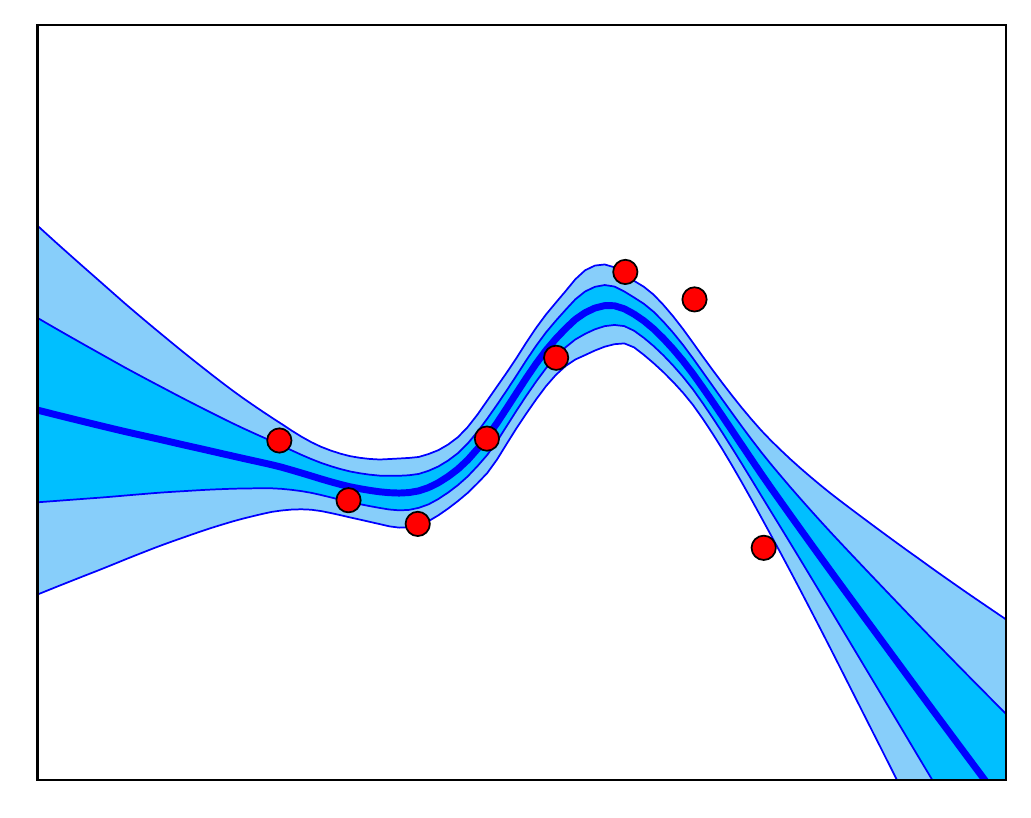}
        \vspace{-4ex}
       \caption*{\scriptsize MLPs w/ 1 hidden layer}
        % \label{fig:ood-imagenet}
    \end{subfigure}
    \begin{subfigure}[b]{0.23\linewidth}
    \centering
       \includegraphics[width=\textwidth]{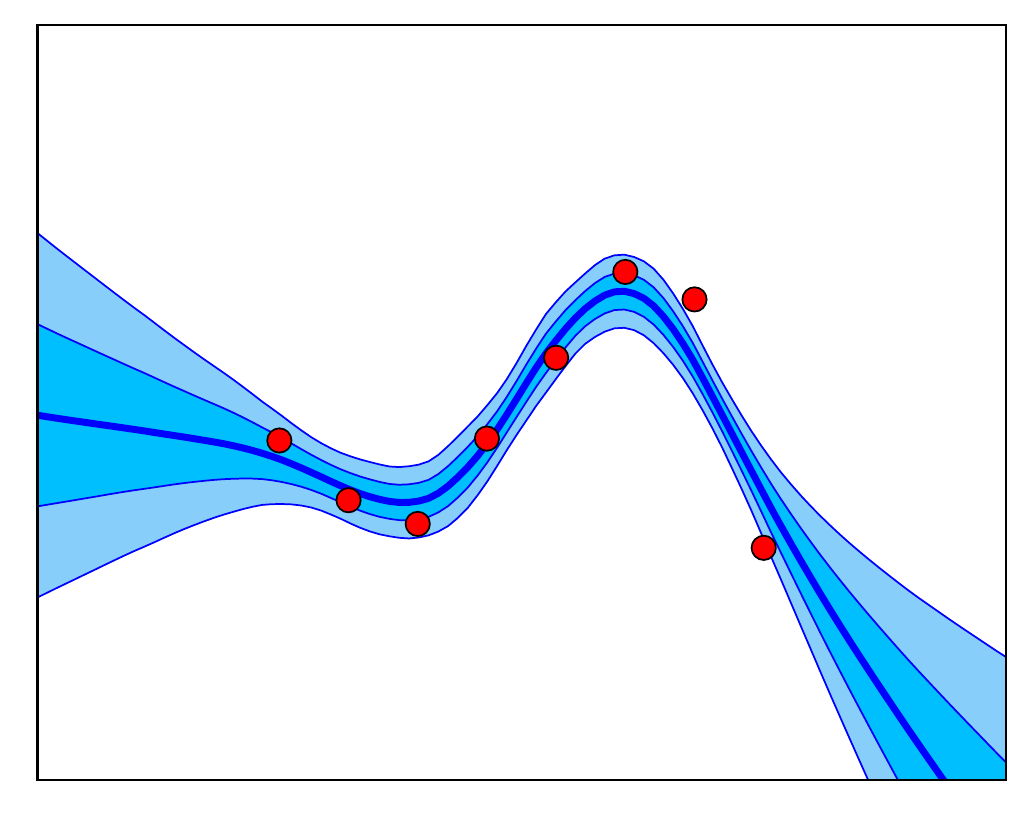}
        \vspace{-4.ex}
       \caption*{\scriptsize MLPs w 2 hidden layers}
        % \label{fig:ood-reg-cifar}
    \end{subfigure}
    \begin{subfigure}[b]{0.23\linewidth}
    \centering
       \includegraphics[width=\textwidth]{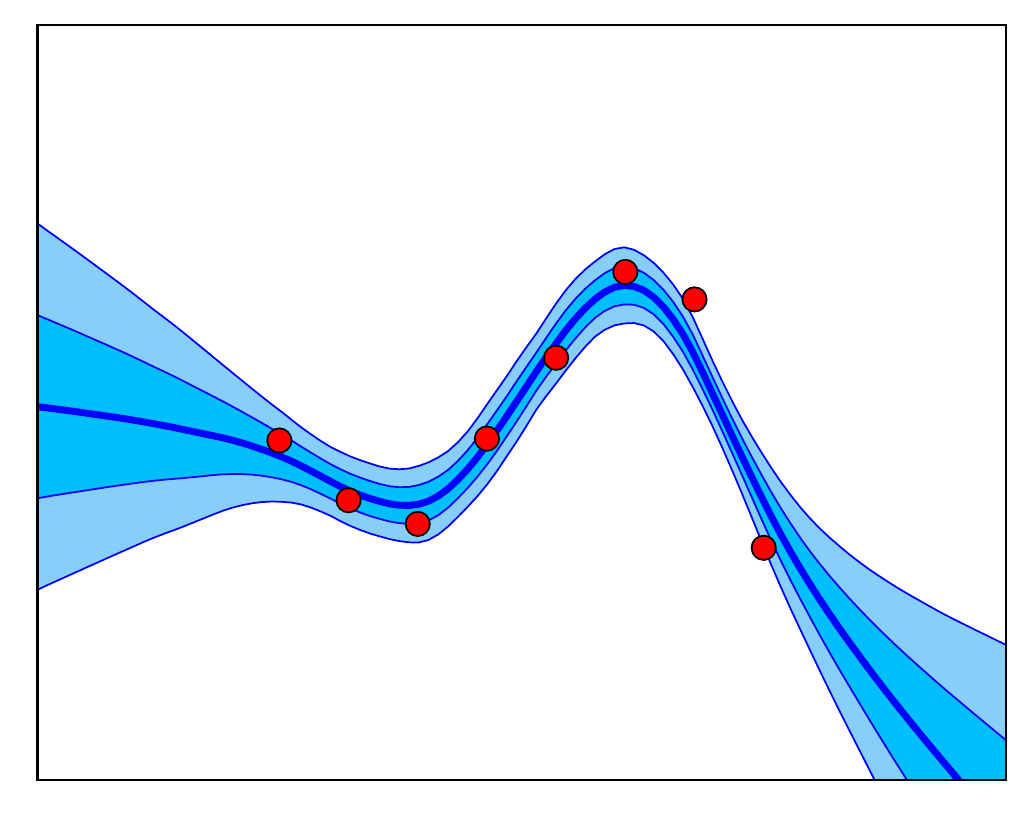}
        \vspace{-4.ex}
       \caption*{\scriptsize MLPs w 3 hidden layers}
        % \label{fig:ood-reg-cifar}
    \end{subfigure}
\vspace{-1.5ex}
\caption{\footnotesize Results of \emph{DE-GP without using extra measurement points}. 
The settings are equivalent to those in \cref{fig:illu}. }
\label{fig:illu-abl}
\vspace{-1.5ex}
\end{figure}

\begin{figure}[t]
% \vspace{-0.1cm}
\centering
\begin{minipage}{0.99\linewidth}
% \begin{figure}
\centering
    \begin{subfigure}[b]{0.49\linewidth}
    \centering
    \includegraphics[width=\linewidth]{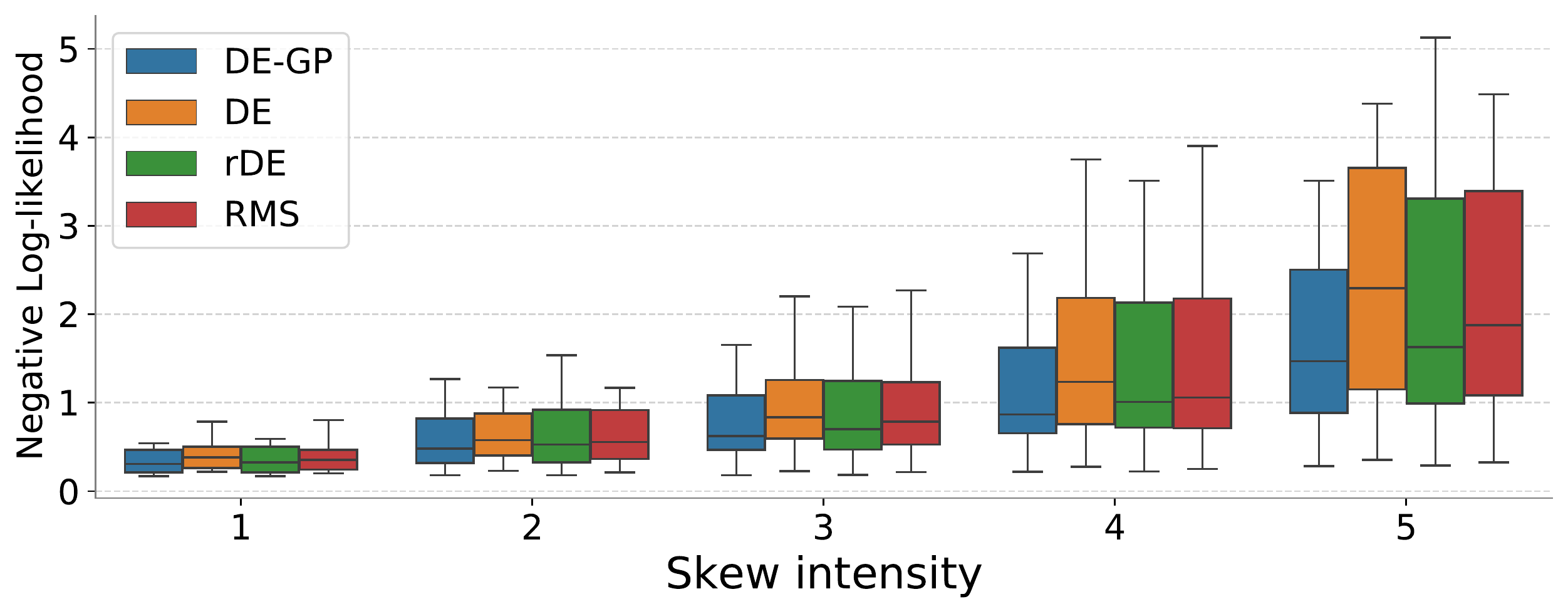}
        % \vspace{-4.ex}
    % \caption{\scriptsize }
        % \label{fig:ood-cifar}
    \end{subfigure}
    \begin{subfigure}[b]{0.49\linewidth}
    \centering
    \includegraphics[width=\linewidth]{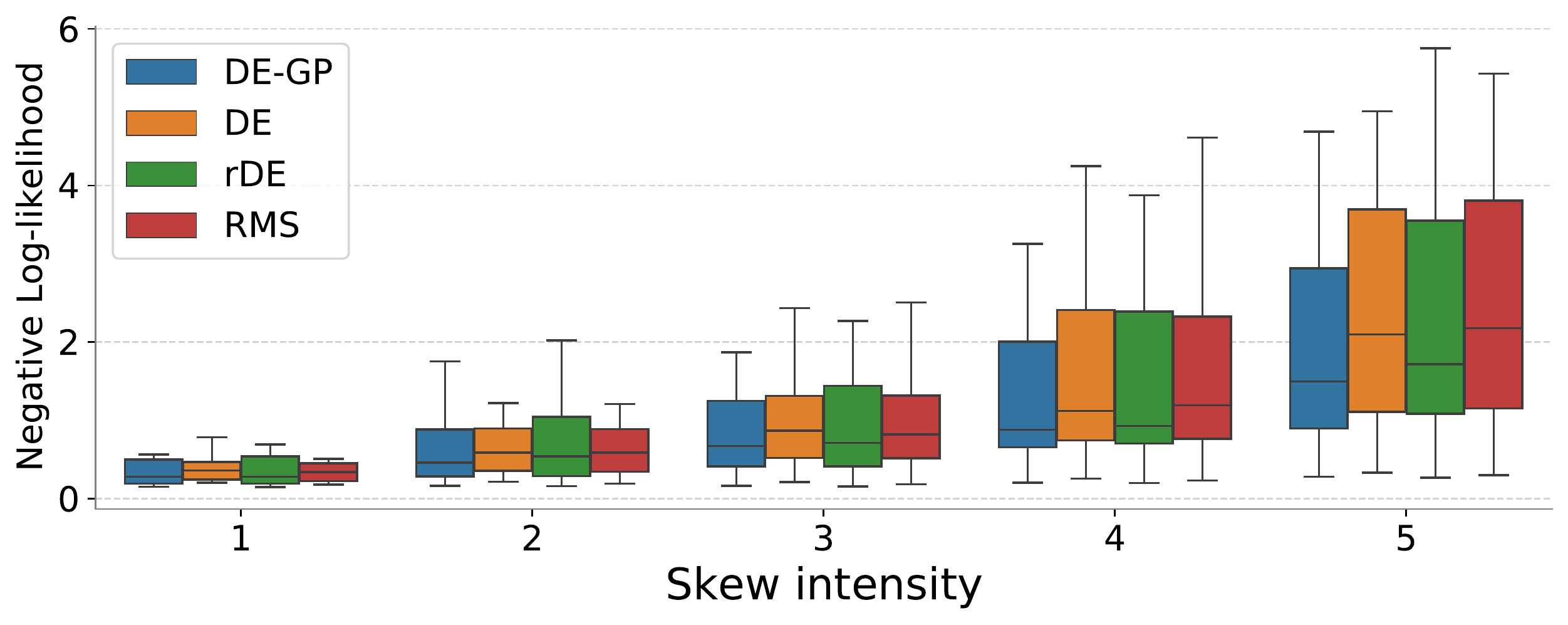}
        % \vspace{-4.ex}
    % \caption{\scriptsize }
        % \label{fig:ood-cifar}
    \end{subfigure}
\end{minipage}
\vspace{-1.5ex}
\caption{\small Negative log-likelihood on CIFAR-10 corruptions for models trained with ResNet-20 (Left) or ResNet-56 (Right) architecture. We summarize the results across 19 types of skew in each box.}
\label{fig:cifar-corruptions-nll}
\end{figure}

\begin{figure}[t]
% \vspace{-0.1cm}
\centering
\begin{minipage}{0.99\linewidth}
% \begin{figure}
\centering
    \begin{subfigure}[b]{0.49\linewidth}
    \centering
    \includegraphics[width=\linewidth]{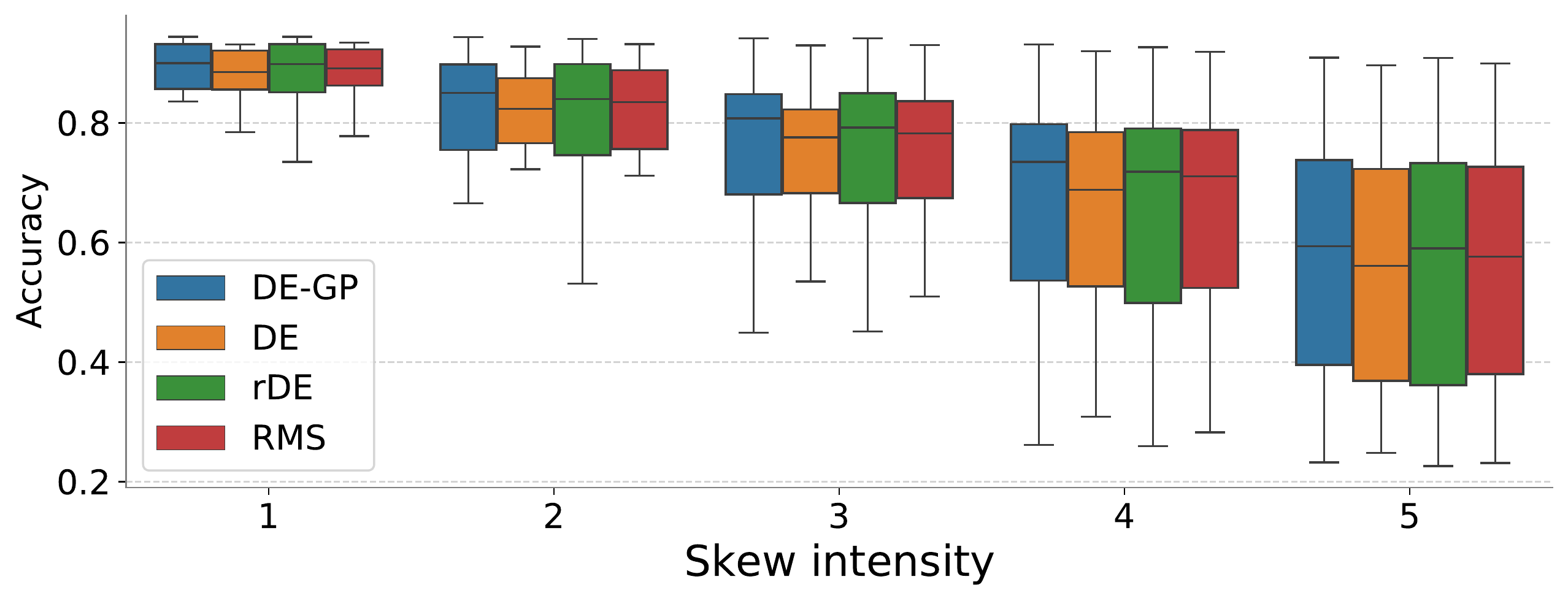}
        % \vspace{-4.ex}
    % \caption{\scriptsize }
        % \label{fig:ood-cifar}
    \end{subfigure}
    \begin{subfigure}[b]{0.49\linewidth}
    \centering
    \includegraphics[width=\linewidth]{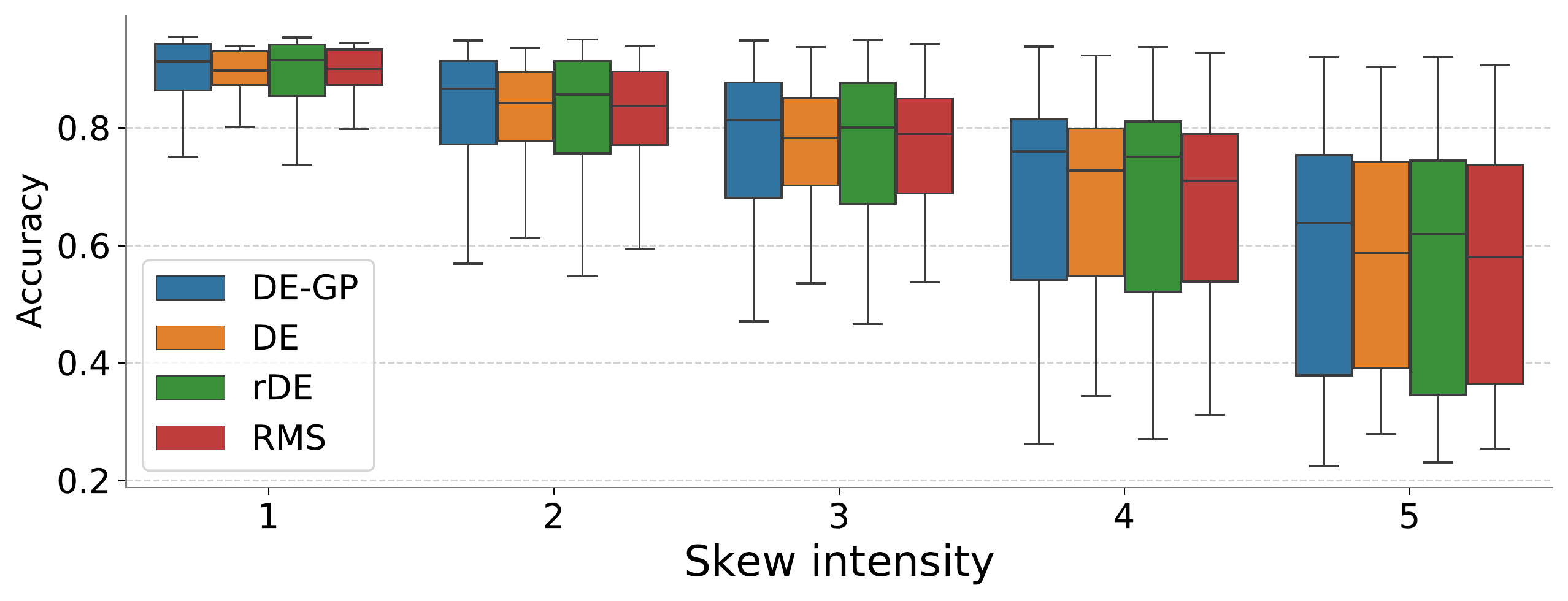}
        % \vspace{-4.ex}
    % \caption{\scriptsize }
        % \label{fig:ood-cifar}
    \end{subfigure}
\end{minipage}
\vspace{-1.5ex}
\caption{\small Test accuracy on CIFAR-10 corruptions for models trained with ResNet-20 (Left) or ResNet-56 (Right) architecture. We summarize the results across 19 types of skew in each box.}
\label{fig:cifar-corruptions-acc}
\end{figure}

\subsection{Results of DE-GP without using Extra Measurement Points}
\label{app:illu-abl}
We depict the results of \emph{DE-GP without using extra measurement points} on $y=\sin 2x+\epsilon, \epsilon \sim \mathcal{N}(0, 0.2)$ problem in \cref{fig:illu-abl}. 
The results are promising.

\subsection{More Results on CIFAR-10 Classification}
\label{app:c10-clf}
We plot the negative log-likelihood and test accuracy on CIFAR-10 corruptions for models trained with ResNet-20 and ResNet-56 in \cref{fig:cifar-corruptions-nll} and \cref{fig:cifar-corruptions-acc}.
As shown, DE-GP outperforms the baselines in aspect of negative log-likelihood, but yields similar test accuracy to the baselines.
Recapping the results in \cref{fig:cifar-corruptions}, DE-GP indeed has improved OOD robustness, but may still face problems in OOD generalization.

\begin{figure}[t]
% \vspace{-0.1cm}
\centering
\begin{minipage}{0.99\linewidth}
% \begin{figure}
\centering
    \begin{subfigure}[b]{0.54\linewidth}
    \centering
    \includegraphics[width=\linewidth]{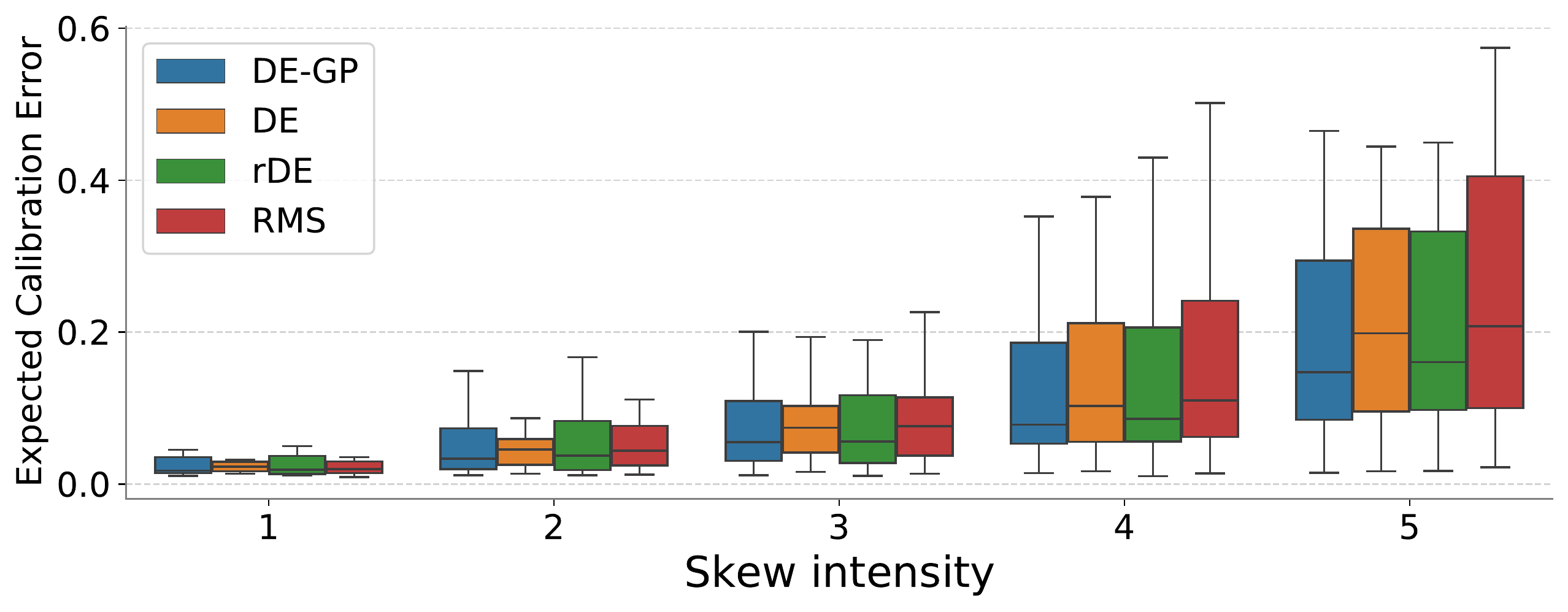}
        % \vspace{-4.ex}
    % \caption{\scriptsize }
        % \label{fig:ood-cifar}
    \end{subfigure}
    \begin{subfigure}[b]{0.45\linewidth}
    \centering
    \includegraphics[width=0.9\linewidth]{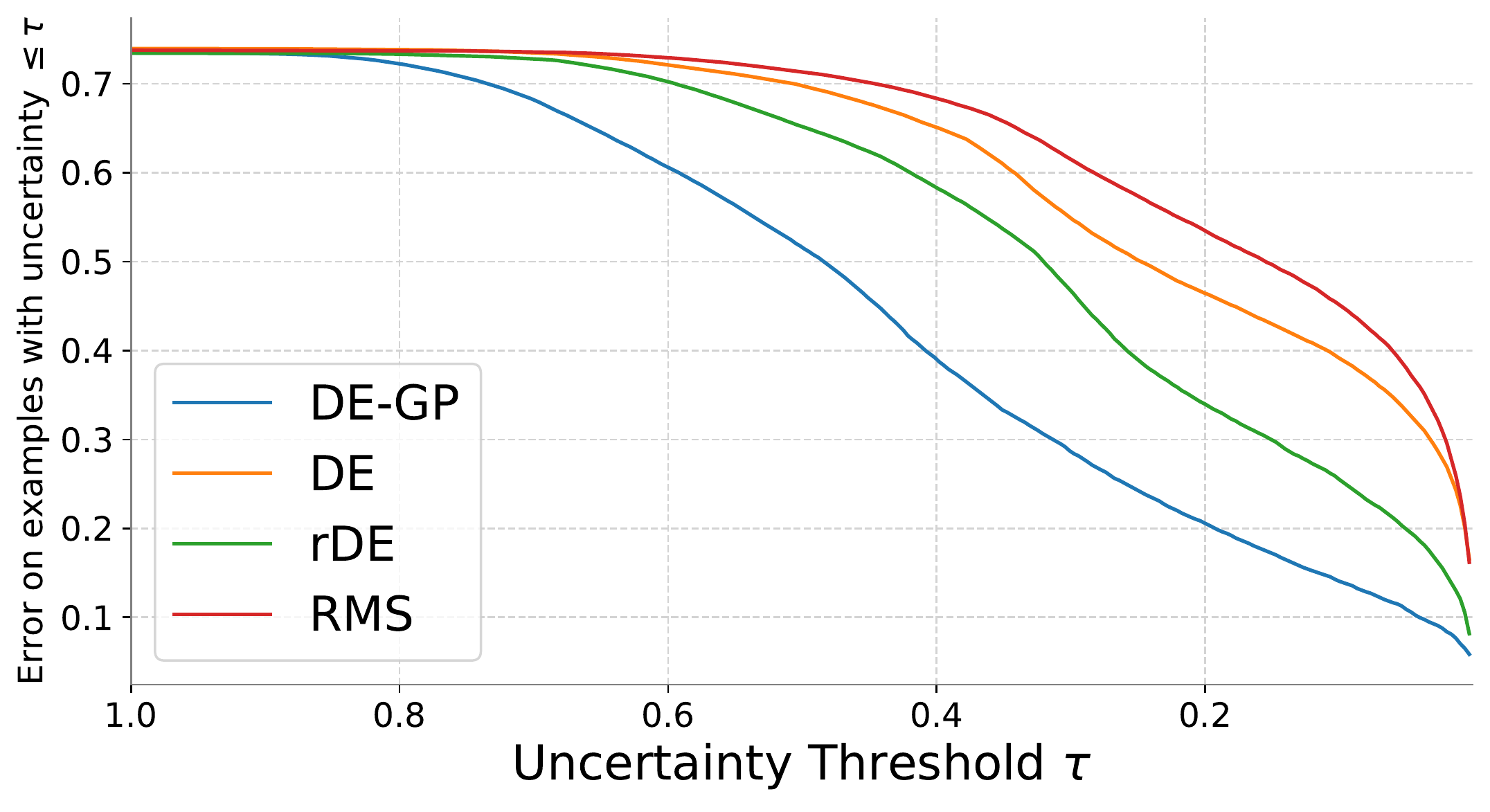}
        % \vspace{-4.ex}
    % \caption{\scriptsize }
        % \label{fig:ood-cifar}
    \end{subfigure}
\end{minipage}
\vspace{-1.5ex}
\caption{\small (Left): Expected Calibration Error on CIFAR-10 corruptions for models trained with ResNet-110 architecture. We summarize the results across 19 types of skew in each box.
(Right): Test error versus uncertainty plots for methods trained on CIFAR-10 and tested on both CIFAR-10 and SVHN with ResNet-110 architecture.
Ensemble size is fixed as 5 for these experiments.}
\label{fig:cifar-rn110}
\end{figure}

We then conduct experiments with the deeper ResNet-110 architecture.
Due to resource constraint, we use $5$ ensemble members. 
The other settings are roughly the same as those for ResNet-56.
The results are offered in \cref{fig:cifar-rn110}, which validate the effectiveness of DE-GP for large networks.

\iffalse
We then outline more results of DE-GP and the baselines on CIFAR-10 corruptions under weight sharing in \cref{fig:cifar-corruption-llrn20}. 
We note that DE-GP has the smallest average Expected Calibration Error, while the variance is high.
The negative log-likelihood of DE-GP is clearly better than that of DE or rDE.
\fi

\iffalse
\begin{figure}[t]
% \vspace{-0.1cm}
\centering
\begin{minipage}{0.99\linewidth}
% \begin{figure}
\centering
    \begin{subfigure}[b]{0.49\linewidth}
    \centering
    \includegraphics[width=\linewidth]{cifar/cifar_llresnet20_corruption_2.pdf}
        % \vspace{-4.ex}
    % \caption{\scriptsize }
        % \label{fig:ood-cifar}
    \end{subfigure}
    \begin{subfigure}[b]{0.49\linewidth}
    \centering
    \includegraphics[width=\linewidth]{cifar/cifar_llresnet20_corruption_0.pdf}
        % \vspace{-4.ex}
    % \caption{\scriptsize }
        % \label{fig:ood-cifar}
    \end{subfigure}
\end{minipage}
\caption{\small Expected Calibration Error (Left) and negative log-likelihood (Right) on CIFAR-10 corruptions for models trained under weight sharing with ResNet-20 architecture. We summarize the results across 19 types of skew in each box.}
\label{fig:cifar-corruption-llrn20}
\end{figure}
\fi

\begin{figure}[t]
\vspace{-0.1cm}
\centering
\begin{minipage}{0.99\linewidth}
% \begin{figure}
\centering
    \begin{subfigure}[b]{0.49\linewidth}
    \centering
    \includegraphics[width=\linewidth]{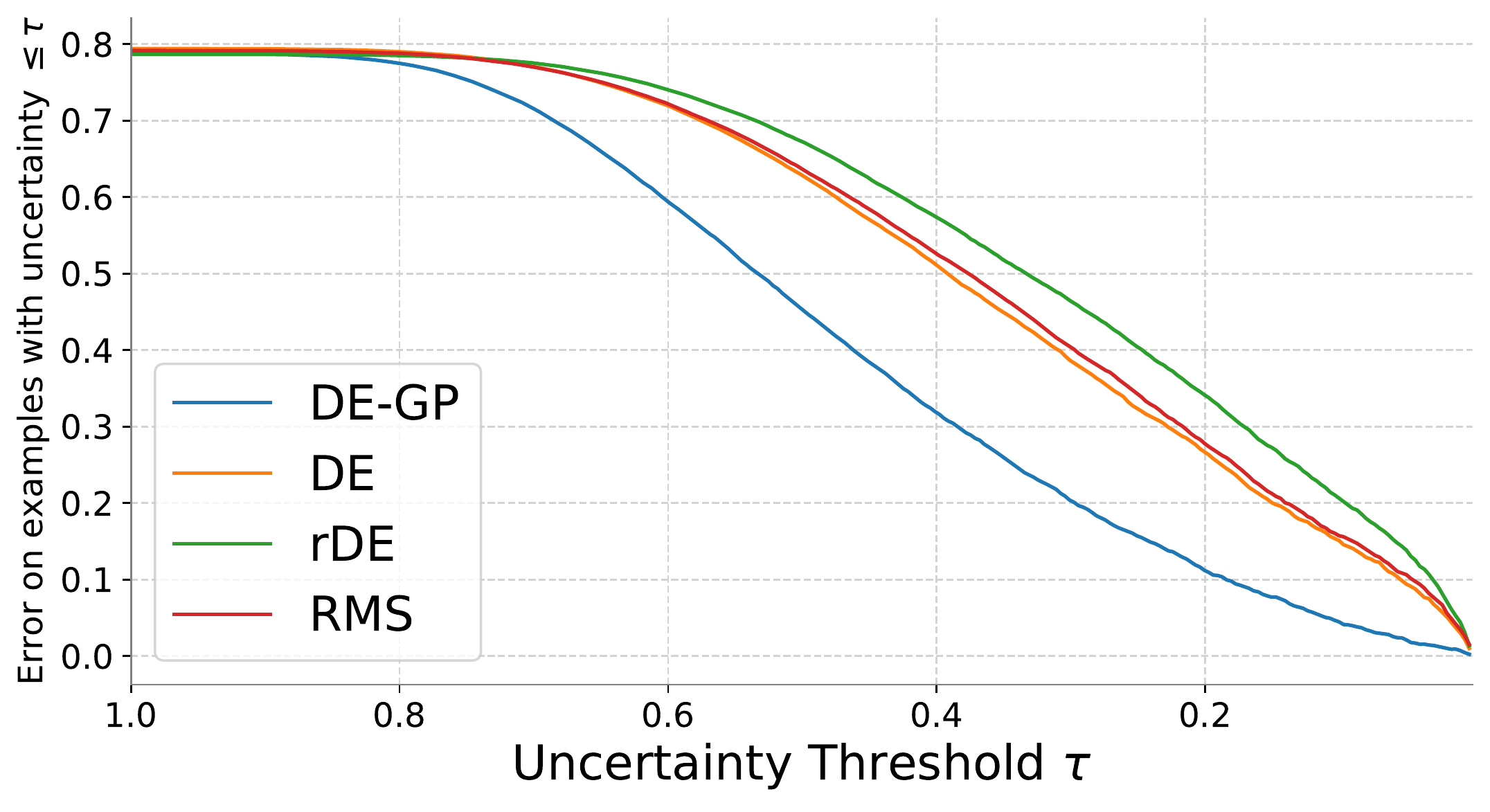}
        % \vspace{-4.ex}
    % \caption{\scriptsize }
        % \label{fig:ood-cifar}
    \end{subfigure}
    \begin{subfigure}[b]{0.49\linewidth}
    \centering
    \includegraphics[width=\linewidth]{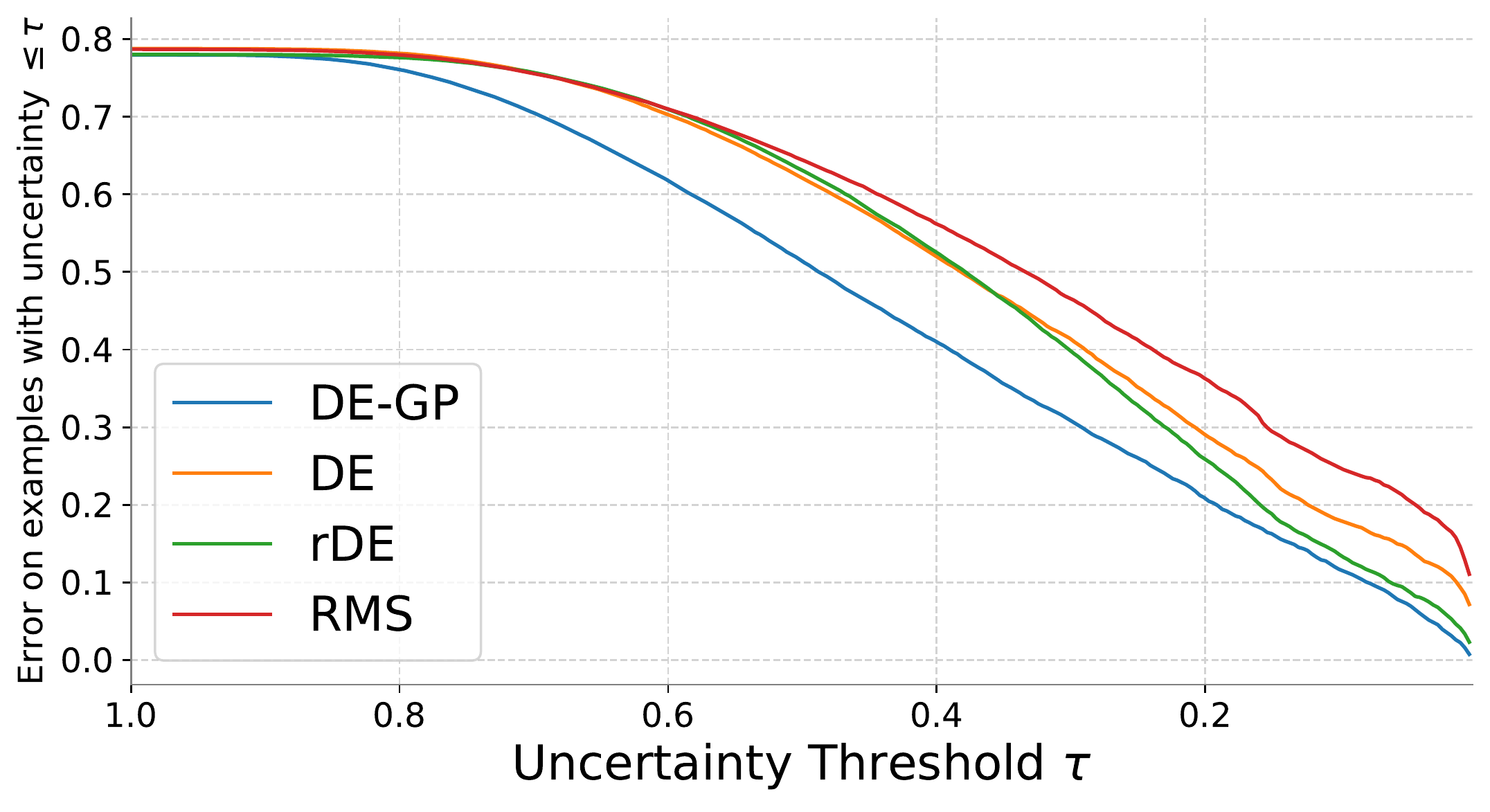}
        % \vspace{-4.ex}
    % \caption{\scriptsize }
        % \label{fig:ood-cifar}
    \end{subfigure}
    % \begin{subfigure}[b]{0.32\linewidth}
    % \centering
    % \includegraphics[width=\linewidth]{cifar/acc_vs_conf_cifar_resnet110.pdf}
    %     % \vspace{-4.ex}
    % \caption{\scriptsize }
    %     % \label{fig:ood-cifar}
    % \end{subfigure}
\end{minipage}
\vspace{-2.ex}
\caption{\small Test error versus uncertainty plots for methods trained on CIFAR-100 and tested on both CIFAR-100 and SVHN with ResNet-20 (Left) or ResNet-56 (Right) architecture.}
\label{fig:acc-vs-conf-cifar100}
\end{figure}

\begin{figure}[t]
% \vspace{-0.1cm}
\centering
\begin{minipage}{0.99\linewidth}
% \begin{figure}
\centering
    \begin{subfigure}[b]{0.49\linewidth}
    \centering
    \includegraphics[width=\linewidth]{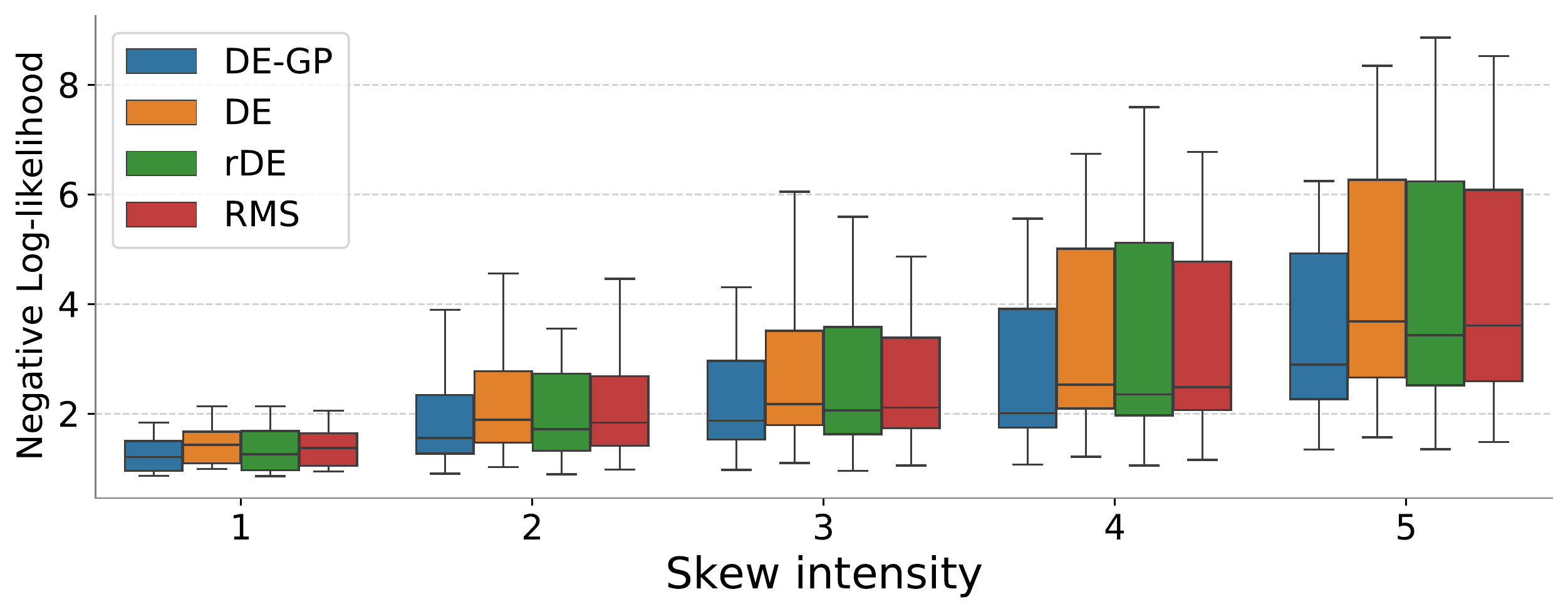}
        % \vspace{-4.ex}
    % \caption{\scriptsize }
        % \label{fig:ood-cifar}
    \end{subfigure}
    \begin{subfigure}[b]{0.49\linewidth}
    \centering
    \includegraphics[width=\linewidth]{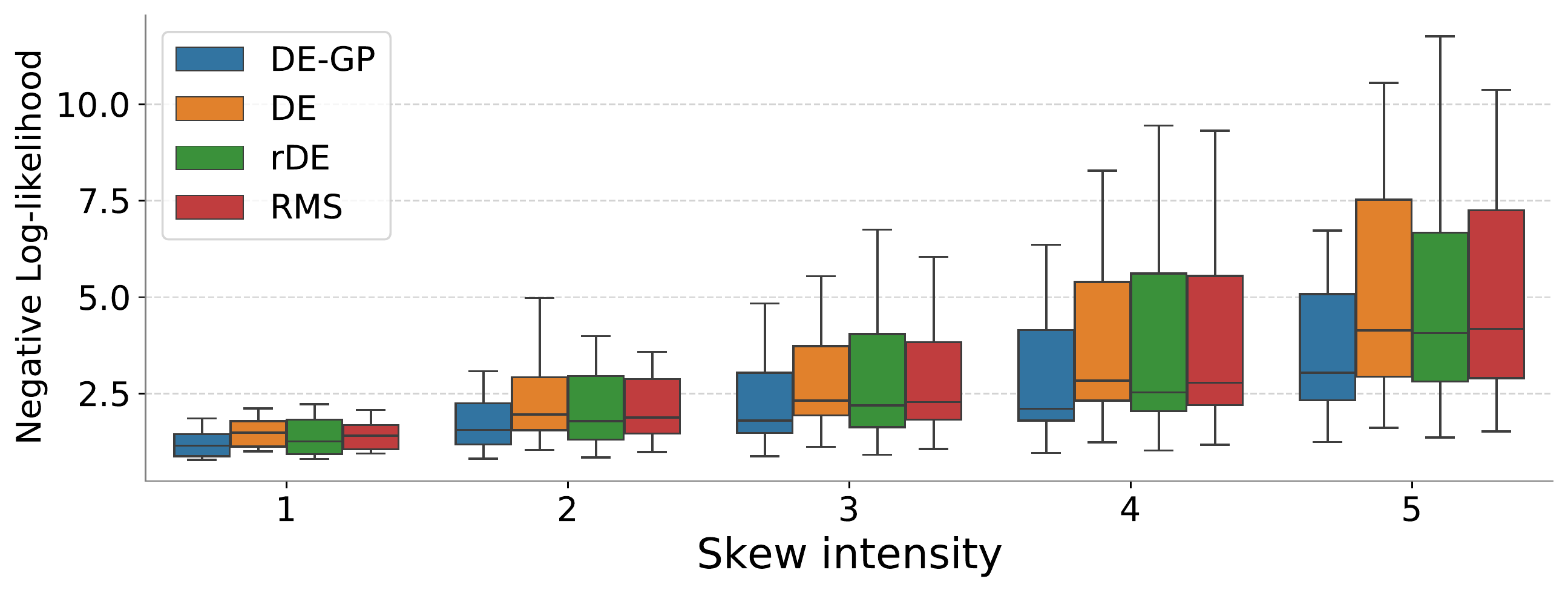}
        % \vspace{-4.ex}
    % \caption{\scriptsize }
        % \label{fig:ood-cifar}
    \end{subfigure}
    \begin{subfigure}[b]{0.49\linewidth}
    \centering
    \includegraphics[width=\linewidth]{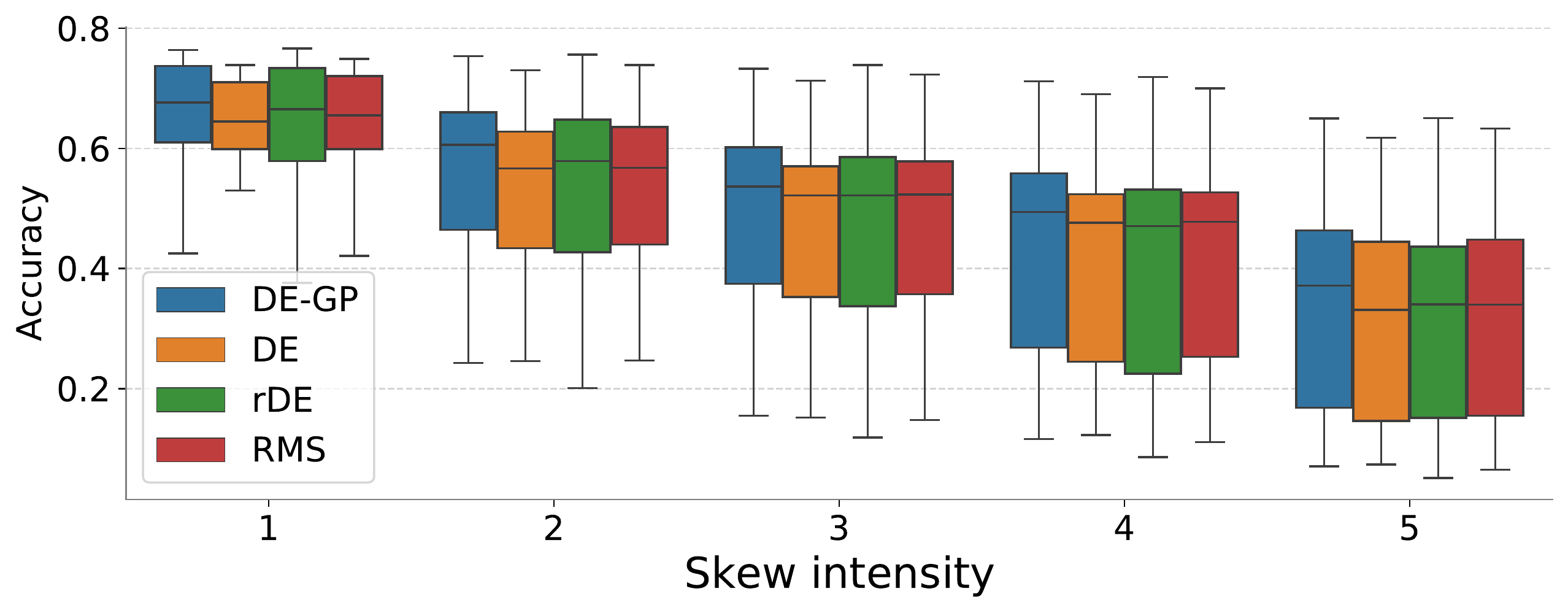}
        % \vspace{-4.ex}
    % \caption{\scriptsize }
        % \label{fig:ood-cifar}
    \end{subfigure}
    \begin{subfigure}[b]{0.49\linewidth}
    \centering
    \includegraphics[width=\linewidth]{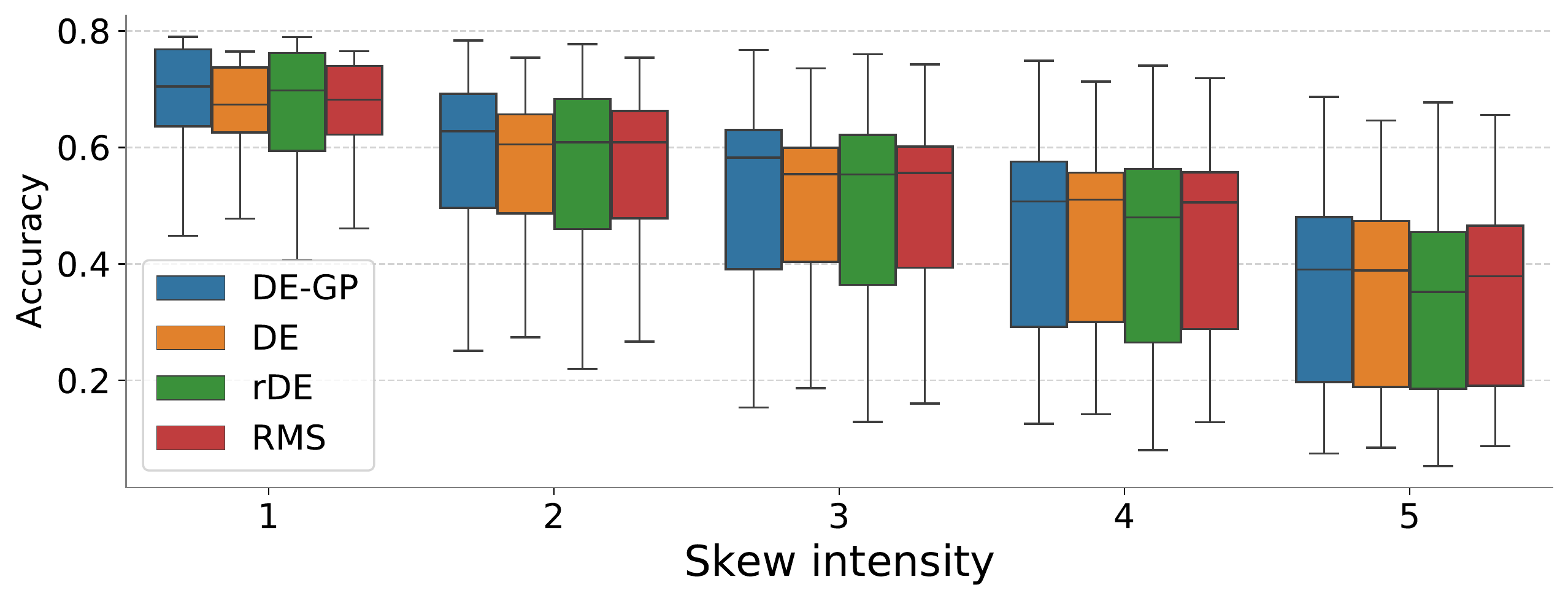}
        % \vspace{-4.ex}
    % \caption{\scriptsize }
        % \label{fig:ood-cifar}
    \end{subfigure}
\end{minipage}
\vspace{-1.5ex}
\caption{\small First row: test NLL on CIFAR-100 corruptions for models trained with ResNet-20 (Left) or ResNet-56 (Right) architecture. Second row: test accuracy on CIFAR-100 corruptions for models trained with ResNet-20 (Left) or ResNet-56 (Right) architecture. We summarize the results across 19 types of skew in each box.}
\label{fig:cifar100-corruptions}
\end{figure}

\subsection{Results on CIFAR-100}
\label{app:cifar100}
We further perform experiments on the more challenging CIFAR-100 benchmark.
We present the in-distribution test accuracy of DE-GP as well as the baselines in \cref{table:cifar100}.
We can see that DE-GP ($\beta=0.1$) is still on par with rDE.
We depict the error versus uncertainty plots on the combination of CIFAR-100 and SVHN test sets in \cref{fig:acc-vs-conf-cifar100}. 
It is shown that the uncertainty estimates yielded by DE-GP for OOD data are more calibrated than the baselines.
We further test the trained methods on CIFAR-100 corruptions~\citep{hendrycks2018benchmarking}, and present the comparisons in aspects of test accuracy and NLL in \cref{fig:cifar100-corruptions}. 
It is evident that DE-GP reveals lower NLL than the baselines at various levels of skew. 
\begin{table*}[h]
  \vspace{-0.2cm}
  \centering
 \footnotesize
\caption{\small Test accuracy comparison on CIFAR-100.}
\vspace{-2ex}
  \label{table:cifar100}
  \begin{tabular}{c||p{17ex}<{\centering}|p{17ex}<{\centering}|p{17ex}<{\centering}|p{17ex}<{\centering} }%|p{17.5ex}<{\centering}}
  \hline
{Architecture}& \emph{DE-GP ($\beta=0.1$)} & \emph{DE} & \emph{rDE} & \emph{RMS} \\% & ResNet-110\\
\hline
ResNet-20 & 76.59\% & 74.14\% & \textbf{76.81}\% & 75.08\% \\
ResNet-56 & \textbf{79.51}\% & 76.46\% & 79.21\% & 76.77\% \\ %use results of resnet20 prior
  \hline
   \end{tabular}
    % \vspace{-0.1cm}
%   \vspace{-2ex}
% \vspace{-0.2cm}
\end{table*}

\subsection{Ablation Study on $\alpha$}
\label{app:alphaabs}
We have conducted an ablation study on $\alpha$ (using ResNet-20 on CIFAR-10).
The results are presented in \cref{table:ablalpha}.
We can see that DE-GP is not sensitive to the value of $\alpha$. 
We in practice set $\alpha=0.1$ in the CIFAR experiments.
We did not use a smaller $\alpha$ as it may result in colder posteriors and in turn worse uncertainty estimates.
\begin{table*}[h!]
%   \vspace{-0.2cm}
  \centering
 \footnotesize
\caption{\small Ablation study on $\alpha$ for DE-GP ($\beta=0.1$) (using ResNet-20 on CIFAR-10).}
\vspace{-2ex}
  \label{table:ablalpha}
  \begin{tabular}{c||p{17ex}<{\centering}|p{17ex}<{\centering}|p{17ex}<{\centering}|p{17ex}<{\centering} }%|p{17.5ex}<{\centering}}
  \hline
{$\alpha$}& 0.1 & 0.05 & 0.01 & 0.005 \\
\hline
Accuracy &94.67$\pm$0.09\% &94.66$\pm$0.07\% & 94.67$\pm$0.04\% & 94.83$\pm$0.10\%
\\% & ResNet-110\\
  \hline
   \end{tabular}
    % \vspace{-0.1cm}
%   \vspace{-2ex}
% \vspace{-0.2cm}
\end{table*}

\subsection{Ablation Study on the Architecture of Prior Kernel}
We perform an ablation study on the architecture for defining the prior MC NN-GP kernel, with the results listed in \cref{table:abl}.
Surprisingly, using the cheap ResNet-20 architecture results in DE-GP with better test accuracy. 
We deduce this is because a deeper prior architecture induces more complex, black-box correlation for the function, which may lead to over-regularization.

\begin{table*}[h!]
  \vspace{-0.2cm}
  \centering
 
 \footnotesize
\caption{\small Ablation study on the architecture of the prior MC NN-GP kernel.}
\vspace{-0.2cm}
  \label{table:abl}
  \begin{tabular}{c||p{12ex}<{\centering}|p{12ex}<{\centering}|p{12ex}<{\centering}}%|p{17.5ex}<{\centering}}
  \hline
\backslashbox{DE-GP architecture}{Prior kernel architecture}& ResNet-20 & ResNet-56 & ResNet-110 \\% & ResNet-110\\
\hline
ResNet-56 (10 ensemble member) & 95.50\% & 95.28\% & - \\
\hline
ResNet-110 (5 ensemble member) & 95.54\% & - & 94.87\%\\ %use results of resnet20 prior
  \hline
   \end{tabular}
    % \vspace{-0.1cm}
%   \vspace{-2ex}
% \vspace{-0.2cm}
\end{table*}
%%%%%%%%%%%%%%%%%%%%%%%%%%%%%%%%%%%%%%%%%%%%%%%%%%%%%%%%%%%%%%%%%%%%%%%%%%%%%%%
%%%%%%%%%%%%%%%%%%%%%%%%%%%%%%%%%%%%%%%%%%%%%%%%%%%%%%%%%%%%%%%%%%%%%%%%%%%%%%%

\end{document}

%% file: math_commands.tex
\usepackage{amsmath,amsfonts,bm}

\def\eqref#1{equation~\ref{#1}}
\def\1{\bm{1}}

\def\vw{{\bm{w}}}
\def\vx{{\bm{x}}}
\def\vy{{\bm{y}}}

\DeclareMathAlphabet{\mathsfit}{\encodingdefault}{\sfdefault}{m}{sl}
\SetMathAlphabet{\mathsfit}{bold}{\encodingdefault}{\sfdefault}{bx}{n}

\newcommand{\KL}{D_{\mathrm{KL}}}